%% file: ms.tex
\documentclass{article}
\pdfoutput=1
\usepackage{ijcai23}

\usepackage[dvipsnames]{xcolor}
\usepackage{multirow}
\usepackage{pgfplots}
\usepackage{tikz}
\usepackage{subcaption}
\usepackage{breakurl}
\usepackage{natbib}

\pgfplotsset{compat=1.18}
\usetikzlibrary{shapes.misc, positioning, shapes, shadows, fit, positioning}

\usepackage{microtype}
\usepackage{graphicx}
\usepackage{booktabs}

\usepackage{hyperref}

\usepackage{amsmath}
\usepackage{amssymb}
\usepackage{mathtools}
\usepackage{amsthm}

\theoremstyle{plain}

\theoremstyle{definition}

\theoremstyle{remark}

\newcommand\independent{\!\perp\!\!\!\perp}

\newcommand\docalc[1]{\mathbf{do}(#1)}

\usepackage{ifthen}

\providecommand\anonym{false}

\newcommand{\ifanonymous}[2]{
    \ifthenelse{\equal{\anonym}{true}}{
        #1
    }{
        #2
    }
}

\title{Can Large Language Models Learn Independent Causal Mechanisms?}

\author{
Ga\"{e}l~Gendron
\quad
Bao~Trung~Nguyen
\quad
Alex~Yuxuan~Peng
\quad
Michael~Witbrock
\quad
Gillian~Dobbie\\
\affiliations
NAOInstitute, University of Auckland
}

\begin{document}

\maketitle

\input{abstract}

\input{introduction}

\input{related_work}

\input{causal_routing}

\input{theory}

\input{experiments}

\input{conclusion}

\input{limitations}

\bibliographystyle{named}
\bibliography{references/introduction, references/related_work, references/causal_routing,references/theory, references/experiments, references/appendix}

\newpage
\appendix
\onecolumn

\input{appendix}


\end{document}

%% file: abstract.tex
\begin{abstract}

Despite impressive performance on language modelling and complex reasoning tasks, Large Language Models (LLMs) fall short on the same tasks in uncommon settings or with distribution shifts, exhibiting a lack of generalisation ability. 
By contrast, systems such as causal models, that learn abstract variables and causal relationships, can demonstrate increased robustness against changes in the distribution. 
One reason for this success is the existence and use of Independent Causal Mechanisms (ICMs) representing high-level concepts that only sparsely interact. 
In this work, we apply two concepts from causality to learn ICMs within LLMs. We develop a new LLM architecture composed of multiple sparsely interacting language modelling modules. 
We show that such causal constraints can improve out-of-distribution performance on abstract and causal reasoning tasks. We also investigate the level of independence and domain specialisation and show that LLMs rely on pre-trained partially domain-invariant mechanisms resilient to fine-tuning.

\end{abstract}

%% file: introduction.tex
\section{Introduction}

The latest generation of Large Language Models (LLMs) with over several billion parameters has demonstrated impressive performance on an extensive range of in-context language and reasoning tasks \citep{bubeck2023sparks, 10.5555/3495724.3495883, NEURIPS2022_9d560961, 51119, DBLP:journals/corr/abs-2302-13971} and an even greater range when fine-tuned for a specific task \citep{DBLP:journals/corr/abs-2307-09288, DBLP:conf/iclr/HuSWALWWC22}.
However, these observations do not hold for tasks that fall outside the training data distribution, sometimes even when the task is only slightly perturbed. 
In particular, standard LLMs perform poorly on complex reasoning tasks, such as abstract, causal, or logical reasoning \citep{DBLP:journals/corr/abs-2307-02477, DBLP:journals/corr/abs-2305-19555, DBLP:journals/corr/abs-2308-13067, DBLP:journals/corr/abs-2306-05836, DBLP:journals/corr/abs-2304-03439, DBLP:journals/corr/abs-2310-09430}. \citet{DBLP:journals/corr/abs-2305-19555, DBLP:journals/corr/abs-2306-05836,DBLP:journals/corr/abs-2307-02477} showed that fine-tuning LLMs can increase their in-distribution performance, but the improvement does not transfer to different distributions, highlighting that LLMs do not generalise as we might expect a person to when applied to domains requiring complex reasoning.
Several hypotheses have been proposed to explain this flaw, such as the lack of abstract or symbolic representations within the latent space of LLMs \citep{DBLP:journals/corr/abs-2307-02477, DBLP:journals/corr/abs-2305-19555, DBLP:journals/corr/abs-2011-15091}. These claims are supported by the brittleness that LLMs can exhibit; when changing the wording of a question, the performance of an LLM can vary drastically \citep{NEURIPS2022_9d560961, DBLP:journals/corr/abs-2306-05836}. This observation hints that LLMs may rely on domain-specific information or spurious correlations in the training data that do not generalise to other distributions. 

Causal models rely on the concept that causal mechanisms invariant under changes in environment exist. The \textit{Independent Causal Mechanisms} principle further states that \textit{``the causal generative process of a system’s variables is composed of autonomous modules that do not inform or influence each other."} \citep{peters2017elements, DBLP:journals/corr/abs-2102-11107}. These principles are applied in diverse ways in the field of causality, either in the structure of the model, which may be built in a modular fashion to respect causal relationships, as in Structural Causal Models \citep{pearl2009causality}, or in the distribution of the data, which may be rendered independent and identically distributed (i.i.d) from an unbalanced distribution by division into subgroups \citep{doi:10.1080/00273171.2011.568786, DBLP:journals/corr/abs-2302-00293}. Integrating these methods into the architecture of a Large Language Model could increase its robustness and out-of-distribution (o.o.d) generalisation. 

We investigate this idea in this work: we aim to better understand how LLMs reason in and out-of-distribution and whether they can behave as models of Independent Causal Mechanisms under certain constraints and with fine-tuning. To this end, we propose an LLM architecture integrating the concept of mechanisms as independent, self-contained LLM modules. 
This model is summarised in Figure \ref{fig:causal-routing}.
We aim to answer the following questions: (i) Can LLMs be used as self-routers for specialised mechanisms, and does it improve their performance? (ii) Can LLMs capture domain-invariant abstractions with information-based regularisation? (iii) How useful is domain-specific knowledge on reasoning tasks? and (iv) Can our proposed architecture approximate Independent Causal Mechanisms?
Our contributions can be summarised as follows:
\begin{itemize}
    \item We propose a modular LLM architecture yielding modularity and abstraction in LLMs using routing and regularisation mechanisms.
    \item We investigate the ability of LLMs to behave as Independent Causal Mechanisms on reasoning tasks and show that it can lead to improved performance and o.o.d generalisation.
    \item We show that LLMs approximate independent mechanism up to an extent but always partially rely on pre-trained domain-invariant mechanisms for reasoning tasks.
\end{itemize}

\ifanonymous{
Our code is available at this anonymous repository: \url{https://anonymous.4open.science/r/iclm-649F}.
}
{
Our code is available at \url{https://github.com/Strong-AI-Lab/modular-lm}.
}

\begin{figure*}
    \centering
    \includegraphics[width=0.9\linewidth]{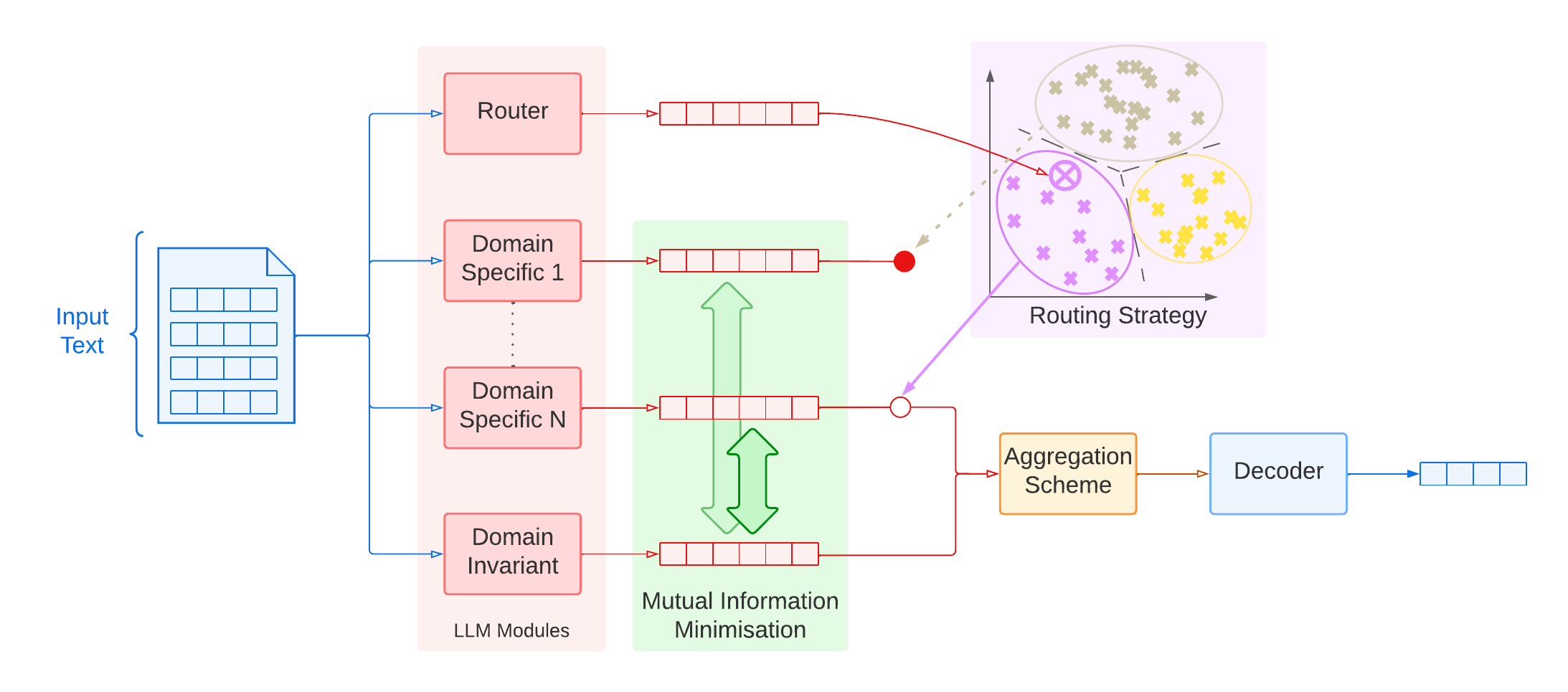}
    \caption{Proposed Independent Causal Language Models (ICLM) architecture for language-modelling tasks. The input text (on the left, in \textcolor{Cyan}{blue}) is fed to multiple pretrained LLM modules (in \textcolor{Salmon}{red}). A router uses clustering on input text embeddings (in \textcolor{Purple}{purple}) to activate a domain-specific module for this input. The domain-invariant module is always activated. The latent representations generated by the activated modules are combined using an aggregation scheme (in \textcolor{Dandelion}{orange}) and converted into a probability distribution for the next word (on the right, in \textcolor{Cyan}{blue}). An additional loss (in \textcolor{Green}{green}) minimises the Mutual Information between the domain-invariant and the domain-specific representations. The router ensures that the domain-specific modules only gain in-domain knowledge while the Mutual Information loss regularises the domain-invariant module towards learning abstract representations.  }
    \label{fig:causal-routing}
\end{figure*}

%% file: related_work.tex
\section{Related Work}

\paragraph{LLM Mixtures-of-Experts}

Modular architectures divide the computations of a network into sub-networks. The Switch Transformer \citep{DBLP:journals/jmlr/FedusZS22} separates the feed-forward layers of the transformer model \citep{DBLP:conf/nips/VaswaniSPUJGKP17} into multiple \textit{expert} modules. 
This strategy allows training larger models at a lower cost, but the expert modules are not guaranteed to specialise in specific domains. Multiple sparse architectures have followed but mainly focus on optimising the training of LLMs for reduced resources and not inducing domain specialisation \citep{DBLP:journals/corr/abs-2209-01667}. 
One exception is the work of \citet{DBLP:conf/naacl/GururanganLHSZ22}, which conditions the activation of an expert module on the input domain. However, only the feed-forward layers are used as experts and the domains are assumed to be known during training. 
\citet{DBLP:conf/icml/ClarkCGMPHDHCB022} investigate the performance of various routing strategies for LLMs and show that the gain from using specialised modules is high for small models but decreases as the model size increases. Introduced recently, Mixtral-of-Experts is a modular LLM using the same routing principle as the Switch Transformer. It outperforms dense LLMs of similar size on reading comprehension, commonsense knowledge and reasoning tasks \citep{DBLP:journals/corr/abs-2401-04088}. However, the authors observe that the routing process does not lead to domain-specialised modules. The assignment of experts is not based on domain information. Our work differs in that it is not directed at optimising LLM training but at inducing functional modularity and studying its effects on generalisation for reasoning tasks. 

\paragraph{Modular Neural Networks}

Other classes of modular neural models are designed to learn specialised sub-networks for specific domains. Recurrent Independent Mechanisms \citep{DBLP:conf/iclr/GoyalLHSLBS21} attempt to learn models of independent mechanisms with an LSTM architecture \citep{DBLP:journals/neco/HochreiterS97} to model the dynamics of physical objects. \citet{DBLP:conf/nips/MittalBL22} investigate routing mechanisms for Mixture-of-Experts models. They find that specialisation can yield better results as the number of tasks increases. However, the learned routing strategies do not capture domain specialisation. In particular, approaches based on backpropagation to the task loss often collapse to a single module.
\citet{DBLP:conf/nips/MittalBL22}'s experiments are restricted to small models and synthetic binary classification and regression tasks; we study a novel routing method via vector quantisation and perform our experiments on architectures over 1B parameters on reasoning tasks.

\paragraph{Causal Models}

Causal models aim to answer queries requiring knowledge of the causal relationships linking the data \citep{DBLP:books/acm/22/BareinboimCII22}. \citet{DBLP:journals/corr/abs-2102-11107, DBLP:journals/corr/abs-2011-15091} argue that for artificial systems to achieve robust and o.o.d reasoning, they must reason in terms of causes and effects and not only correlations, which current LLMs cannot do yet \citep{DBLP:books/acm/22/BareinboimCII22, DBLP:journals/corr/abs-2308-13067}.
Structural Causal Models (SCMs) are graphical models representing causal relationships as mapping functions from parent nodes to their child nodes in a Directed Acyclic Graph \citep{pearl2009causality}. 
If fully specified, an SCM can represent the complete inner workings of a system. However, building an SCM requires access to high-level causal variables, which is not the case in many deep learning tasks that take low-level observations as inputs \citep{DBLP:journals/corr/abs-2102-11107}.
The \textbf{do-calculus}, defined by \citet{pearl1995causal, pearl2009causality}, is used to identify the causal effect of a variable on another with the help of the \textbf{do} operator:
$\docalc{\cdot}$  represents an intervention, i.e. the forced attribution of a value to a variable. If $P(Y|\docalc{X}) = P(Y)$, then $X$ has no causal effect on $Y$ (they may still be correlated if they share common ancestors). 
Another class of causal models relies on determining the flow of \textit{information} in a system \citep{shannon1948mathematical, paluvs2001synchronization, schreiber2000measuring}. However, these concepts have yet to be applied to language models. In the domain of transformers, the Causal Transformer \mbox{\citep{DBLP:conf/icml/MelnychukFF22}} and Causal Attention \mbox{\citep{DBLP:conf/cvpr/YangZQ021}} introduce cross-attention mechanisms to reduce biases from the training distribution.

%% file: causal_routing.tex
\section{Causal Information Routing for LLMs}

We now describe our proposed modular architecture: \textit{Independent Causal Language Models} (ICLM), where each module is an LLM fine-tuned for a specific specialisation or generalisation objective. We aim to build a system that can adapt to changing distributions and capture better abstractions.
Our architecture is separated into $N+2$ \textbf{LLM modules} connected by three main \textbf{components}. The LLM modules are composed of a \textit{router} that generates embeddings of the inputs, a \textit{domain-invariant module} trained to learn abstractions and $N$ \textit{domain-specific modules} trained to specialise on a single task or domain. 
The other components are the \textit{routing strategy}, \textit{Mutual Information loss} and the \textit{aggregation scheme}.
The \textit{routing strategy} uses the embeddings from the LLM router to redirect the inputs to a specific module. Specifically, routing is performed in an unsupervised fashion: each embedding is projected into a clustered space. The centroid of each cluster is associated with a domain-specific module to which it assigns a binary activation weight for a given input. If the input belongs to a cluster, the corresponding module is activated. By contrast, the domain-invariant module processes all inputs. The \textit{Mutual Information loss} induces abstraction within the domain-invariant module; minimising this loss reduces shared information between the domain-specific and domain-invariant modules. I.e. it is intended to cause the domain-invariant module to gain domain-invariant knowledge and the domain-specific modules domain-specific knowledge. Finally, the \textit{aggregation scheme} combines the output of the activated domain-specific module and the domain-invariant module to produce the final output. Figure \ref{fig:causal-routing} shows an overview of our method.
We describe the routing strategy in Section \ref{sec:clustering}, the Mutual Information minimisation in Section \ref{sec:information_flow} and the aggregation scheme in Section \ref{sec:aggregation}. Section \ref{sec:theory} discusses how the architecture reflects Independent Causal Mechanisms.

\subsection{Routing Strategy}
\label{sec:clustering}

The routing strategy redirects the input tokens to a domain-specific module. This step divides the inference into independent modules to increase the specialisation of each module and reduce spurious distribution biases. In particular, the distribution may be imbalanced: a data class may dominate the training distribution and spuriously drive the gradients in a dense model. The routing module is used to balance out the distribution. The inputs belonging to the dominant class are restricted to a single module and cannot adversarially affect the other modules. In parallel, data points far from this data class are trained on a specialised module.

We use a pre-trained LLM (the router) with no final language modelling layer to build an input embedding space. 
The embeddings serve as inputs to the unsupervised routing strategy.
In the strategy, all modules receive the inputs, but the outputs of non-activated modules are \textit{blocked}. I.e. their outputs are associated with a weight of zero (activated modules have a weight equal to one). This activation process by weighting allows us to study more complex (non-binary) weighting schemes, discussed in Appendix \ref{sec:supp_routing_strategy}.
This unsupervised learning method grants more flexibility than the matrix multiplication used in sparse transformers, as any clustering algorithm can be used. In particular, in continual learning settings, one could imagine using a varying number of clusters \citep{ester1996density} and dynamically allocating new modules as data is being fed to the router. In our work, we restrict ourselves to simple clustering methods as we find that they are sufficiently fine-grained for our tasks.
We perform clustering at the input level, i.e. each point in the clustering space represents a complete input context.

\paragraph{Vector Quantisation} We use the vector quantisation procedure introduced for the VQ-VAE \cite{DBLP:conf/nips/OordVK17} as a clustering method. $N$ vectors $h_c$ are arbitrarily initialised in the embedding space, acting as cluster centroids. The attribution of an input to a cluster is determined by measuring the shortest Euclidean distance between them. The router generates an embedding for each token in the input so we measure the distance between a centroid and each token and sum them to obtain the total distance. The location of the centroids is iteratively updated to move closer to the input embeddings using vector quantisation. The corresponding routing loss is defined as follows:

\begin{equation}
    \mathcal{L}_R = \text{MSE}(sg(h_c),h_r) + \nu \cdot \text{MSE}(h_c,sg(h_r))
    \label{eq:loss_r}
\end{equation}

\noindent with $h_r$ one token embedding and $h_c$ the coordinates of the selected centroid, $sg$ is the \textit{stop\_gradients} operation, and $\nu$ is a hyperparameter.
This method has been very successful in transposing high-level concepts from a continuous to a discrete space \citep{DBLP:conf/iclr/Bao0PW22, DBLP:conf/icml/RameshPGGVRCS21} and in building disentangled or interpretable semantic spaces \citep{DBLP:conf/ijcai/GendronWD23, DBLP:journals/corr/abs-2306-17842}.
This approach is simple and assumes clusters with non-overlapping convex hulls. We consider more complex strategies in Appendix \ref{sec:supp_routing_strategy}.

\subsection{Mutual Information Minimisation}
\label{sec:information_flow}

The second aim of the architecture is to induce abstraction and domain-invariance in LLMs. To this end, we introduce a regularisation process based on information theory.
We minimise the Mutual Information (I) \citep{shannon1948mathematical, 1057418} between the domain-specific and domain-invariant modules. Specifically, we minimise the information between the last hidden states of the modules. The idea is to drive the domain-specific modules to gain knowledge specific to their distribution only, while the domain-invariant module gains knowledge common to all distributions and discards the domain-specific information that could be detrimental to generalisation.  
The Mutual Information between two random processes corresponds to the dependence between the two processes, i.e. the amount of information gained on the first process by observing the second one. 
The Mutual Information between two random variables $H_I \in \mathcal{H}$ and $H_S \in \mathcal{H}$ is given by:
\begin{equation}
    \text{I}(H_I, H_S) = \text{KL}(P_{H_I, H_S} ||P_{H_I} \otimes P_{H_S} )
\end{equation}
\noindent where KL is the Kullback-Leibler divergence \citep{kullback1951information}, $H_I$ is the random variable representing the last hidden state of the domain-invariant module and $H_S$ is its counterpart in one domain-specific module.
The hidden states are interpreted as logits distributed in a feature space $\mathcal{H}$. $P_{H_I, H_S}$ is their joint distribution, and $P_{H_I}$ and $P_{H_S}$ are their marginals.
They are later decoded using a final linear layer into the space of possible next words corresponding to the vocabulary of the LLM.
The total loss $\mathcal{L}_\text{I}$ is given by the total information shared between the domain-invariant and all $N$ domain-specific modules:

\begin{equation}
    \mathcal{L}_\text{I} = \sum_{n \in [1,N]} \text{I}(H_I, H_{S_n})
    \label{eq:loss_mi}
\end{equation}

The probabilities $P_{H_I}(h)$ and $P_{H_S}(h)$ cannot be directly computed for any given hidden state $h \in \mathcal{H}$. We can only access the probabilities $P_{H_I}(h|c)$ and $P_{H_S}(h|c)$ for a given input context $c \in \mathcal{C}$. The marginalisation on $\mathcal{C}$ is intractable because of the exponential input space: $|\mathcal{C}| = V^L$, with $V$ the vocabulary size of the LLM and $L$ the maximum length of the input sequence (typically $V^L = (32.10^3)^{4096}$). 
We can approximate it by sampling $\mathcal{C}$ at the batch level $\mathcal{B}$: $P(h) = \sum_{c \in \mathcal{C}} P(h|c) \cdot P(c) \approx \frac{1}{|\mathcal{B}|} \sum_{c \in \mathcal{B}} P(h|c)$ with $|\mathcal{B}| \ll |\mathcal{C}|$. We do the same with the joint distribution $P_{H_I, H_S}$.

\subsection{Aggregation of Outputs}
\label{sec:aggregation}

Before aggregating the domain-invariant and domain-specific modules, we perform a shared batch normalisation \citep{DBLP:conf/icml/IoffeS15} between their last hidden states. For a batch of size $|\mathcal{B}|$, one domain-specific active module and one domain-invariant module, batch normalisation is operated on $2 \times |\mathcal{B}|$ samples. Batch normalisation ensures that the module outputs have the same mean and variance.
We then use a standard language modelling head that converts the hidden states into a probability distribution for the next token. The language modelling head is a  fully connected layer that takes the concatenated hidden states as inputs and outputs a probability distribution in the vocabulary of the language model.
This is a simple aggregation method with great expressivity due to the shared final dense layer. However, this layer can be subject to biases, e.g. if prioritising information from one module at the expense of the others. We study other aggregation schemes, less expressive but more resilient to this issue, in Appendix \ref{sec:supp_aggregation}.

\paragraph{Loss Function}

The total training loss is composed of five components:

\begin{equation}
    \mathcal{L} = \mathcal{L}_o + \alpha \cdot \mathcal{L}_{inv} + \beta \cdot \mathcal{L}_{dom} + 
    \gamma \cdot \mathcal{L}_R + \epsilon \cdot \mathcal{L}_{\text{I}}
    \label{eq:losses}
\end{equation}
\noindent $\mathcal{L}_o$ is the self-supervised cross-entropy loss between the output logits of ICLM and the target text. 
$\mathcal{L}_{inv}$ and $\mathcal{L}_{dom}$ are cross-entropy losses between the output logits of the invariant module and those of the activated domain-specific module. 
$\mathcal{L}_R$ is the vector quantisation loss obtained from the routing strategy (Eq. \ref{eq:loss_r}). $\mathcal{L}_{\text{I}}$ is the Mutual Information loss (Eq. \ref{eq:loss_mi}).
We consider three separate self-supervised losses $\mathcal{L}_o$, $\mathcal{L}_{inv}$ and $\mathcal{L}_{dom}$ to induce the modules to match the target distributions individually and prevent collapse to a single useful module.
$\alpha$, $\beta$, $\gamma$ and $\epsilon$ are constant hyperparameters.

%% file: theory.tex
\section{Theoretical Perspective}
\label{sec:theory}

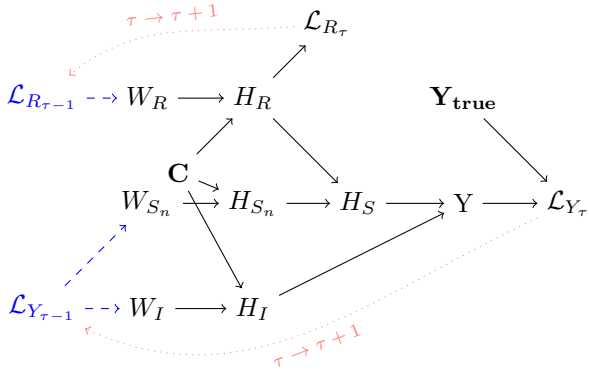
\begin{figure}
    \centering
    \begin{tikzpicture}[node distance=1.4cm]

        \node (sn){$H_{S_n}$};
        \node (s) [right of=sn] {$H_{S}$};
        \node (r) [above of=sn] {$H_R$};
        \node (c) [below left of=r] {\textbf{C}};
        \node (i) [below of=sn] {$H_I$};
        \node (y) [right of=s] {Y};
        \node (yt) [above of=y] {$\mathbf{Y_{true}}$};
        \node (l) [right of=y] {$\mathcal{L}_{Y_{\tau}}$};
        \node (lr) [above right of=r] {$\mathcal{L}_{R_{\tau}}$};
        \node (wn) [left of=sn] {$W_{S_n}$};
        \node (wr) [left of=r] {$W_R$};
        \node (wi) [left of=i] {$W_I$};
        \node (lp) [left of=wi] {\textcolor{blue}{$\mathcal{L}_{Y_{\tau-1}}$}};
        \node (lrp) [left of=wr] {\textcolor{blue}{$\mathcal{L}_{R_{\tau-1}}$}};
        
        \draw[->] (c) -- (r);
        \draw[->] (c) -- (sn);
        \draw[->] (c) -- (i);
        \draw[->] (r) -- (s);
        \draw[->] (sn) -- (s);
        \draw[->] (s) -- (y);
        \draw[->] (i) -- (y);
        \draw[->] (y) -- (l);
        \draw[->] (yt) -- (l);
        \draw[->] (r) -- (lr);
        \draw[->] (wn) -- (sn);
        \draw[->] (wr) -- (r);
        \draw[->] (wi) -- (i);
        \draw[->,blue,dashed] (lp) -- (wn);
        \draw[->,blue,dashed] (lp) -- (wi);
        \draw[->,blue,dashed] (lrp) -- (wr);
        \draw[->,dotted,red!50] (lr) to[out=185,in=40] node[midway, above, sloped] {\scriptsize$\tau \rightarrow \tau+1$} (lrp);
        \draw[->,dotted,red!50] (l) to[out=205,in=335] node[midway, below, sloped] {\scriptsize$\tau \rightarrow \tau+1$} (lp);
        
    \end{tikzpicture}
    \caption{Simplified temporal causal graph $\mathcal{G}$ during training before adding Mutual Information minimisation. C is the input context. $H_R$, $H_I$, $H_{S_n}$, $H_S$ are the latent states of the router, domain-invariant, domain-specific and activated domain-specific (after router weighting) modules. For simplicity, we only show the state $H_{S_n}$ of the activated domain-specific module $n$. $Y$ and $Y_{true}$ are the output and true distributions. $W_R$, $W_{S_n}$ and $W_I$ are the trainable parameters of the modules. $\mathcal{L}_Y = \mathcal{L}_o + \alpha \cdot \mathcal{L}_{inv} + \beta \cdot \mathcal{L}_{dom}$ and $\mathcal{L}_R$ are the output and router losses. Black edges show the forward pass at step $\tau$. \textcolor{blue}{Blue} dashed edges show the backward pass at step $\tau$. \textcolor{red}{Red} dotted edges illustrate the causal links between the forward and backward passes. 
    }
    \label{fig:simplified_training_causal_grah_forward}
\end{figure}

In this section, we provide theoretical evidence on how our model approximates Independent Causal Mechanisms and under what assumptions.
Independent Causal Mechanisms consist of autonomous modules that work independently. In our case, all domain-specific modules are trained for specific tasks/distributions. The domain-invariant module is trained only to use domain-invariant knowledge. The router module is tasked to split the input distribution into $N$ more balanced distributions. We aim to verify that the modules are not causally related.
More formally, we aim to study under what conditions the following holds:

\begin{align}
    &P(H_R|\docalc{H_{S_n}}) = P(H_R)~\forall n \in [1,N]
    \label{eq:indep_router_dom} \\
    &P(H_R|\docalc{H_I}) = P(H_R)
    \label{eq:indep_router_inv} \\
    &\begin{aligned}
    &P(H_{S_n}|\docalc{H_{S_{\hat{n}}}}) = P(H_{S_n}) \\
    &~~~~~~\forall \hat{n} \in [1,N] \setminus \{n\} ~\forall n \in [1,N]
    \end{aligned}
    \label{eq:indep_doms} \\
    &P(H_I|\docalc{H_{S_n}})=P(H_I)~\forall n \in [1,N]
    \label{eq:indep_inv_dom}
\end{align}

$H_R$, $H_I$ and $H_{S_n}~\forall n \in [1,N]$ are the respective representations generated by the router, domain-invariant and $N$ domain-specific modules.

Equations  \ref{eq:indep_router_dom} and \ref{eq:indep_router_inv} are verified. The proof is provided in Appendix \ref{sec:theory_full}; the main idea is that the use of a separate loss function for training the router prevents the other modules from causally acting on the router, either in the forward or backward passes. It can be verified in Figure \ref{fig:simplified_training_causal_grah_forward}.
However, if an invariant module is part of the model, Equations \ref{eq:indep_doms} and \ref{eq:indep_inv_dom} do not hold. The domain-specific modules do not directly influence each other because the routing mechanism allows a single module to go through the forward and backward passes. Nevertheless, a causal path can be drawn through the domain-invariant module as it is always activated. For example, assuming a model with two domain-specific modules, $S_0$ and $S_1$, activated one after the other, a path exists and can be represented in a simplified version as $H_{S_1} \xrightarrow{\dots} \mathcal{L}_{Y_\tau} \xrightarrow{\dots} H_I \xrightarrow{\dots} \mathcal{L}_{Y_{\tau+1}} \xrightarrow{\dots} H_{S_2}$. Again, details of the proof are given in Appendix \ref{sec:theory_full}.
As $H_I$ and $H_{S_n}$ are causally related, we need to reduce the dependency between the two quantities using a regularisation term. Minimising the Mutual Information between $H_I$ and $H_{S_n}$ amounts to reducing the mutual dependence between the variables. $I(H_I,H_{S_n}) = 0$ if and only if $H_I$ and $H_{S_n}$ are independent. If verified, the loss $\mathcal{L}_Y$ can be divided into two independent components and Equations \ref{eq:indep_doms} and \ref{eq:indep_inv_dom} hold. We verify experimentally in Appendix \ref{sec:mi_variation} that the Mutual Information is close to zero after $\sim 50$ training steps.

%% file: experiments.tex
\section{Abstract Reasoning with ICLM}

\input{experiments/setup}

\input{experiments/abstract_reasoning}

\input{experiments/continual_learning}

\section{Routing and Independence Analysis}

\input{experiments/evol_corr}

\input{experiments/routing_visualisation}

%% file: experiments/setup.tex
\subsection{Experimental Setup}
\label{sec:llm}

By default, we use $N=2$ domain-specific modules and one domain-invariant module, as the datasets we use contain two subdomains each. We also perform experiments with an ablated model that does not have a domain-invariant module. In addition, we study the individual performance of the domain-invariant and domain-specific modules.
We use a pretrained LLaMA2-7B \citep{DBLP:journals/corr/abs-2307-09288} for all our modules. We use Low-Rank Approximation of LLMs (LoRA) \citep{DBLP:conf/iclr/HuSWALWWC22} to fine-tune the modules on their respective tasks.
All models are fine-tuned for 3 epochs with AdamW \citep{DBLP:conf/iclr/LoshchilovH19} and a batch size of 16. Loss hyperparameters are $\alpha=0.1$, $\beta = 0.1$, $\gamma = 0.1$, $\epsilon = 0.01$, $\nu = 0.25$.
It is worth noting that the number of parameters used is only marginally higher than that of the base LLaMA2, as only low-memory LoRA adapter weights are learned during training.

\subsection{Datasets}

We perform experiments on the text-based ACRE and RAVEN datasets \citep{DBLP:conf/cvpr/0017JEZZ21, DBLP:conf/cvpr/ZhangGJZZ19, DBLP:journals/corr/abs-2305-19555}\footnote{We use the data provided at \url{https://github.com/Strong-AI-Lab/Logical-and-abstract-reasoning}.}.
ACRE and RAVEN are adapted from Visual Question Answering datasets to be used by language models. The visual ACRE \citep{DBLP:conf/cvpr/0017JEZZ21} is an abstract causal reasoning dataset where the model must deduce the causal mechanisms from a small set of image examples.
The visual RAVEN \citep{DBLP:conf/cvpr/ZhangGJZZ19} is an abstract reasoning dataset where the model must complete a sequence of Raven Progressive Matrices \citep{raven1938raven}.
The text ACRE and RAVEN contain descriptions of the images and instructions for solving the task. The descriptions are provided in two formats: symbolic and natural language. 
The chosen datasets require knowledge of the underlying causal mechanisms to be solved and have o.o.d sets to challenge this ability in the tested systems.

\paragraph{Out-of-Distribution Regimes}

Each dataset has two o.o.d regimes. In ACRE, the \textit{compositionality} split changes the composition of the context examples: combinations of figure shapes and colours unseen in the training set are proposed; the \textit{systematicity} split alters the distribution of the context example activations: the context contains more positive examples than in the training set. In RAVEN, the \textit{four} split contains four figures instead of one; the \textit{in-center} split describes two figures with one containing the other instead of being placed next to each other.

%% file: experiments/abstract_reasoning.tex
\subsection{Abstract and Causal Reasoning}

The results obtained on the ACRE and RAVEN datasets are shown in Table \ref{tab:abstract_reasoning_specific_results}. 
The proposed ICLM can outperform the baseline, particularly on the most challenging o.o.d sets. Moreover, the performance of the individual domain-invariant modules highlights that the modules have learned more generalisable knowledge than with standard training. The domain-specific modules compete with the baselines trained on the corresponding specific domain, showing that the router accurately distributes the inputs to the right modules. The modules even outperform the oracle router on RAVEN in almost all settings. We investigate a potential reason for this phenomenon in Section \ref{sec:routing_alignment}.

\begin{table*}[ht]
    \centering
    \scriptsize
    \addtolength{\tabcolsep}{-0.1em}
    \begin{tabular}{lcccccc}
        \hline
         & \multicolumn{2}{c}{ACRE} & \multicolumn{2}{c}{-o.o.d-Comp} & \multicolumn{2}{c}{-o.o.d-Sys} \\
        \cline{2-7}
         & Text & Symb & Text & Symb & Text & Symb \\
        \hline
        LLaMA2-Base & 0.014 & 0.003 & 0.244 & 0.001 & 0.288 & 0.001 \\
        -Finetuned-All* & 0.832 & 0.891 & 0.832 & 0.881 & \textit{0.911} & \textit{0.891} \\
        ICLM* (ours) & 0.653 & \textbf{0.950} & 0.663 & \textit{0.931} & 0.634 & \textbf{0.901} \\
        ICLM-No-Inv* (ours) & \textit{0.871} & \textit{0.921} & \textit{0.842} & \textbf{0.941} & 0.822 & \textit{0.891} \\
        ICLM-Invariant* (ours) & \textbf{0.891} & \textit{0.921} & \textbf{0.851} & \textbf{0.941} & \textbf{0.921} & \textit{0.891} \\
        \hline
        ICLM-Domain* (ours) & 0.871 & 0.911 & 0.822 & 0.901 & 0.822 & 0.891 \\
        \hline
        -Finetuned-Oracle-Router & 0.997 & 1.000 & 1.000 & 1.000 & 0.994 & 0.999 \\
        \hline
    \end{tabular}
    \hspace{0.05cm}
    \begin{tabular}{lcccccc}
        \hline
         & \multicolumn{2}{c}{RAVEN} & \multicolumn{2}{c}{-o.o.d-Four} & \multicolumn{2}{c}{-o.o.d-In-Center} \\
        \cline{2-7}
         & Text & Symb & Text & Symb & Text & Symb \\
        \hline
         & 0.026 & 0.149 & 0.073 & 0.121 & 0.000 & 0.001 \\
         & \textit{0.990} & \textbf{1.000} & \textit{0.673} & \textbf{0.743} & \textbf{0.673} & 0.198 \\
         & \textbf{1.000} & 0.980 & \textbf{0.703} & \textit{0.703} & 0.515 & \textit{0.228} \\
         & \textbf{1.000} &  0.732 & 0.525 & 0.515 & 0.455 & 0.168 \\
         & \textbf{1.000} & \textit{0.990} & 0.634 & 0.693 & \textit{0.554} & \textbf{0.238} \\
        \hline
         & 0.980 & 0.980 & 0.604 & 0.634 & 0.386 & 0.228 \\
        \hline
         & 0.977 & 0.965 & 0.557 & 0.442 & 0.536 & 0.064 \\
        \hline
    \end{tabular}
    \caption{Accuracy on the ACRE and RAVEN i.i.d and o.o.d test sets. ``Finetuned-All'' is a single LLaMA2 model fine-tuned on text and symbolic i.i.d training sets. ``Finetuned-Oracle-Router'' is an ensemble of two LLaMA2 models fine-tuned each i.i.d training set (either text or symbolic) and routed via a ground-truth oracle. ICLM is trained on text and symbolic i.i.d training sets. ICLM-Invariant shows the results for the domain-invariant module alone. ICML-Domain shows the results for the domain-specific module that aligns best with the dataset (see Appendix \ref{sec:supp_routing_alignment}). ICLM-No-Inv is an ablated ICLM with no domain-invariant module. Models with a $^*$ indicate that that this paper introduces the results. The best model is in \textbf{bold}, and the second best is in \textit{italics}. ICLM outperform LLaMA2 on most sets and individual modules even outperform the oracle on the more challenging RAVEN. }
    \label{tab:abstract_reasoning_specific_results}
\end{table*}

%% file: experiments/continual_learning.tex
\subsection{Continual Learning}

We investigate the capacity of our model to be used in continual learning settings. Continual learning consists of training a model with continuous data streams or sets evolving over time, where the model acquires and accumulates knowledge incrementally. The main challenge lies in the \textit{catastrophic forgetting} of the previous knowledge when gaining new information \citep{DBLP:journals/corr/abs-2302-00487}.
We study a simple usecase where we want our model to learn one new task after training on a previous task. We choose the scenario $ACRE \rightarrow RAVEN$ as RAVEN is more challenging, particularly the o.o.d sets. The results are shown in Table \ref{tab:continual_learning_results}.
The domain-invariant module can use general information extracted from ACRE to improve its performance on RAVEN, even outperforming the baseline trained on RAVEN only. The domain-specific modules can also partially mitigate the catastrophic forgetting problem observed in LLaMA2. Their weights are not activated by the router on RAVEN inputs, thus not updated, and their performance on ACRE is preserved. However, the aggregation process is affected, leading to reduced performance on ACRE~Text.

\begin{table*}[!ht]
    \centering
    \scriptsize
    \addtolength{\tabcolsep}{-0.1em}
    \begin{tabular}{lcccccccccccc}
        \hline
          & \multicolumn{2}{c}{ACRE$^T$} & \multicolumn{2}{c}{-Comp} & \multicolumn{2}{c}{-Sys} & \multicolumn{2}{c}{RAVEN$^T$} & \multicolumn{2}{c}{-Four} & \multicolumn{2}{c}{-In-Center} \\
        \cline{2-13}
         & Text & Symb & Text & Symb & Text & Symb & Text & Symb & Text & Symb & Text & Symb \\
        \hline
        ICLM$_{ACRE}$* (Table \ref{tab:abstract_reasoning_specific_results}) & 0.653 & 0.950 & 0.663 & 0.931 & 0.634 & 0.901 & - & - & - & - & - & - \\
        ICLM$_{RAVEN}$* (Table \ref{tab:abstract_reasoning_specific_results}) & - & - & - & - & - & - & 1.000 & 0.980 & 0.703 & 0.703 & 0.515 & 0.228 \\
        \hline
        ${\scriptstyle ACRE \rightarrow RAVEN}$ \\
        ICLM* (ours) & \textit{0.089} & \textbf{0.901} & 0.119 & \textbf{0.931} & 0.050 & \textbf{0.871} & \textbf{1.000} & \textbf{0.990} & \textbf{0.772} & \textbf{0.772} & \textbf{0.833} & \textbf{0.248} \\
        ICLM-Invariant* (ours) & \textbf{0.287} & \textit{0.396} & \textbf{0.277} & \textit{0.416} & \textbf{0.238} & \textit{0.455} & \textbf{1.000} & \textit{0.970} & \textit{0.673} & \textit{0.723} & \textit{0.723} & \textit{0.238} \\
        LLaMA2-Finetuned-Sequential* & 0.079 & 0.376 & \textit{0.149} & 0.386 & \textit{0.089} & 0.426 & \textit{0.980} & 0.772 & 0.634 & 0.554 & 0.584 & 0.069 \\
        \hline
    \end{tabular}
    \caption{Accuracy on ACRE and RAVEN when the model is trained sequentially.
    ICLM can use the information from ACRE to improve its performance on RAVEN, outperforming the baseline trained on RAVEN only, while preserving more knowledge from the previous task than the base LLaMA2-7B finetuned sequentially. In particular, the accuracy of ACRE-Symbolic sets is almost untouched.  }
    \label{tab:continual_learning_results}
\end{table*}

%% file: experiments/evol_corr.tex
\subsection{Evolution of Module Independence}

\begin{figure}[ht]
    \centering
    \begin{subfigure}[t]{0.48\linewidth}
        \includegraphics[width=\linewidth]{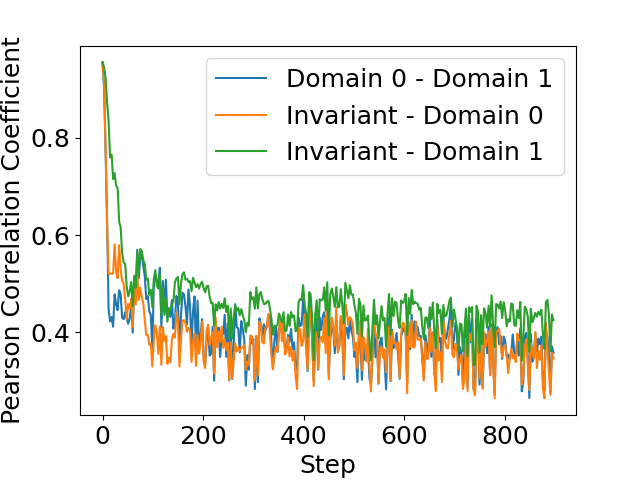}
        \caption{ACRE Correlation. }
    \end{subfigure}
    \hfill
    \begin{subfigure}[t]{0.48\linewidth}
        \includegraphics[width=\linewidth]{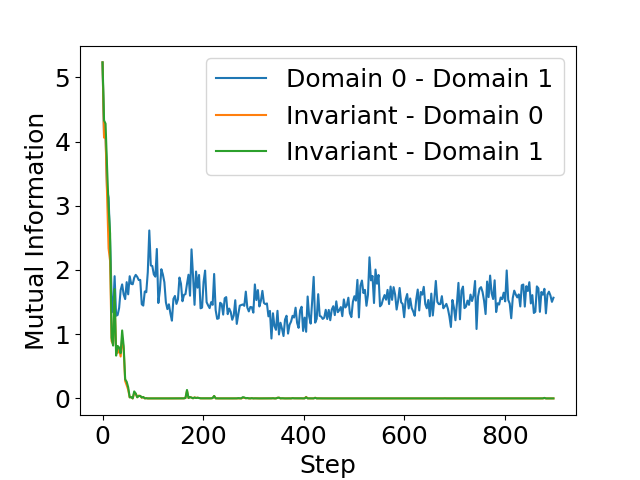}
        \caption{ACRE MI. }
    \end{subfigure}
    \hfill
    \begin{subfigure}[t]{0.48\linewidth}
        \includegraphics[width=\linewidth]{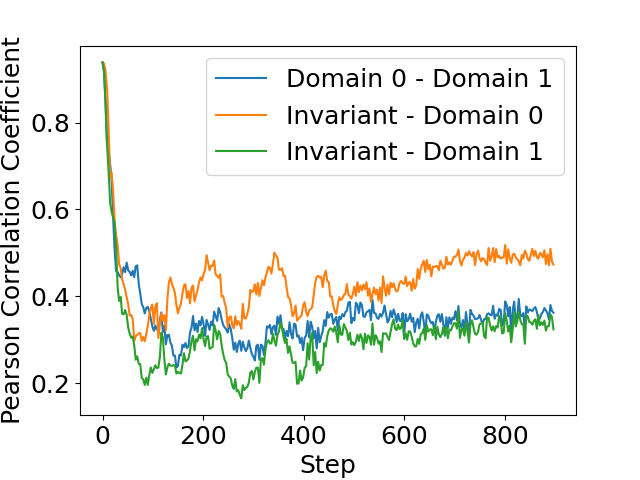}
        \caption{RAVEN Correlation. }
    \end{subfigure}
    \hfill
    \begin{subfigure}[t]{0.48\linewidth}
        \includegraphics[width=\linewidth]{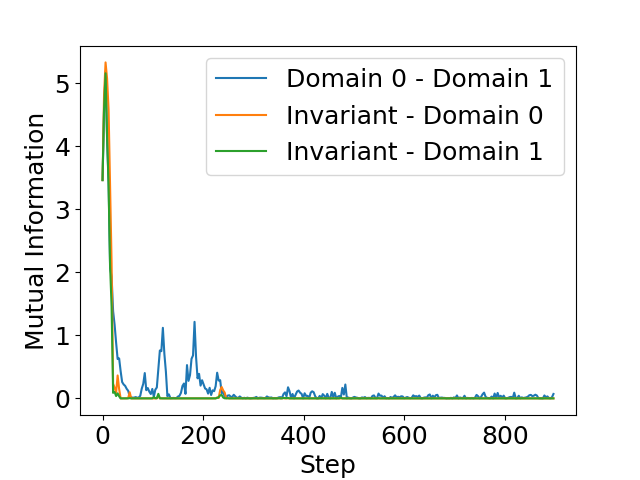}
        \caption{RAVEN MI. }
    \end{subfigure}
    \caption{Evolution of independence measures between modules during fine-tuning on ACRE and RAVEN. We measure independence on the last hidden states of the modules. Correlation and MI are highly reduced but modules remain correlated. }
    \label{fig:module_independence}
\end{figure}

We study the independence of the module hidden states during training (Figure \ref{fig:module_independence}) and inference (Figure \ref{fig:test-correlations}). we focus on two complementary measures: Mutual Information and the Pearson Correlation Coefficient that measures linear correlation between variables. The latter is limited to linear dependence but is more easily interpretable.
The shared Mutual Information as well as the Pearson Correlation Coefficient between modules are effectively reduced by the regularisation scheme during fine-tuning. However, the module hidden states remain correlated, in particular at inference time. Further investigation in Appendix \ref{sec:supp_corr} further shows that this correlation is maintained across most layers, which indicates the presence of a general domain-invariant mechanism shared by all modules and composing the basis of their reasoning abilities. Its influence is reduced via the fine-tuning procedure that develops domain-specific knowledge but it remains the main mechanism used, particularly at test time. 

\newcommand{\heatmap}[1]{
        \begin{tikzpicture}
            \begin{axis}[
                colormap/viridis,
                yticklabels={, Dom 1, Dom 0, Inv},
                xtick=\empty,
                enlargelimits=false,
                point meta min=0.5,
                point meta max=1.0,
                width=\linewidth,
                every tick label/.append style={font=\scriptsize}
            ]
                \addplot [
                    matrix plot*,
                    point meta=explicit,
                    nodes near coords,
                    every node near coord/.append style={
                        font=\scriptsize,
                        text=black,
                        anchor=center,
                    },
                    ] file {#1};
            \end{axis}
        \end{tikzpicture}
}

\begin{figure}[ht]
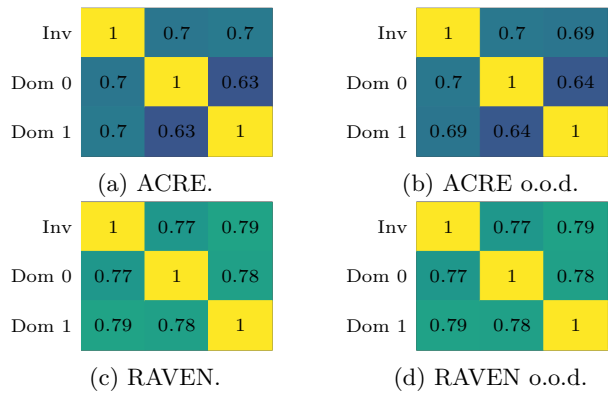

    \centering
    \begin{subfigure}{0.48\linewidth}
        \heatmap{acre.dat}
        \caption{ACRE. }
    \end{subfigure}
    \hfill
    \begin{subfigure}{0.48\linewidth}
        \heatmap{acre_ood.dat}
        \caption{ACRE o.o.d. }
    \end{subfigure}
    \hfill
    \begin{subfigure}{0.48\linewidth}
        \heatmap{raven.dat}
        \caption{RAVEN. }
    \end{subfigure}
    \hfill
    \begin{subfigure}{0.48\linewidth}
        \heatmap{raven_ood.dat}
        \caption{RAVEN o.o.d. }
    \end{subfigure}
    \caption{Correlation between the last hidden states of the modules during inference at test time. Module states are more correlated than during training. }
    \label{fig:test-correlations}
\end{figure}

%% file: experiments/routing_visualisation.tex
\subsection{Routing Alignment}
\label{sec:routing_alignment}

\begin{figure}[ht!]
    \centering
    \begin{subfigure}[t]{0.48\linewidth}
        \fbox{\includegraphics[width=0.9\linewidth,height=0.55\linewidth,trim={2.95cm 2.3cm 2.4cm 2.6cm},clip]{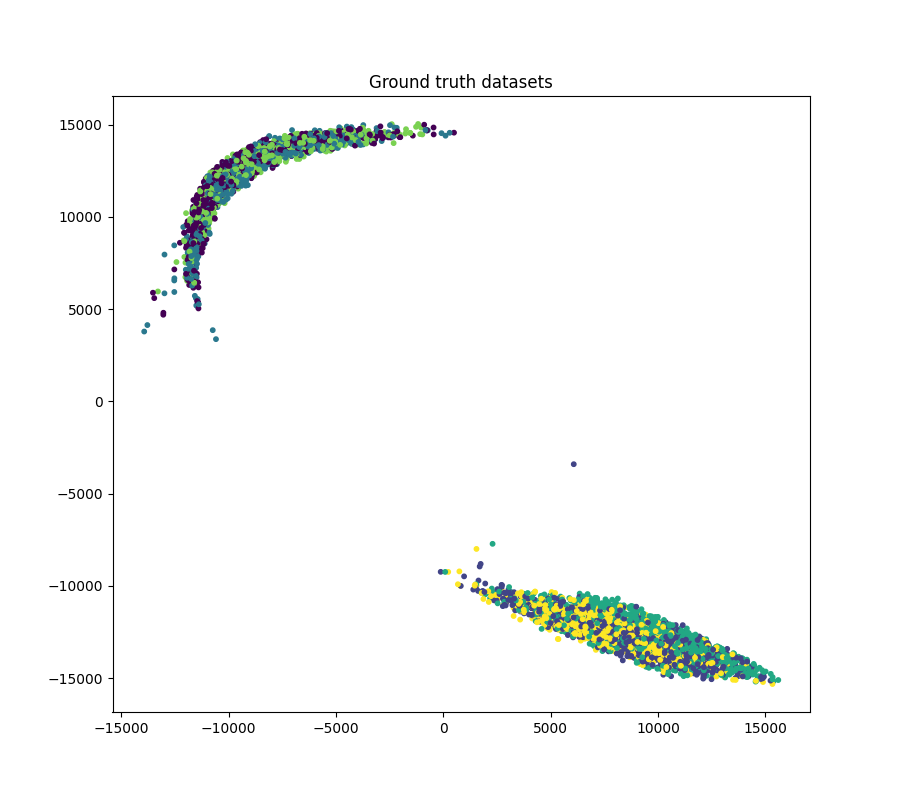}}
        \caption{ACRE Ground Truth}
    \end{subfigure}
    \begin{subfigure}[t]{0.48\linewidth}
        \fbox{\includegraphics[width=0.9\linewidth,height=0.55\linewidth,trim={2.95cm 2.3cm 2.4cm 2.6cm},clip]{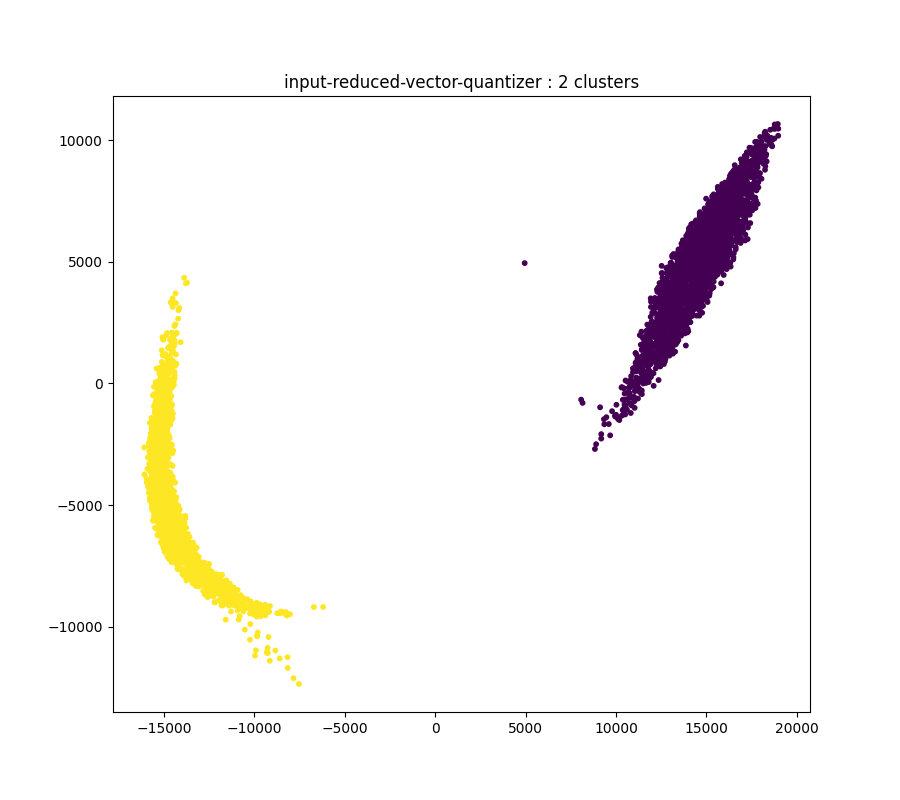}}
        \caption{Vector quantisation router}
    \end{subfigure}
    \hfill
    \begin{subfigure}[t]{0.48\linewidth}
        \fbox{\includegraphics[width=0.9\linewidth,height=0.55\linewidth,trim={2.95cm 2.3cm 2.4cm 2.6cm},clip]{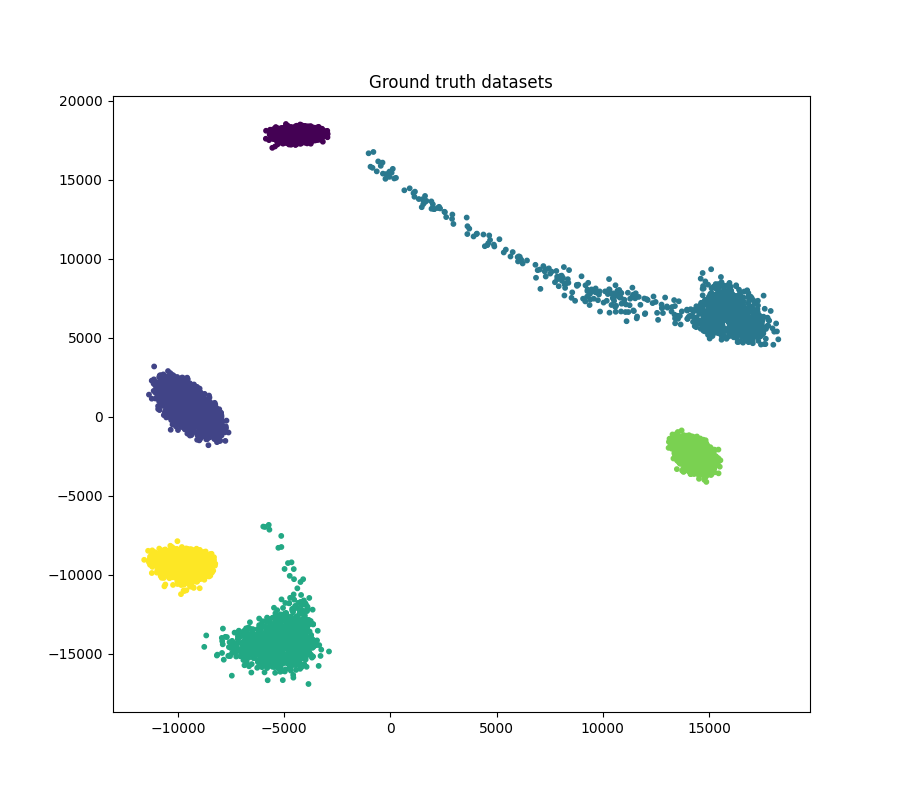}}
        \caption{RAVEN Ground Truth}
    \end{subfigure}
    \begin{subfigure}[t]{0.48\linewidth}
        \fbox{\includegraphics[width=0.9\linewidth,height=0.55\linewidth,trim={2.95cm 2.3cm 2.4cm 2.6cm},clip]{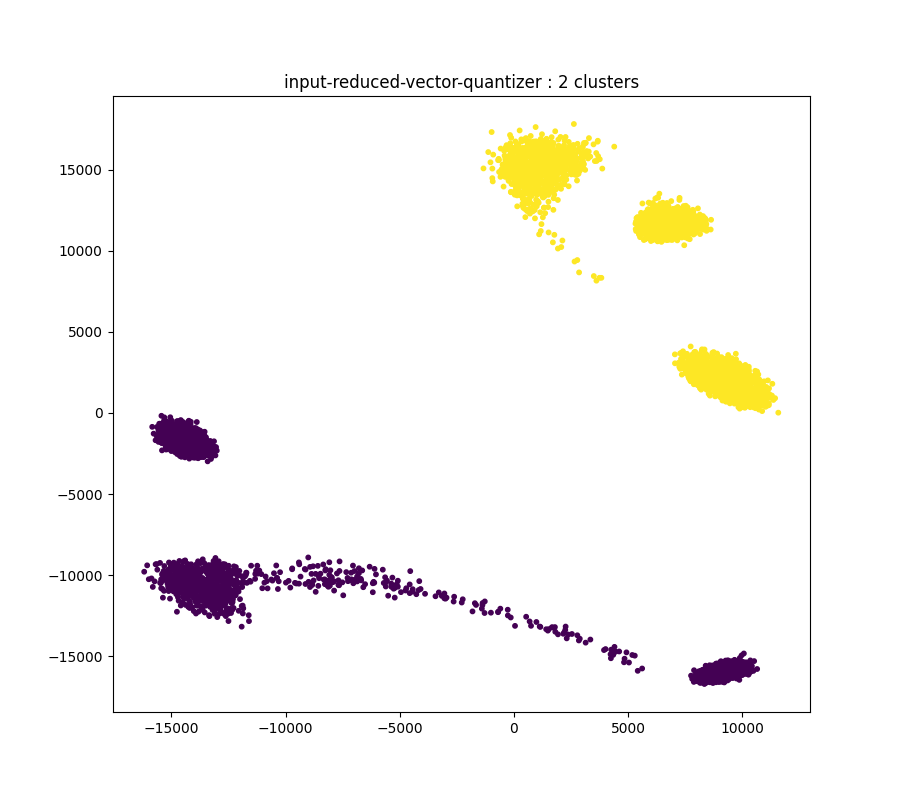}}
        \caption{Vector quantisation router}
    \end{subfigure}
    \caption{2D projection of the hidden states of LLaMA2 on ACRE and RAVEN i.i.d and o.o.d sets. Ground truth samples are labelled as in Table \ref{tab:abstract_reasoning_specific_results} (text/symbolic i.i.d/o.o.d sets). Text and symbolic inputs are always clustered separately. i.i.d and o.o.d sets are clustered together in ACRE and separated in RAVEN. The router follows the text and symbolic division. }
    \label{fig:ood_clusters}
    \vspace{-0.5cm}
\end{figure}

We look deeper at the embedding space in the routing module, projected into a 2D space using Multidimensional Scaling (MDS) \citep{borg2005modern}. Figure \ref{fig:ood_clusters} shows the embedding spaces and their attribution to the domain-specific modules. Detailed attributions are shown in Appendix \ref{sec:supp_routing_alignment}.
Both datasets have a clear division between text and symbolic embeddings. However, the o.o.d sets are not well separated in ACRE while they are in RAVEN. This division can explain the similarity in the results between the ACRE i.i.d and o.o.d sets, as shown in Table \ref{tab:abstract_reasoning_specific_results}. Moreover, as the distributions are very similar, the impact of the router and the need for abstraction are reduced. On the other hand, there is a clear separation between the i.i.d and o.o.d RAVEN embeddings, explaining the differences in behaviours from the models across the sets. Adding more modules could allow taking more advantage of this separation, with each module specialising to a subdomain closer to one of the o.o.d embeddings.

%% file: conclusion.tex
\section{Conclusion}

Performing strong out-of-distribution reasoning is a challenging task, and despite their impressive performance on a wide range of problems, LLMs have not demonstrated this ability yet. Combining this popular model with causal models could help bridge this gap. This work presents a modular architecture yielding LLMs to behave as Independent Causal Mechanisms. 
We show theoretically that the proposed model generates causally-independent modules. We perform experiments on abstract and causal reasoning tasks in o.o.d and continual learning settings and show that these principles increase strong reasoning and generalisation.
We further show that the proposed modules specialise to their domain with fine-tuning but still partially rely on a shared domain-invariant mechanism, highlighting a limitation for representing ICMs with LLMs.

%% file: limitations.tex
\section*{Limitations}
\label{sec:limitations}

The model proposed in this paper is constrained to work in a modular manner. All modules are sparsely connected at the level of the language modelling head. This single connection offers a useful initial framework to study the ICMs within the context of LLMs and can represent a wide range of problems (requiring the composition of several independent reasoning processes) but it can only represent causal DAGs with a single layer depth, potentially hindering the expressivity of more complex mechanism  interactions. Generating complex causal computation graphs tailored to the task at hand may improve performance, but this problem is out of the scope of this paper.

We also focus our investigation on the independence and accuracy of the modules and do not attempt to directly represent the true causal mechanisms of the tasks as they are unknown. Moreover, we conduct experiments on reasoning tasks to verify if inducing high-level modularity can yield increases in performance and generalisation. We aim not to outperform the state-of-the-art on the problems but to study whether the proposed mechanisms can yield such increases. 

In addition, training and fine-tuning Large Language Models has a high computational cost. Due to this high cost, we perform a single fine-tuning run per task and conduct experiments on this model.

%% file: appendix.tex
\input{theory_full}

\input{supp_routing_strategy}

\input{supp_aggregation}

\section{Additional Experiments}

\input{experiments/mi_variation}

\input{experiments/supp_routing_visualisation}

\input{experiments/cluster_alternatives}

\input{experiments/aggregation_schemes}

\input{experiments/supp_corr}

%% file: theory_full.tex
\section{Supplement to the Theoretical Perspective}
\label{sec:theory_full}

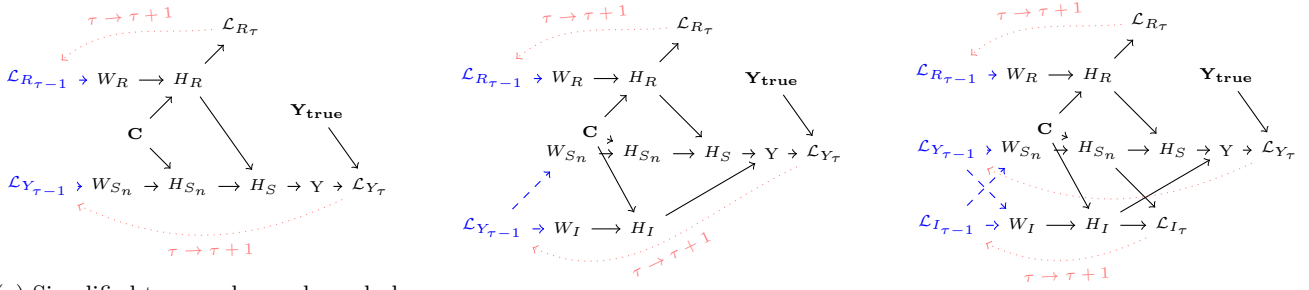
\begin{figure}
    \centering
    
    \begin{subfigure}{0.32\linewidth}
    \begin{tikzpicture}[node distance=1.0cm]
    \tikzstyle{every node}=[font=\tiny]

        \node (c) {\textbf{C}};
        \node (sn) [below right of=c] {$H_{S_n}$};
        \node (s) [right of=sn] {$H_{S}$};
        \node (r) [above right of=c] {$H_R$};
        \node (y) [right of=s, node distance=0.71cm] {Y};
        \node (yt) [above of=y] {$\mathbf{Y_{true}}$};
        \node (l) [right of=y, node distance=0.71cm] {$\mathcal{L}_{Y_{\tau}}$};
        \node (lr) [above right of=r] {$\mathcal{L}_{R_{\tau}}$};
        \node (wn) [left of=sn] {$W_{S_n}$};
        \node (wr) [left of=r] {$W_R$};
        \node (lp) [left of=wn] {\textcolor{blue}{$\mathcal{L}_{Y_{\tau-1}}$}};
        \node (lrp) [left of=wr] {\textcolor{blue}{$\mathcal{L}_{R_{\tau-1}}$}};
        
        \draw[->] (c) -- (r);
        \draw[->] (c) -- (sn);
        \draw[->] (r) -- (s);
        \draw[->] (sn) -- (s);
        \draw[->] (s) -- (y);
        \draw[->] (y) -- (l);
        \draw[->] (yt) -- (l);
        \draw[->] (r) -- (lr);
        \draw[->] (wn) -- (sn);
        \draw[->] (wr) -- (r);
        \draw[->,blue,dashed] (lp) -- (wn);
        \draw[->,blue,dashed] (lrp) -- (wr);
        \draw[->,dotted,red!50] (lr) to[out=185,in=40] node[midway, above, sloped] {$\tau \rightarrow \tau+1$} (lrp);
        \draw[->,dotted,red!50] (l) to[out=205,in=335] node[midway, below, sloped] {$\tau \rightarrow \tau+1$} (lp);
        
    \end{tikzpicture}
    \caption{Simplified temporal causal graph during training with no domain-invariant module.}
    \end{subfigure}
    \hfill
    \begin{subfigure}{0.32\linewidth}
    \begin{tikzpicture}[node distance=1.0cm]
    \tikzstyle{every node}=[font=\tiny]

        \node (sn){$H_{S_n}$};
        \node (s) [right of=sn] {$H_{S}$};
        \node (r) [above of=sn] {$H_R$};
        \node (c) [below left of=r] {\textbf{C}};
        \node (i) [below of=sn] {$H_I$};
        \node (y) [right of=s, node distance=0.71cm] {Y};
        \node (yt) [above of=y] {$\mathbf{Y_{true}}$};
        \node (l) [right of=y, node distance=0.71cm] {$\mathcal{L}_{Y_{\tau}}$};
        \node (lr) [above right of=r] {$\mathcal{L}_{R_{\tau}}$};
        \node (wn) [left of=sn] {$W_{S_n}$};
        \node (wr) [left of=r] {$W_R$};
        \node (wi) [left of=i] {$W_I$};
        \node (lp) [left of=wi] {\textcolor{blue}{$\mathcal{L}_{Y_{\tau-1}}$}};
        \node (lrp) [left of=wr] {\textcolor{blue}{$\mathcal{L}_{R_{\tau-1}}$}};
        
        \draw[->] (c) -- (r);
        \draw[->] (c) -- (sn);
        \draw[->] (c) -- (i);
        \draw[->] (r) -- (s);
        \draw[->] (sn) -- (s);
        \draw[->] (s) -- (y);
        \draw[->] (i) -- (y);
        \draw[->] (y) -- (l);
        \draw[->] (yt) -- (l);
        \draw[->] (r) -- (lr);
        \draw[->] (wn) -- (sn);
        \draw[->] (wr) -- (r);
        \draw[->] (wi) -- (i);
        \draw[->,blue,dashed] (lp) -- (wn);
        \draw[->,blue,dashed] (lp) -- (wi);
        \draw[->,blue,dashed] (lrp) -- (wr);
        \draw[->,dotted,red!50] (lr) to[out=185,in=40] node[midway, above, sloped] {$\tau \rightarrow \tau+1$} (lrp);
        \draw[->,dotted,red!50] (l) to[out=205,in=335] node[midway, below, sloped] {$\tau \rightarrow \tau+1$} (lp);
        
    \end{tikzpicture}
    \caption{Simplified temporal causal graph $\mathcal{G}$. The graph is the same as Figure \ref{fig:simplified_training_causal_grah_forward}.}
    \end{subfigure}
    \hfill
    \begin{subfigure}{0.32\linewidth}
    \begin{tikzpicture}[node distance=1.0cm]
    \tikzstyle{every node}=[font=\tiny]

        \node (sn){$H_{S_n}$};
        \node (s) [right of=sn] {$H_{S}$};
        \node (r) [above of=sn] {$H_R$};
        \node (c) [below left of=r] {\textbf{C}};
        \node (i) [below of=sn] {$H_I$};
        \node (y) [right of=s, node distance=0.71cm] {Y};
        \node (yt) [above of=y] {$\mathbf{Y_{true}}$};
        \node (l) [right of=y, node distance=0.71cm] {$\mathcal{L}_{Y_{\tau}}$};
        \node (lr) [above right of=r] {$\mathcal{L}_{R_{\tau}}$};
        \node (li) [right of=i] {$\mathcal{L}_{I_{\tau}}$};
        \node (wn) [left of=sn] {$W_{S_n}$};
        \node (wr) [left of=r] {$W_R$};
        \node (wi) [left of=i] {$W_I$};
        \node (lp) [left of=wn] {\textcolor{blue}{$\mathcal{L}_{Y_{\tau-1}}$}};
        \node (lrp) [left of=wr] {\textcolor{blue}{$\mathcal{L}_{R_{\tau-1}}$}};
        \node (lip) [left of=wi] {\textcolor{blue}{$\mathcal{L}_{I_{\tau-1}}$}};
        
        \draw[->] (c) -- (r);
        \draw[->] (c) -- (sn);
        \draw[->] (c) -- (i);
        \draw[->] (r) -- (s);
        \draw[->] (sn) -- (s);
        \draw[->] (s) -- (y);
        \draw[->] (i) -- (y);
        \draw[->] (y) -- (l);
        \draw[->] (yt) -- (l);
        \draw[->] (r) -- (lr);
        \draw[->] (sn) -- (li);
        \draw[->] (i) -- (li);
        \draw[->] (wn) -- (sn);
        \draw[->] (wr) -- (r);
        \draw[->] (wi) -- (i);
        \draw[->,blue,dashed] (lp) -- (wn);
        \draw[->,blue,dashed] (lp) -- (wi);
        \draw[->,blue,dashed] (lrp) -- (wr);
        \draw[->,blue,dashed] (lip) -- (wn);
        \draw[->,blue,dashed] (lip) -- (wi);
        \draw[->,dotted,red!50] (lr) to[out=185,in=40] node[midway, above, sloped] {$\tau \rightarrow \tau+1$} (lrp);
        \draw[->,dotted,red!50] (l) to[out=205,in=335] (lp);
        \draw[->,dotted,red!50] (li) to[out=205,in=335] node[midway, below, sloped] {$\tau \rightarrow \tau+1$} (lip);
        
    \end{tikzpicture}
    \caption{Temporal causal graph with Information Minimisation loss added.}
    \end{subfigure}
    \caption{Causal graphs with and without domain-invariant module and Mutual Information minimisation loss. C is the input context. $H_R$, $H_I$, $H_{S_n}$, $H_S$ are the latent states of the router, domain-invariant, domain-specific and activated domain-specific (after router weighting) modules. For simplicity, we only show the state $H_{S_n}$ of the activated domain-specific module $n$. $Y$ and $Y_{true}$ are the output and true distributions. $W_R$, $W_{S_n}$ and $W_I$ are the trainable parameters of the modules. $\mathcal{L}_Y = \mathcal{L}_o + \alpha \cdot \mathcal{L}_{inv} + \beta \cdot \mathcal{L}_{dom}$ and $\mathcal{L}_R$ are the output and router losses. Black edges show the forward pass at step $\tau$. \textcolor{blue}{Blue} dashed edges show the backward pass at step $\tau$. \textcolor{red}{Red} dotted edges illustrate the causal links between the forward and backward passes. For simplicity, we only show the step for the loss variables as they appear twice. All other variables are at step $\tau$. }
    \label{fig:simplified_training_causal_grah_forward_2}
\end{figure}

In this section, we prove the assertions made in Section \ref{sec:theory} of the main paper. We study under what conditions the following equations (repeated from Section \ref{sec:theory}) hold:

\begin{align}
    &P(H_R|\docalc{H_{S_n}}) = P(H_R)~\forall n \in [1,N]
    \label{eq:indep_router_dom_2} \\
    &P(H_R|\docalc{H_I}) = P(H_R)
    \label{eq:indep_router_inv_2} \\
    &\begin{aligned}
    &P(H_{S_n}|\docalc{H_{S_{\hat{n}}}}) = P(H_{S_n}) \\
    &~~~~~~\forall \hat{n} \in [1,N] \setminus \{n\} ~\forall n \in [1,N]
    \end{aligned}
    \label{eq:indep_doms_2} \\
    &P(H_I|\docalc{H_{S_n}})=P(H_I)~\forall n \in [1,N]
    \label{eq:indep_inv_dom_2}
\end{align}

$H_R$, $H_I$ and $H_{S_n}~\forall n \in [1,N]$ are the respective representations generated by the router, domain-invariant and $N$ domain-specific modules.

The rules of do-calculus, defined in \citet{pearl1995causal}, allow one to reduce interventional queries (with the $\docalc{\cdot}$ operator) to observational queries. We will only use the \textit{deletion of actions} rule. A simplified rule is shown in Equation \ref{eq:do_rule3}: 

\begin{align}
    \begin{aligned}
        &P(Y|\docalc{X}) = P(Y) \\&
        \text{~~~~~if~} (Y \independent X)_{\mathcal{G}_{\overline{X}}}
    \end{aligned}
\label{eq:do_rule3}
\end{align}

\noindent $\mathcal{G}_{\overline{X}}$ represents the causal graph $\mathcal{G}$ with the incoming edges of $X$ removed.

Let us first address the causal relationships of the router. Equations \ref{eq:indep_router_inv_2} and \ref{eq:indep_router_dom_2} can be verified using the simplified causal graph in Figure \ref{fig:simplified_training_causal_grah_forward_2}. 
They are a direct application of rule \ref{eq:do_rule3}. When removing the parents of $H_{S_n}$ or $H_I$, $H_R$ is \textbf{d-separated} \citep{pearl1988probabilistic} from them: the backward path through $C$ is blocked and the forward path through $W_R$ is not connected to $H_{S_n}$ or $H_I$. This is due to the use of a separate loss function for training the router when using the vector quantisation routing strategy.
One could notice that we do not represent the sum of losses of Equation \ref{eq:losses}. We omit it in the simplified graph. Its impact on the backward pass is incidental since each element can be optimised independently.

Let us now address the causal relationships of one activated domain-specific module $n$ with its counterparts (Equation \ref{eq:indep_doms_2}). Again, under graph $\mathcal{G}_{\overline{H_{S_{\hat{n}}}}}$, $H_{S_n}$, the backward path through $C$ between is blocked. In addition, we make the assumption that only the module $n$ is activated and is connected to $Y$ as in Figure \ref{fig:simplified_training_causal_grah_forward_2}. This assumption is verified when using the vector quantisation routing strategy. As a consequence, the routing process $H_S$ can be decomposed into multiple subgraphs $H_{S_n} \xrightarrow{H_R} Y$ and $H_{S_{\overline{n}}} \not\xrightarrow{H_R} Y~\forall \overline{n} \in [1,N] \setminus \{n\}$, with $A \xrightarrow{H_R} B$ equivalent to having $A \rightarrow X \leftarrow H_R$ and $X \rightarrow B$. $A \not\xrightarrow{H_R} B$ removes the second link. Therefore, during the backward pass, there is only one link $\mathcal{L}_Y \rightarrow W_{S_n} \rightarrow H_{S_n}$ and no path to the other domain-specific modules $\overline{n}$. 
A last type of path can exist; here is an example: assuming a model with two domain-specific modules, $S_0$ and $S_1$, activated one after the other, the following path exists: $H_{S_1} \xrightarrow{H_R} Y_\tau \rightarrow \mathcal{L}_{Y_\tau} \rightarrow W_I \xrightarrow{C} H_I \rightarrow Y_{\tau+1} \rightarrow \mathcal{L}_{Y_{\tau+1}} \rightarrow W_{S_2} \xrightarrow{C} H_{S_2}$. There is a causal path forward path from $H_{S_1}$ to $H_{S_2}$. The path does not exist if there is no invariant module, and Equation \ref{eq:indep_doms_2} holds.

If an invariant module is part of the model, Equations \ref{eq:indep_doms_2} and \ref{eq:indep_inv_dom_2} do not hold because of the path above: $H_I$ and $H_{S_n}$ are not independent in the causal graph $\mathcal{G}_{H_{S_n}}$. Independence is achieved by minimising the Mutual Information between $H_I$ and $H_{S_n}$, as discussed in the main paper.

%% file: supp_routing_strategy.tex
\section{Additional Routing Strategies}
\label{sec:supp_routing_strategy}

In our main experiments, we use a simple routing strategy based on computing vector quantisation from Euclidean distance. In this section, we consider several additional routing strategies:
\begin{itemize}
    \item K-Means Clustering
    \item Euclidean Distance Weighting
\end{itemize}

\paragraph{K-Means Clustering} This strategy computes the clusters using K-Means (Lloyd's algorithm) \citep{lloyd1982least}. We first learn the cluster centroids on the training set separately. Then, we fine-tune the other modules. Therefore, the clustering mechanism is independent of the gradient descent during fine-tuning, and the quality of the clusters with respect to the data distribution depends solely on the robustness of the clustering method.
For efficiency reasons, we do not directly perform the clustering on the hidden states of the router module. Before clustering an input embedding, we project it to a more dense space with fewer dimensions (typically 64). We apply Multidimensional Scaling with the SMACOF algorithm \citep{borg2005modern}. The algorithm requires us to provide a base of the input space to perform the projection. We span the space using a random set of vectors from the training space. Because the distribution is skewed, we do not have a warranty to build a base. To remain computation-efficient, we sample $8 \times M$ vectors with $M$ the dimensionality of the reduced space.

\paragraph{Euclidean Distance Weighting} This strategy differs from the other ones as it does not use vector quantisation. Instead, we compute the Euclidean distance between the embeddings and the centroid coordinates (randomly initialised) and use softmin to convert the distances into continuous weights between zero and one. The lower the distance between the embedding and a centroid, the higher the weight the corresponding domain-specific module will have on the output. Consequently, with this method, all domain-specific modules are always activated. This operation is differentiable and is the closest to the routing process of Mixture-of-Expert models like the Switch Transformer \citep{DBLP:journals/jmlr/FedusZS22}. This method does not follow the causal structure discussed in Section \ref{sec:theory}. Instead, it uses the output loss to update the centroid coordinates.

The results obtained with these two routing strategies are provided in Appendix \ref{sec:routing_variations}.

%% file: supp_aggregation.tex
\section{Additional Aggregation Schemes}
\label{sec:supp_aggregation}

In this section, we describe two additional aggregation schemes between the domain-invariant and domain-specific modules. Instead of using a shared language modelling head, we propose to use a separate head for each module and combine their outputs at the end by a weighted sum.
This method tackles the issue of prioritised modules (e.g. one module being overused at the expense of the others). However, the information from the modules is not linearly combined but added separately, reducing expressivity.

We want to bound the output of each model such that it influences the final prediction by a pre-determined factor (given by the router output for the domain-specific modules and provided as a hyperparameter for the domain-invariant module).
Each module outputs unbounded logits. The lack of bounds prevents them from directly multiplying the logits by their weighting factor and summing them together. Indeed, one module could overcome the weighting by increasing the magnitude of its logits. We consider two combination schemes: in the \textit{logit space} and in the \textit{probability space}.

\paragraph{Combination in the logit space}

The aggregation scheme in the logit space is very similar to the one performed in the latent space in the main paper. We first perform a shared batch normalisation \citep{DBLP:conf/icml/IoffeS15} between the modules to overcome the unbounded issue in the logit space. For a batch of size $|\mathcal{B}|$, one domain-specific active module and one domain-invariant module, batch normalisation is operated on $2 \times |\mathcal{B}|$ samples.
We attribute a weight $w_I$ to the domain-invariant module as a hyperparameter and $w_{S_n} = r_n \cdot (1 - w_I)$ to the domain-specific module, with $r_n$ the weight given by the router. After normalisation, We multiply each logit value by its corresponding weight and sum them together.

\paragraph{Combination in the probability space}

Each module outputs unbounded logits, so we first convert each output into normalised probabilities (that sum to one). We then perform the weighting in each probability space before converting them back to logits (shown in Equation \ref{eq:weighted_logits}). Finally, the outputs from all modules are summed together (shown in Equation \ref{eq:final_logits}). The final probabilities are shown in Equation \ref{eq:final_probas}. 

\begin{align}
    \widetilde{l_a}(Y|c)   &= \log\Big(\dfrac{1-w_a}{2} + w_a \cdot P_a(Y|c))\Big) + \log(B_a) 
    \label{eq:weighted_logits}
    \\
    l(Y|c)  &= \widetilde{l_I}(Y|c) + \sum\limits_{n \in [1,N]} \widetilde{l_{S_n}}(Y|c) 
    \label{eq:final_logits}
    \\
    P(Y|c)  &= \sigma(l(Y|c))
    \label{eq:final_probas}
\end{align}

$c$ is the input context. $P(Y|c)$ is the final output distribution between all words $Y$, obtained using softmax normalisation $\sigma$ on the output logits $l(Y|c)$. The output logits are obtained by summing the weighted logits of the domain-invariant module $\widetilde{l_I}(Y|c)$ and all domain-specific modules $\widetilde{l_{S_n}}(Y|c)$. Equation \ref{eq:weighted_logits} shows the weighting process for all modules ($a \in \{I, S_1, \dots, S_N\}$). The weight $w_I$ is a hyperparameter set prior to training. The weights $w_{S_n} \forall~n \in [1,N]$ combine the weight $w_I$ with the router weights $r_n$: $w_{S_n} = r_n \cdot (1 - w_I)$. 
$B_a$ is a normalisation term that ensures the conversion function between probabilities and logits is invertible.

The results obtained with these two aggregation schemes are provided in Appendix \ref{sec:aggregation_variations}.

%% file: experiments/mi_variation.tex
\subsection{Evolution of the Mutual Information Across Training}
\label{sec:mi_variation}

\begin{figure}
    \begin{minipage}[t]{0.48\linewidth}
    \centering
    \includegraphics[width=1.1\linewidth]{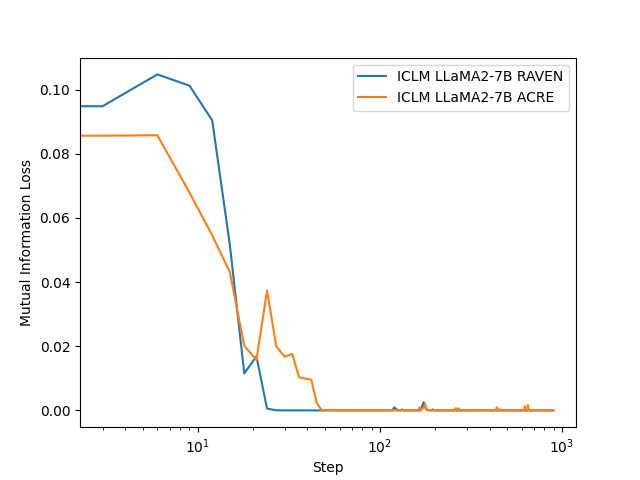}
    \caption{Evolution of the Mutual Information loss during training on ACRE and RAVEN. The x-axis corresponds to the number of training steps and is shown in the log scale. }
    \label{fig:mi_loss_evolution}
    \end{minipage}
    \hfill
    \begin{minipage}[t]{0.48\linewidth}
    \centering
    \includegraphics[width=1.1\linewidth]{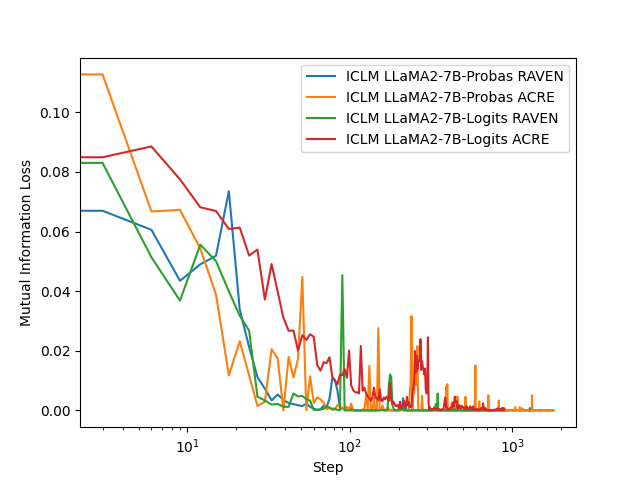}
    \caption{Evolution of the Mutual Information loss during training of variants with probability and logits aggregation on ACRE and RAVEN. The x-axis corresponds to the number of training steps and is shown in the log scale. }
    \label{fig:mi_loss_evolution_logits_probas}
    \end{minipage}
\end{figure}

To ensure the independence between the domain-specific and domain-invariant modules, we minimise the mutual Information between them. Figure \ref{fig:mi_loss_evolution} shows the evolution of Mutual Information during training. We observe that it quickly decreases to reach below $0,0001$. Figure \ref{fig:mi_loss_evolution_logits_probas} shows the same loss for the variants using aggregation in the logit and probability spaces. Unlike for the main model, we observe small spikes in the loss after $100$ training steps. The aggregation scheme that uses a shared language modelling head (our default) seems more stable during training.

%% file: experiments/supp_routing_visualisation.tex
\subsection{Routing Alignment and Visualisation}
\label{sec:supp_routing_alignment}

We study the module attribution performed by the router more deeply. Table \ref{tab:acre_cluster_alignment} shows the alignment between the two domain-specific modules. We first observe that the division is mainly syntactic: each module specialises towards one type of input format, either text or symbolic. It aligns perfectly with the dataset.

\begin{table}[ht]
    \centering
    \caption{Alignment between modules and formats in the ACRE dataset. Each column shows the proportion of activation for each module for a given dataset. Each module specialises perfectly to one dataset. }
    \begin{tabular}{lcccccc}
        \hline
          & \multicolumn{2}{c}{ACRE} & \multicolumn{2}{c}{-Comp} & \multicolumn{2}{c}{-Sys} \\
        \cline{2-7}
         & Text & Symb & Text & Symb & Text & Symb \\
        \hline
        $n=0$ & 0.0 & 1.0 & 0.0 & 1.0 & 0.0 & 1.0 \\
        $n=1$ & 1.0 & 0.0 & 1.0 & 0.0 & 1.0 & 0.0 \\
        \hline
    \end{tabular}
    \label{tab:acre_cluster_alignment}
\end{table}

\begin{figure}
    \centering
    \begin{subfigure}{0.22\textwidth}
        \fbox{\includegraphics[width=0.9\linewidth,height=0.55\linewidth,trim={2.95cm 2.3cm 2.4cm 2.6cm},clip]{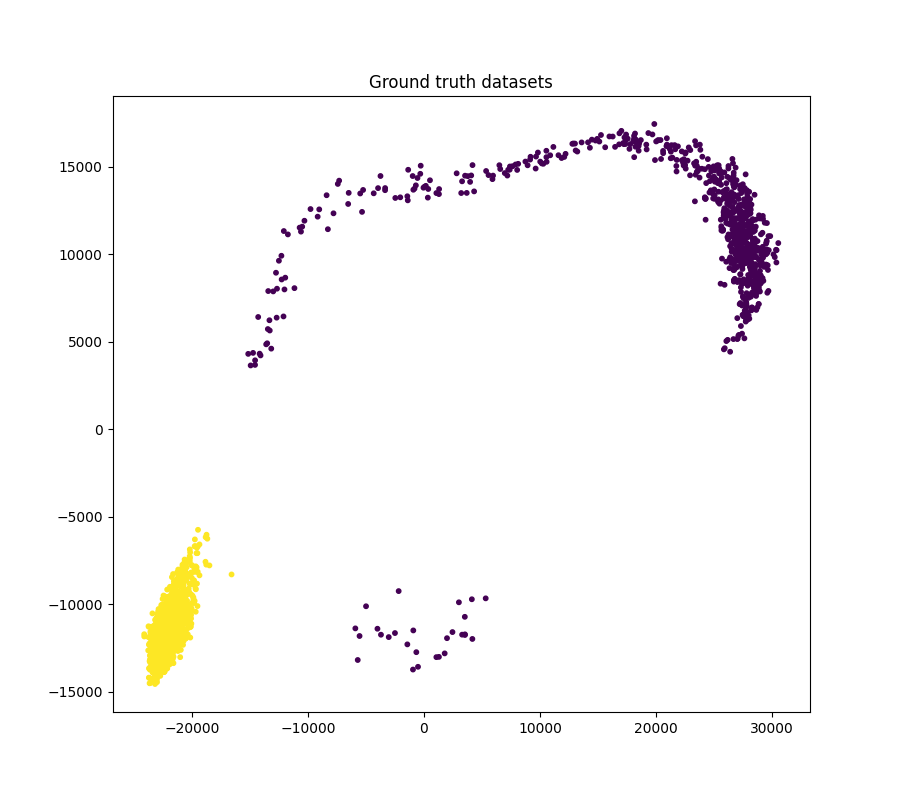}}
        \caption{Ground Truth ($l=-1$)}
    \end{subfigure}
    \hfill
    \begin{subfigure}{0.22\textwidth}
        \fbox{\includegraphics[width=0.9\linewidth,height=0.55\linewidth,trim={7.8cm 26.1cm 39.5cm 16.4cm},clip]{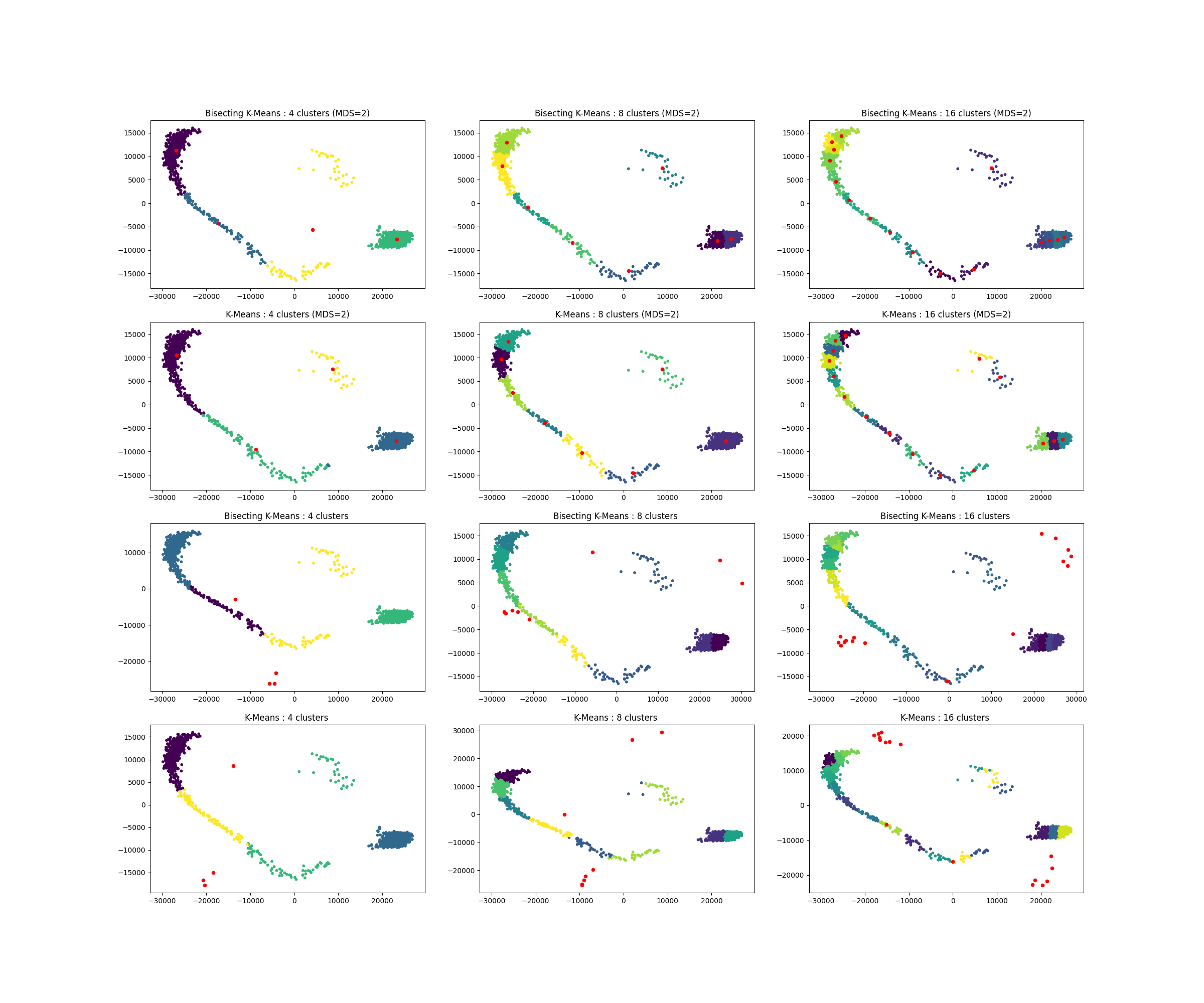}}
        \caption{K-Means ($l=-1$)}
    \end{subfigure}
    \hfill
    \begin{subfigure}{0.22\textwidth}
        \fbox{\includegraphics[width=0.9\linewidth,height=0.55\linewidth,trim={2.95cm 2.3cm 2.4cm 2.6cm},clip]{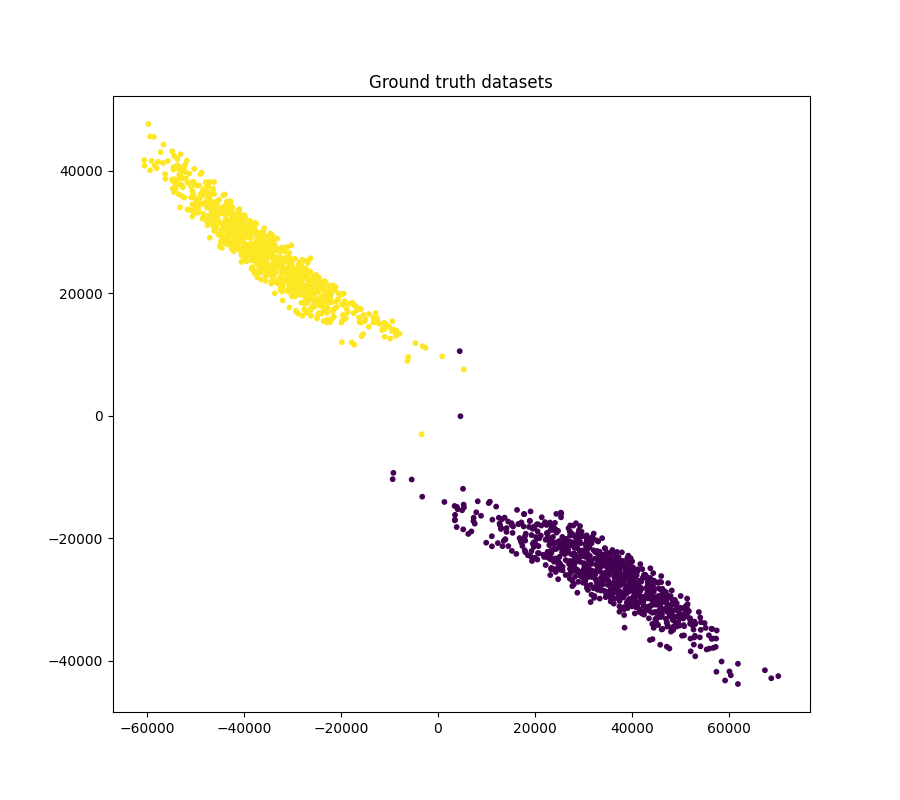}}
        \caption{Ground Truth ($l=-2$)}
    \end{subfigure}
    \hfill
    \begin{subfigure}{0.22\textwidth}
        \fbox{\includegraphics[width=0.9\linewidth,height=0.55\linewidth,trim={7.8cm 26.1cm 39.5cm 16.4cm},clip]{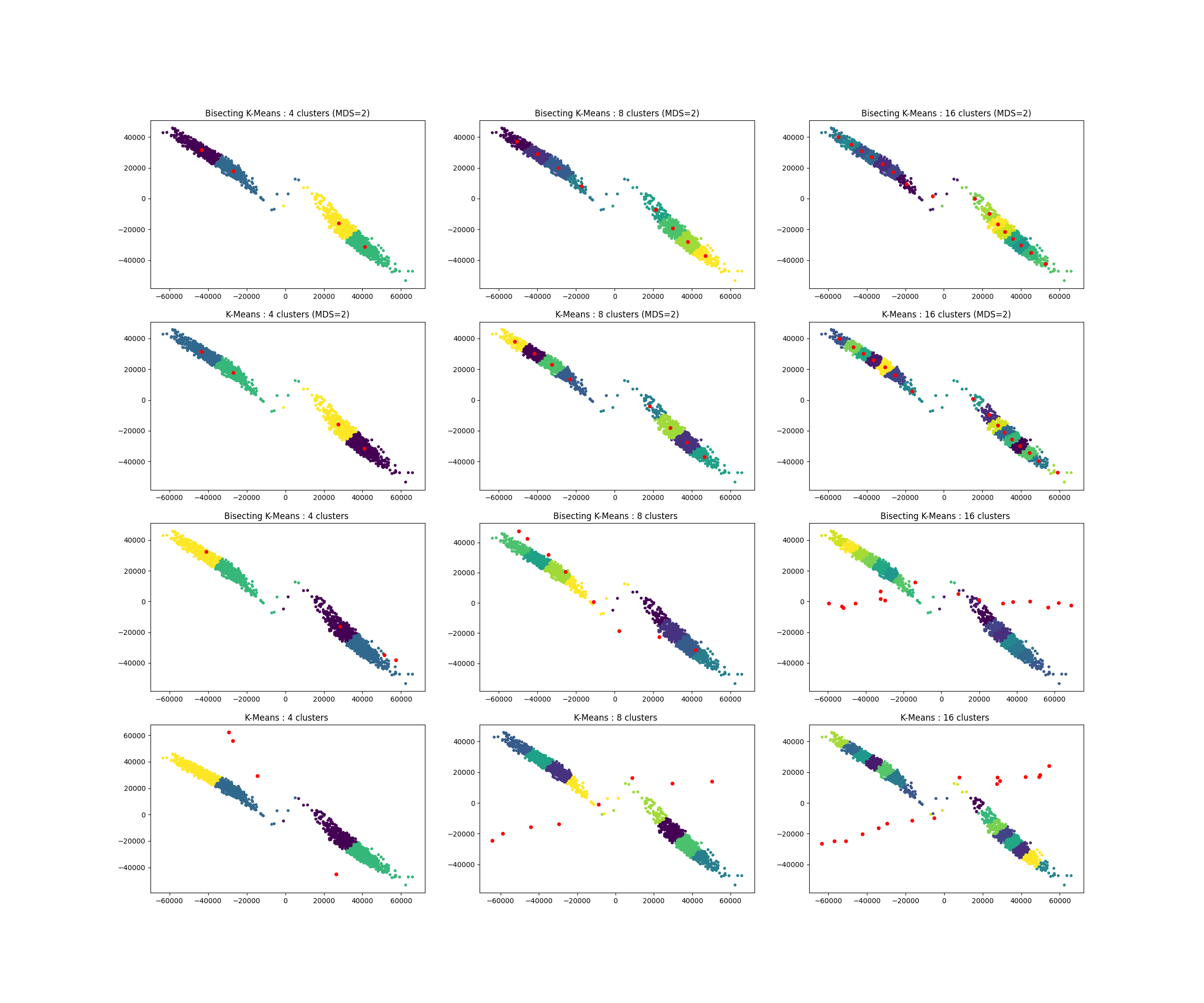}}
        \caption{K-Means ($l=-2$)}
    \end{subfigure}

    \caption{Clusters formed from the hidden states of LLaMA2 on the ACRE training sets. The visualisations contain the last two levels ($l$) of hidden layers. The ground truth shows the true splits (text/symbolic). The learned clusters use 4 centroids.  }
    \label{fig:supp_acre_clusters}
\end{figure}

\begin{figure}
    \centering
    \begin{subfigure}{0.22\textwidth}
        \fbox{\includegraphics[width=0.9\linewidth,height=0.55\linewidth,trim={2.95cm 2.3cm 2.4cm 2.6cm},clip]{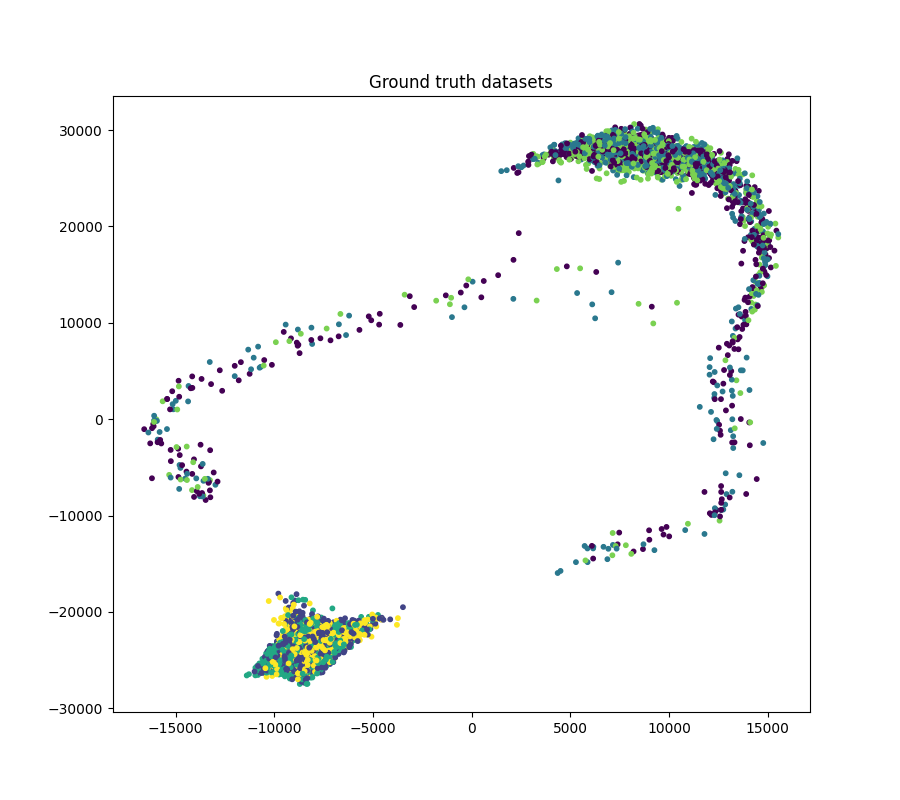}}
        \caption{Ground Truth ($l=-1$)}
    \end{subfigure}
    \hfill
    \begin{subfigure}{0.22\textwidth}
        \fbox{\includegraphics[width=0.9\linewidth,height=0.55\linewidth,trim={7.8cm 26.1cm 39.5cm 16.4cm},clip]{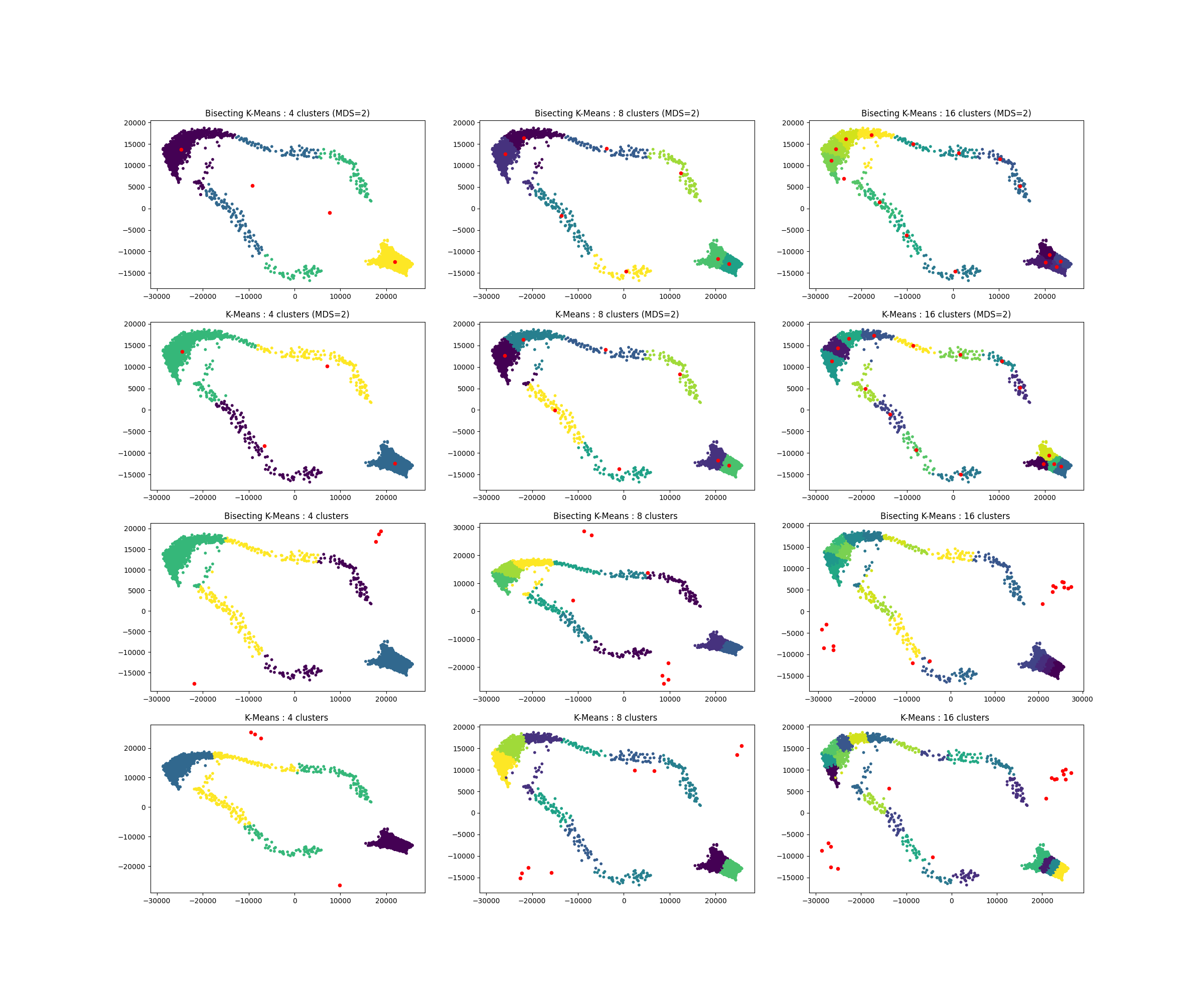}}
        \caption{K-Means ($l=-1$)}
    \end{subfigure}
    \hfill
    \begin{subfigure}{0.22\textwidth}
        \fbox{\includegraphics[width=0.9\linewidth,height=0.55\linewidth,trim={2.95cm 2.3cm 2.4cm 2.6cm},clip]{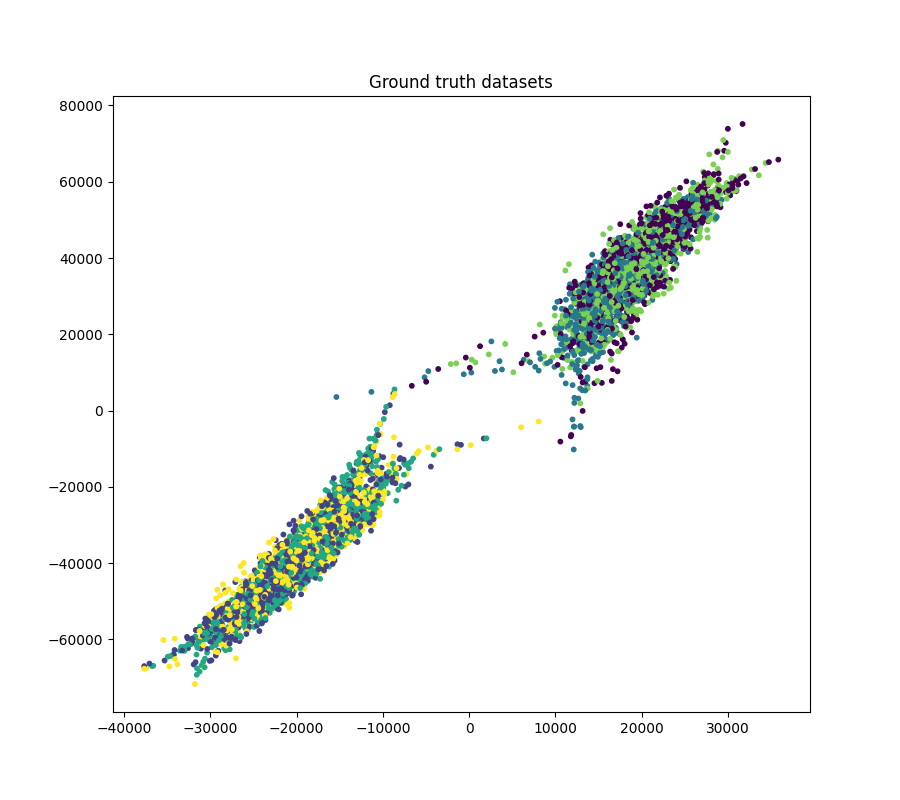}}
        \caption{Ground Truth ($l=-2$)}
    \end{subfigure}
    \hfill
    \begin{subfigure}{0.22\textwidth}
        \fbox{\includegraphics[width=0.9\linewidth,height=0.55\linewidth,trim={7.8cm 26.1cm 39.5cm 16.4cm},clip]{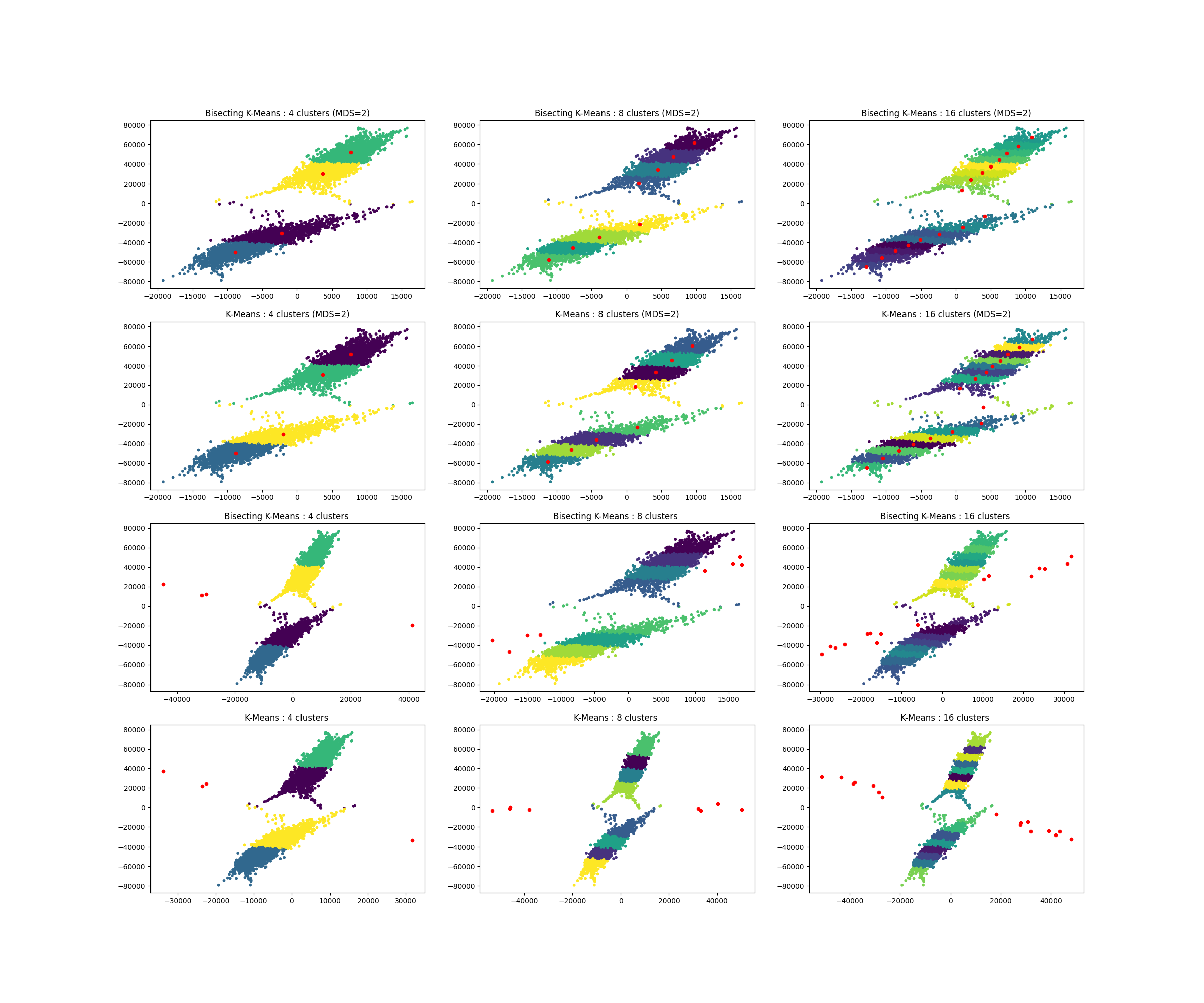}}
        \caption{K-Means ($l=-2$)}
    \end{subfigure}

    \caption{Clusters formed from the hidden states of LLaMA2 on the ACRE o.o.d sets. The visualisations contain the last two levels ($l$) of hidden layers. The ground truth shows the true splits (text/symbolic/o.o.d splits). The learned clusters use 4 centroids.  }
    \label{fig:supp_acre_ood_clusters}
\end{figure}

\begin{figure}
    \centering
    \begin{subfigure}{0.22\textwidth}
        \fbox{\includegraphics[width=0.9\linewidth,height=0.55\linewidth,trim={2.95cm 2.3cm 2.4cm 2.6cm},clip]{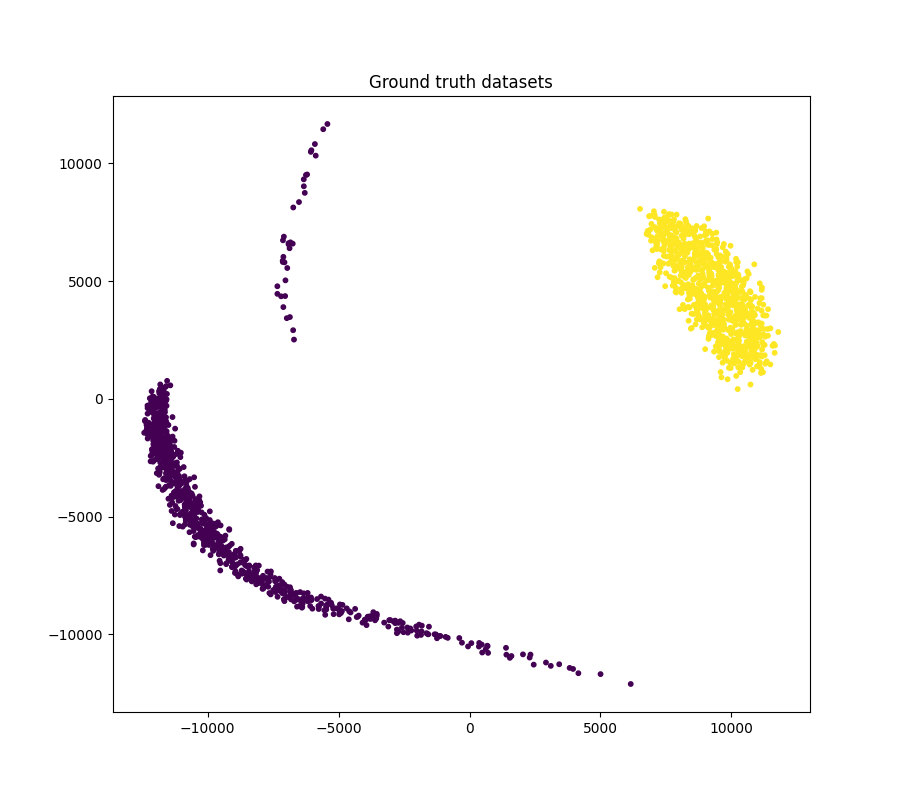}}
        \caption{Ground Truth ($l=-1$)}
    \end{subfigure}
    \hfill
    \begin{subfigure}{0.22\textwidth}
        \fbox{\includegraphics[width=0.9\linewidth,height=0.55\linewidth,trim={7.8cm 26.1cm 39.5cm 16.4cm},clip]{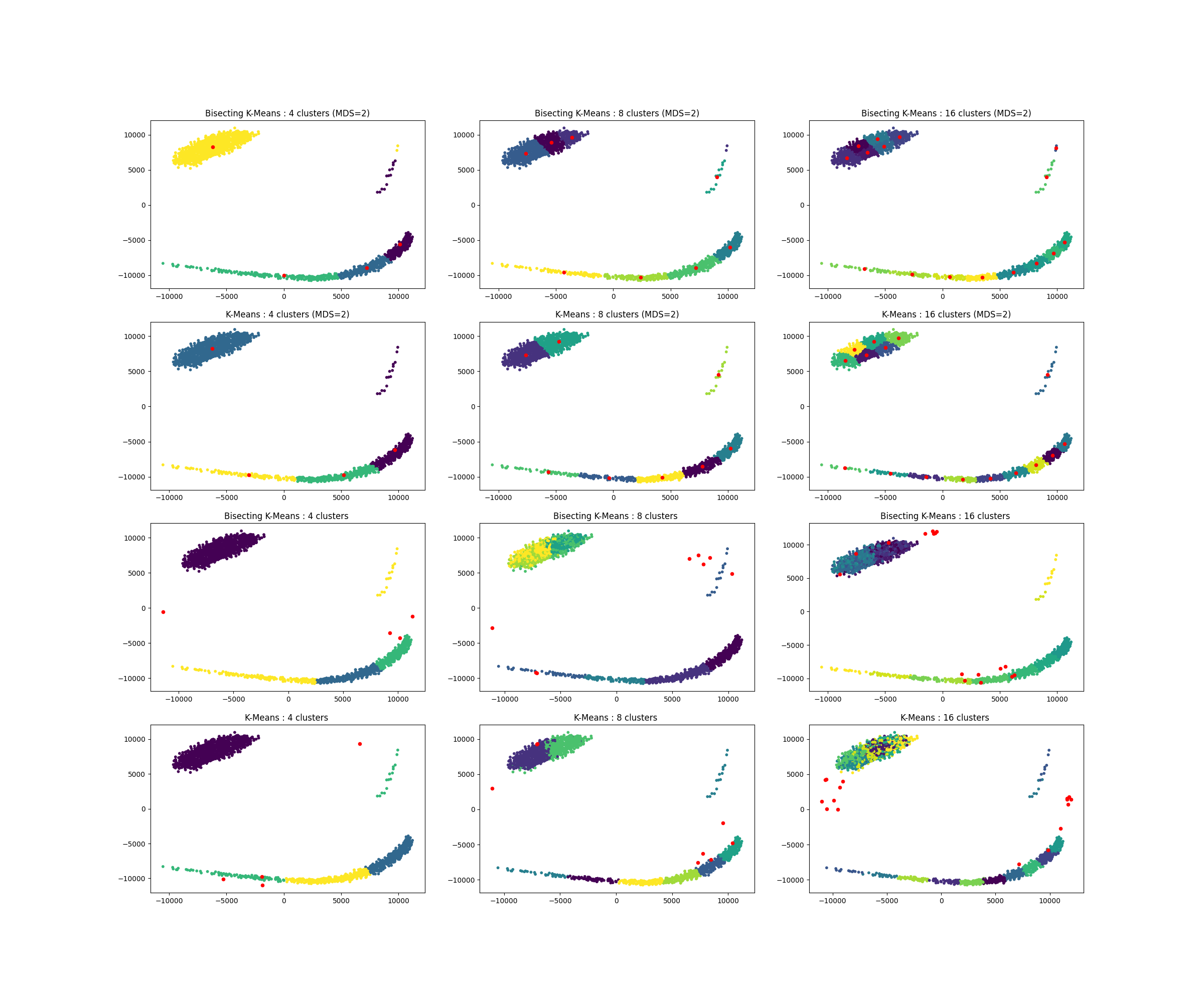}}
        \caption{K-Means ($l=-1$)}
    \end{subfigure}
    \hfill
    \begin{subfigure}{0.22\textwidth}
        \fbox{\includegraphics[width=0.9\linewidth,height=0.55\linewidth,trim={2.95cm 2.3cm 2.4cm 2.6cm},clip]{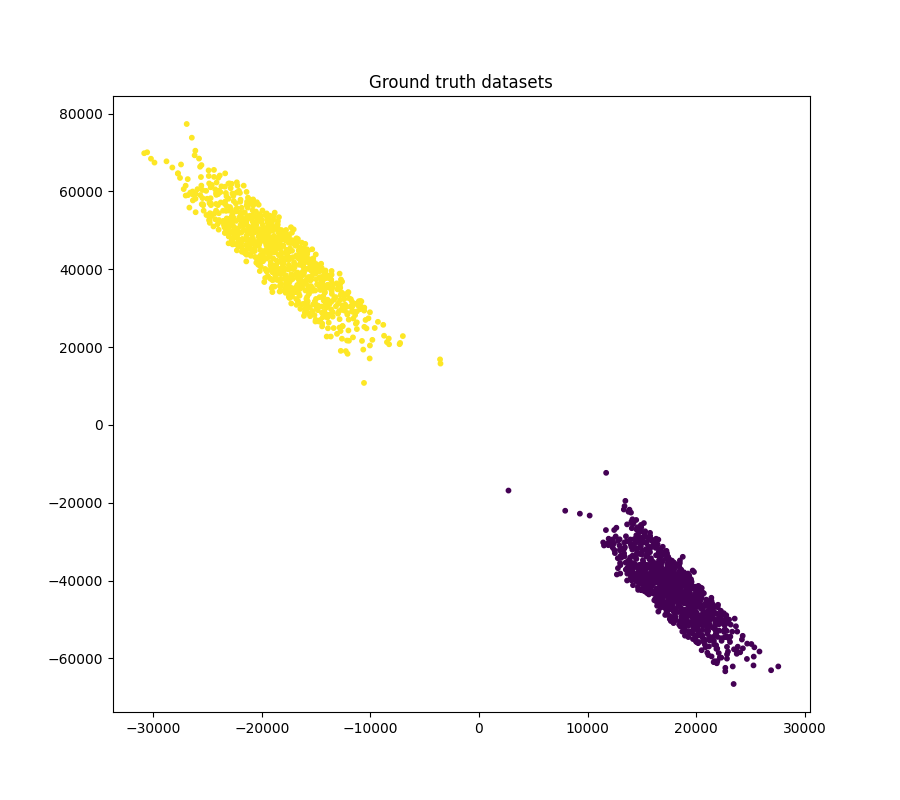}}
        \caption{Ground Truth ($l=-2$)}
    \end{subfigure}
    \hfill
    \begin{subfigure}{0.22\textwidth}
        \fbox{\includegraphics[width=0.9\linewidth,height=0.55\linewidth,trim={7.8cm 26.1cm 39.5cm 16.4cm},clip]{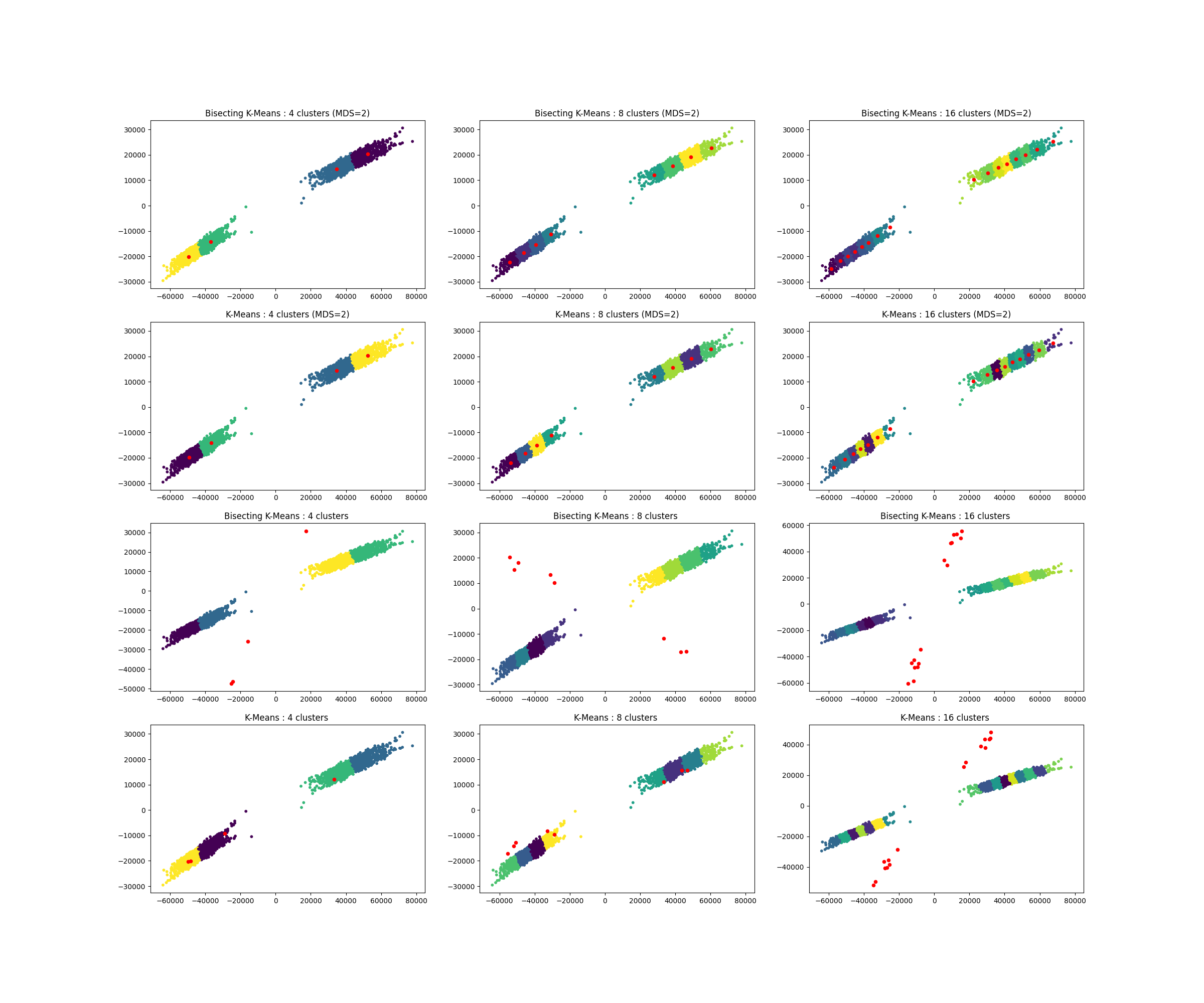}}
        \caption{K-Means ($l=-2$)}
    \end{subfigure}

    \caption{Clusters formed from the hidden states of LLaMA2 on the RAVEN training sets. The visualisations contain the last two levels ($l$) of hidden layers. The ground truth shows the true splits (text/symbolic). The learned clusters use 4 centroids. }
    \label{fig:supp_raven_clusters}
\end{figure}

\begin{figure}
    \centering
    \begin{subfigure}{0.22\textwidth}
        \fbox{\includegraphics[width=0.9\linewidth,height=0.55\linewidth,trim={2.95cm 2.3cm 2.4cm 2.6cm},clip]{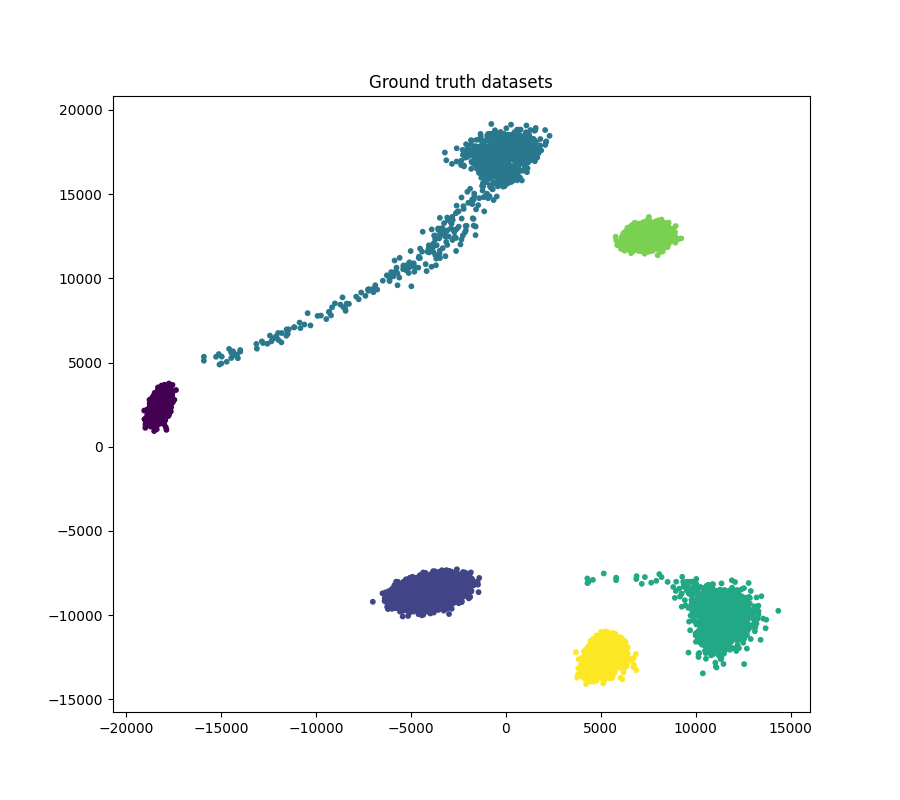}}
        \caption{Ground Truth ($l=-1$)}
    \end{subfigure}
    \hfill
    \begin{subfigure}{0.22\textwidth}
        \fbox{\includegraphics[width=0.9\linewidth,height=0.55\linewidth,trim={7.8cm 26.1cm 39.5cm 16.4cm},clip]{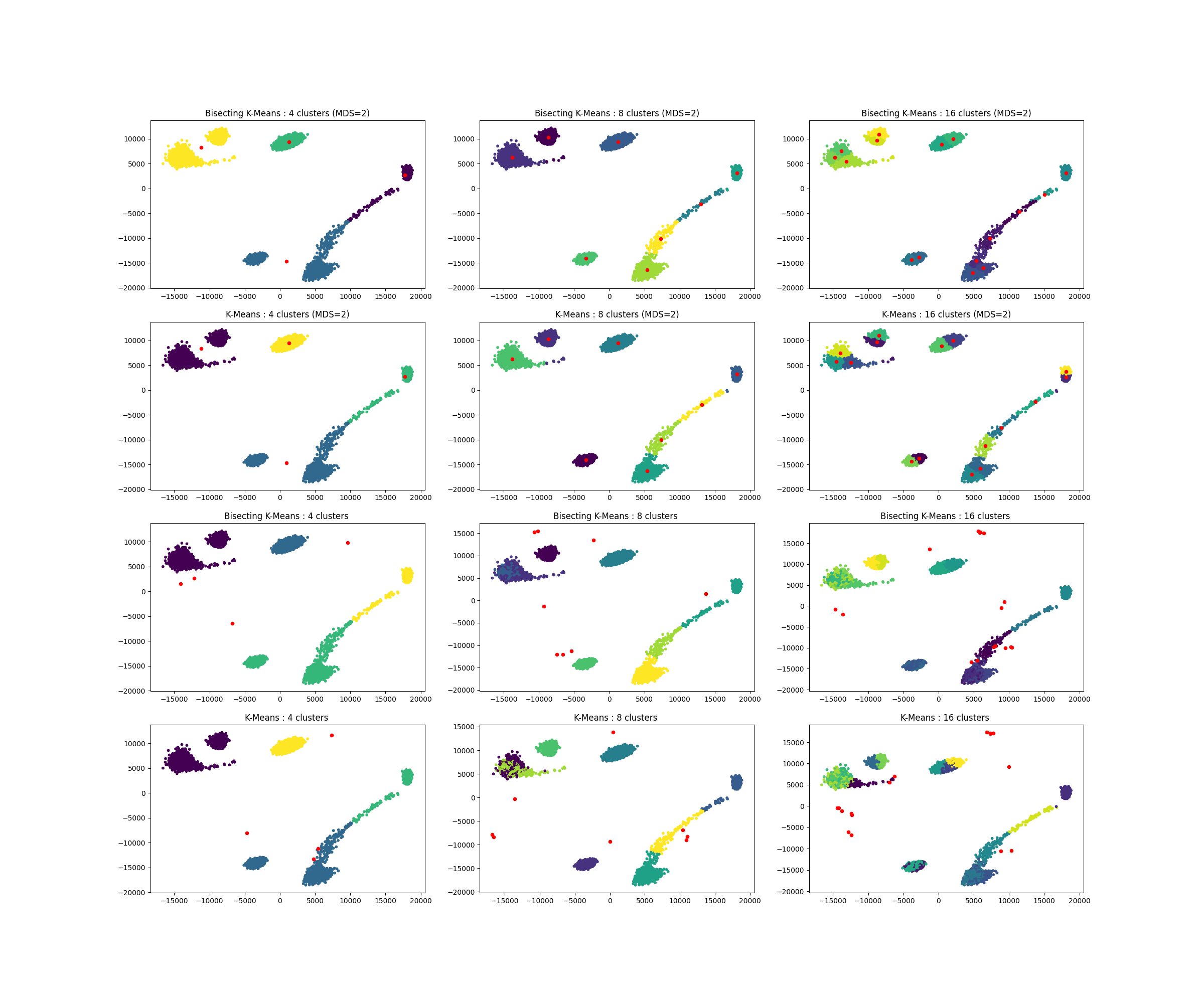}}
        \caption{K-Means ($l=-1$)}
    \end{subfigure}
    \hfill
    \begin{subfigure}{0.22\textwidth}
        \fbox{\includegraphics[width=0.9\linewidth,height=0.55\linewidth,trim={2.95cm 2.3cm 2.4cm 2.6cm},clip]{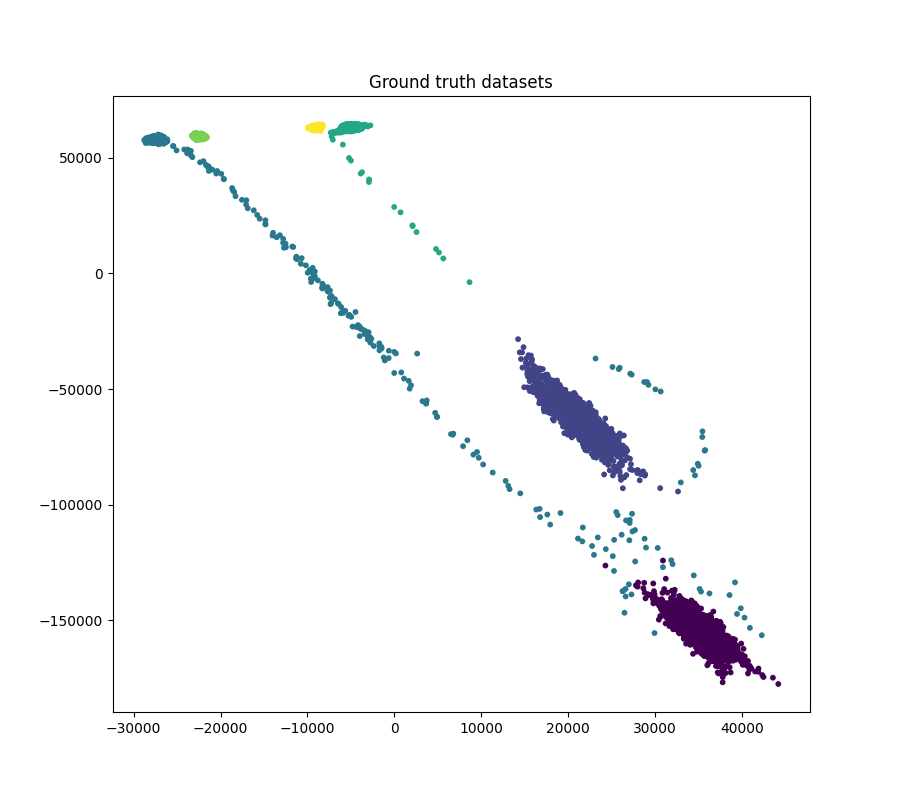}}
        \caption{Ground Truth ($l=-2$)}
    \end{subfigure}
    \hfill
    \begin{subfigure}{0.22\textwidth}
        \fbox{\includegraphics[width=0.9\linewidth,height=0.55\linewidth,trim={7.8cm 26.1cm 39.5cm 16.4cm},clip]{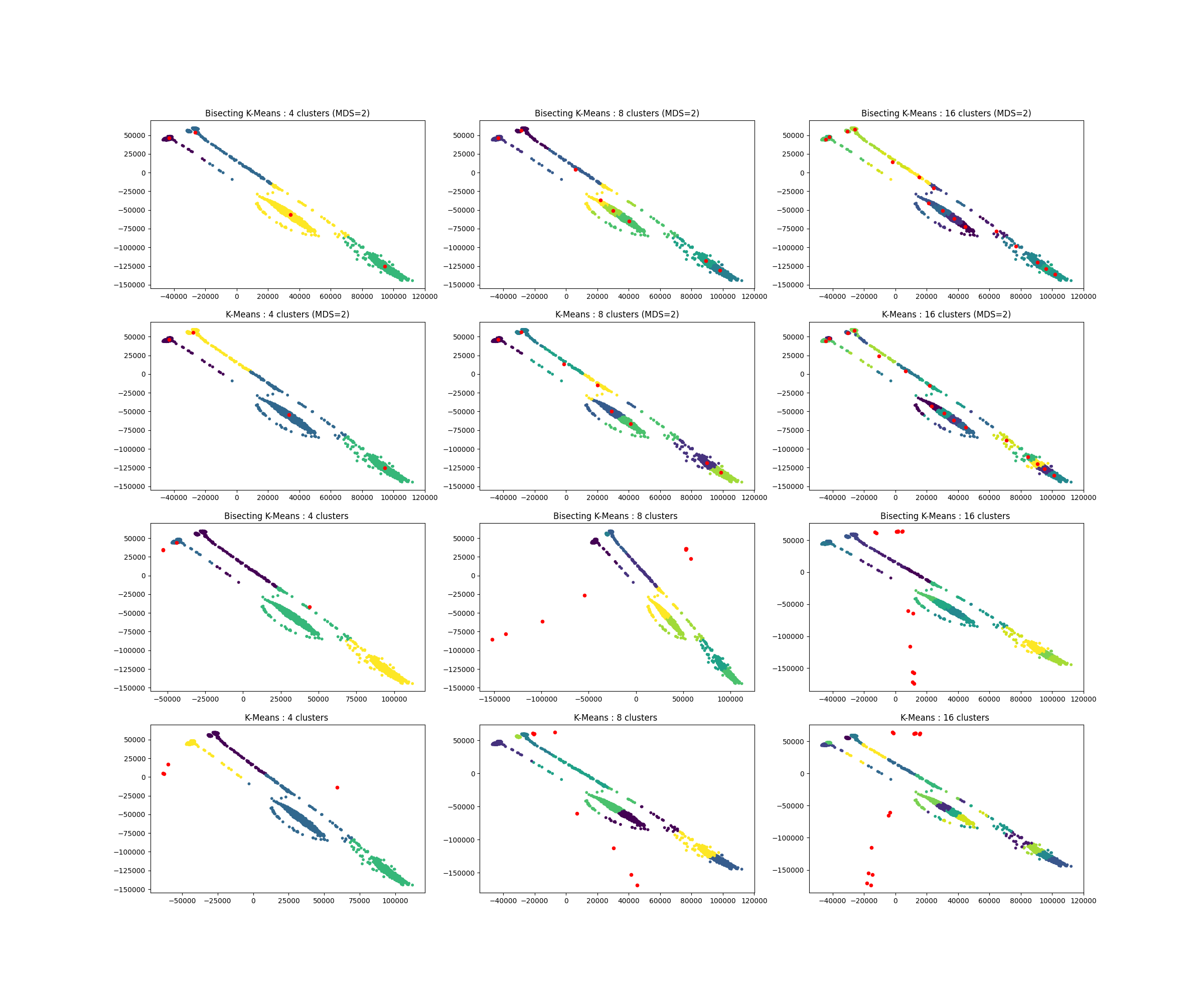}}
        \caption{K-Means ($l=-2$)}
    \end{subfigure}

    \caption{Clusters formed from the hidden states of LLaMA2 on the RAVEN o.o.d sets. The visualisations contain the last two levels ($l$) of hidden layers. The ground truth shows the true splits (text/symbolic). The learned clusters use 4 centroids. }
    \label{fig:supp_raven_ood_clusters}
\end{figure}

Figures \ref{fig:supp_acre_clusters}, \ref{fig:supp_acre_ood_clusters}, \ref{fig:supp_raven_clusters} and \ref{fig:supp_raven_ood_clusters} show visualisations of the clusters in a 2D space. Figures \ref{fig:supp_acre_clusters} and \ref{fig:supp_acre_ood_clusters} show the i.i.d and o.o.d sets of ACRE. Figures \ref{fig:supp_raven_clusters} and \ref{fig:supp_raven_ood_clusters} show the i.i.d and o.o.d sets of RAVEN. As in the main paper, the projection is made using Multidimensional Scaling (MDS) \citep{borg2005modern}. For illustration purposes, we observe the clusters formed by the K-Means method for $N=4$ modules. We also observe the clusters formed from the penultimate hidden states of the router. As discussed above and in the main paper, there is a clear division between text and symbolic embeddings, but the o.o.d sets are not well separated in ACRE while they are in RAVEN. This division (and absence of division) is also present in the previous hidden states, although the separation is less obvious: all embeddings tend to align to a single axis.

\begin{figure}
    \centering
    \begin{subfigure}{0.22\textwidth}
        \fbox{\includegraphics[width=0.9\linewidth,height=0.55\linewidth,trim={2.95cm 2.3cm 2.4cm 2.6cm},clip]{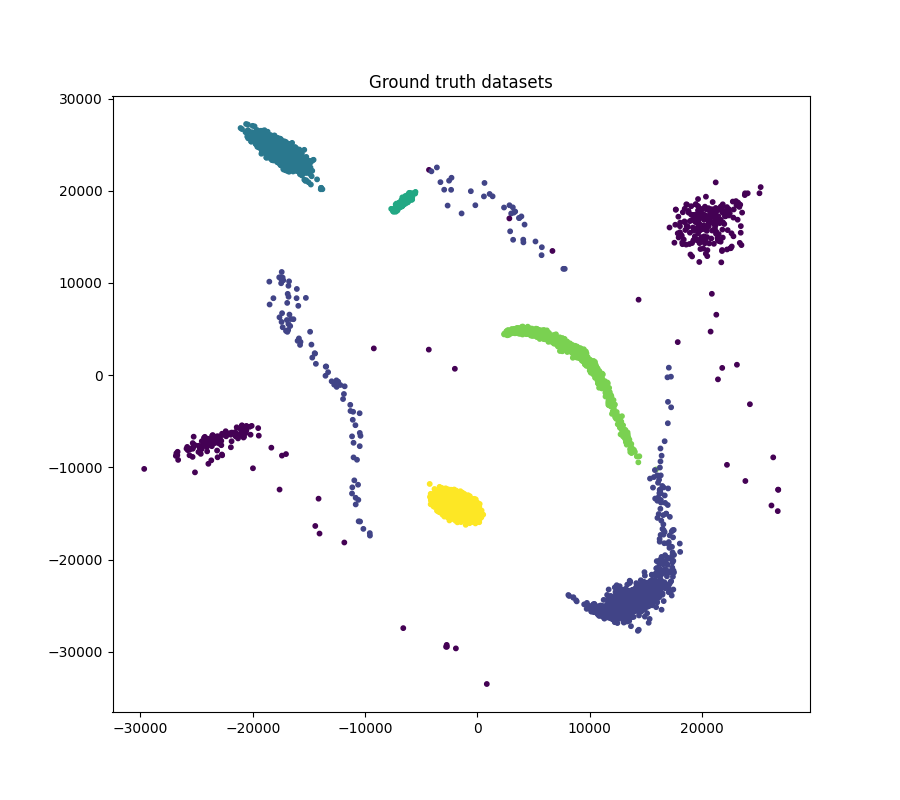}}
        \caption{Ground Truth ($l=-1$)}
    \end{subfigure}
    \begin{subfigure}{0.22\textwidth}
        \fbox{\includegraphics[width=0.9\linewidth,height=0.55\linewidth,trim={7.8cm 26.1cm 39.5cm 16.4cm},clip]{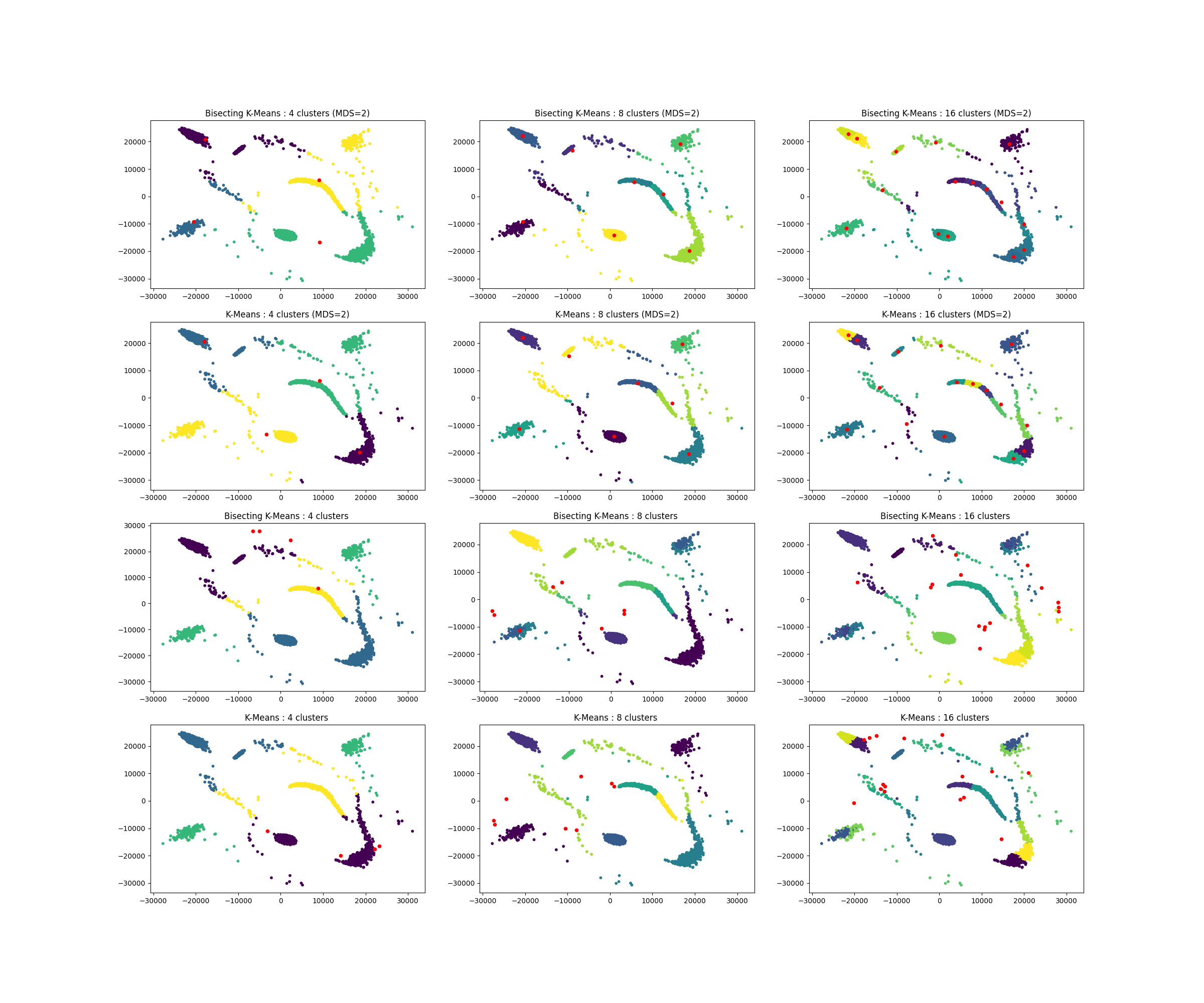}}
        \caption{K-Means ($l=-1$)}
    \end{subfigure}
    \begin{subfigure}{0.22\textwidth}
        \fbox{\includegraphics[width=0.9\linewidth,height=0.55\linewidth,trim={2.95cm 2.3cm 2.4cm 2.6cm},clip]{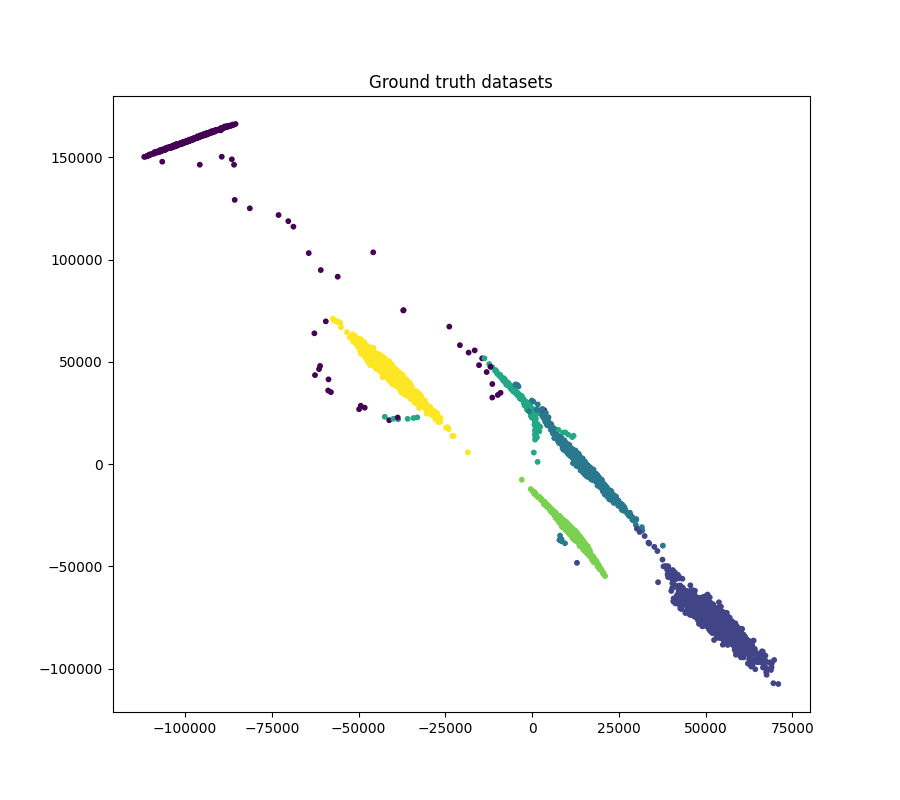}}
        \caption{Ground Truth ($l=-2$)}
    \end{subfigure}
    \begin{subfigure}{0.22\textwidth}
        \fbox{\includegraphics[width=0.9\linewidth,height=0.55\linewidth,trim={7.8cm 26.1cm 39.5cm 16.4cm},clip]{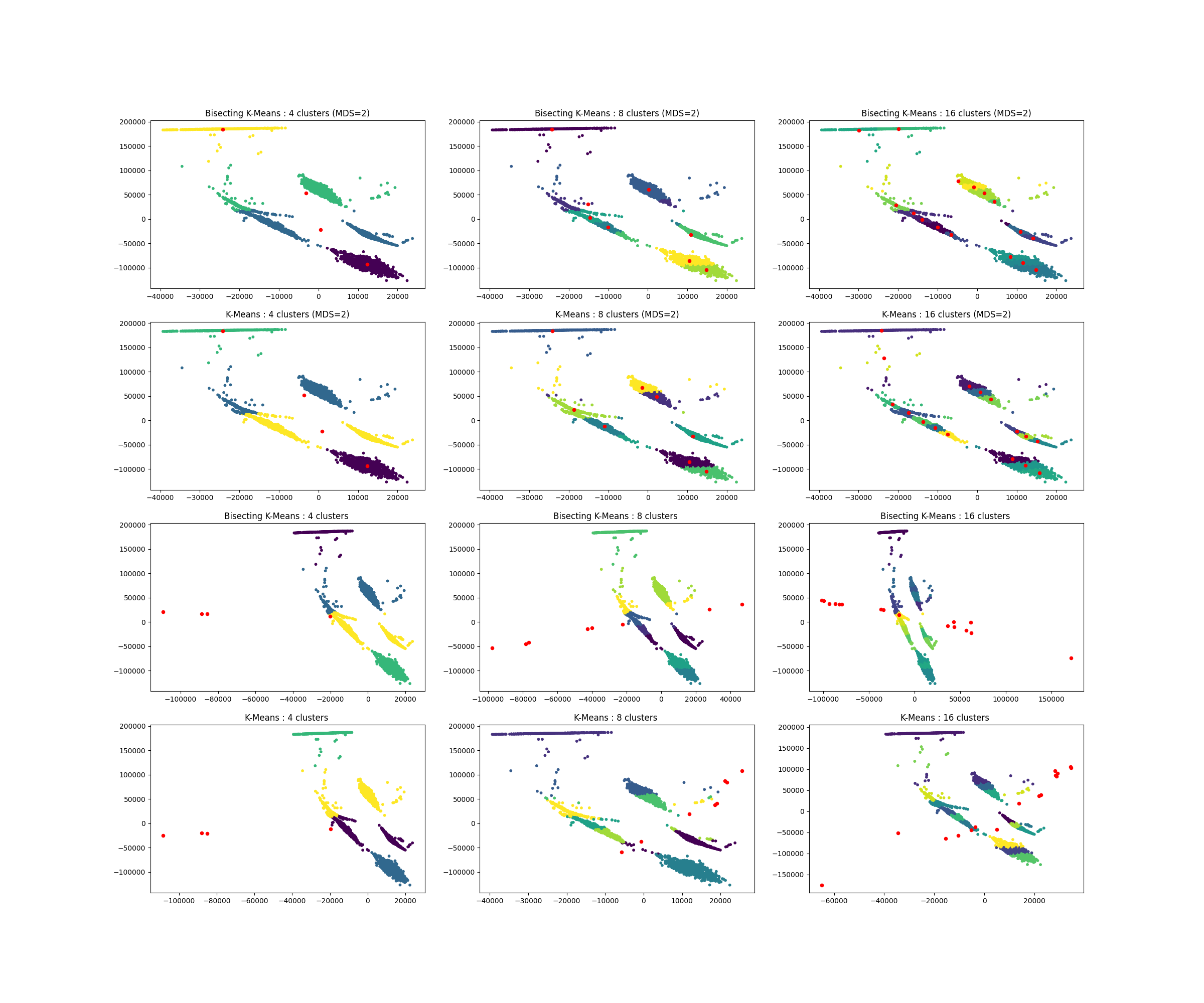}}
        \caption{K-Means ($l=-2$)}
    \end{subfigure}

    \caption{Clusters formed from the hidden states of LLaMA2 on the training sets of ACRE, ARC, PVR and RAVEN. The visualisations contain the last two levels ($l$) of hidden layers. The ground truth shows the true splits between each dataset. The learned clusters use 4 centroids. }
    \label{fig:supp_abstract_clusters}
\end{figure}

We want to study the router's behaviour further when faced with a diverse set of input data. To this end, we feed six different datasets to the model: the i.i.d text and symbolic sets of ACRE and RAVEN, PVR \citep{DBLP:journals/corr/abs-2107-12580} and ARC \citep{DBLP:journals/corr/abs-1911-01547} datasets. The visualisations are in Figure \ref{fig:supp_abstract_clusters}. Overall, the datasets are well separated but have different shapes. While some form dense amalgamates, others spread in the latent space. The observations from ACRE and RAVEN suggest that the distance in the embedding space between a module cluster and an input can be an indicator of the module's performance on the input. The o.o.d sets of ACRE are merged in the latent space, and the model maintains accuracy across the sets. In parallel, the o.o.d sets of RAVEN are separated by clear boundaries, and the accuracy drops as the distance with the i.i.d set increases.
Experiments on a larger scale are needed to validate or invalidate the hypothesis and discriminate the true causes responsible for this behaviour from spurious correlations.

%% file: experiments/cluster_alternatives.tex
\subsection{Variations of the Routing Strategy}
\label{sec:routing_variations}

We perform additional experiments on ACRE and RAVEN datasets using the routing strategies introduced in Appendix \ref{sec:supp_routing_strategy}: K-Means and weighting. Tables \ref{tab:acre_routing_alternatives} and \ref{tab:raven_routing_alternatives} show the results.

\begin{table*}[ht]
    \centering
    \caption{Accuracy on the ACRE i.i.d and o.o.d test sets. Datasets are represented in columns, and models in rows. ICLM is trained on text and symbolic i.i.d training sets. Models with a $^*$ indicate that the results are introduced in this paper. The best model is shown in \textbf{bold}.  }
    \begin{tabular}{lcccccc}
        \hline
         & \multicolumn{2}{c}{ACRE} & \multicolumn{2}{c}{-o.o.d-Comp} & \multicolumn{2}{c}{-o.o.d-Sys} \\
        \cline{2-7}
         & Text & Symb & Text & Symb & Text & Symb \\
        \hline
        ICLM* (ours) & 0.653 & \textbf{0.950} & 0.663 & 0.931 & 0.634 & 0.901 \\
        ICLM-Weighted* (ours) & 0.812 & 0.921 & 0.802 & \textbf{0.960} & 0.842 & \textbf{0.970} \\
        ICLM-K-Means* (ours) & \textbf{0.901} & 0.881 & \textbf{0.911} & 0.911 & \textbf{0.891} & 0.921  \\
        \hline
    \end{tabular}
    \label{tab:acre_routing_alternatives}
\end{table*}

\begin{table*}[ht]
    \centering
    \caption{Accuracy on the RAVEN i.i.d and o.o.d test sets. The characteristics are the same as in Table \ref{tab:acre_routing_alternatives}. }
    \begin{tabular}{lcccccc}
        \hline
         & \multicolumn{2}{c}{RAVEN} & \multicolumn{2}{c}{-o.o.d-Four} & \multicolumn{2}{c}{-o.o.d-In-Center} \\
        \cline{2-7}
         & Text & Symb & Text & Symb & Text & Symb \\
        \hline
        ICLM* (ours) & \textbf{1.000} & 0.980 & 0.703 & \textbf{0.703} & 0.515 & 0.228 \\
        ICLM-Weighted* (ours) & \textbf{1.000} & \textbf{1.000} & \textbf{0.743} & \textbf{0.703} & \textbf{0.653} & 0.248 \\
        ICLM-K-Means* (ours) & \textbf{1.000} & \textbf{1.000} & 0.634 & 0.673 & 0.515 & \textbf{0.287} \\
        \hline
    \end{tabular}
    \label{tab:raven_routing_alternatives}
\end{table*}

The alternative routing strategies achieve similar and sometimes superior performance than the base ICLM model. As observed in the previous section, the router creates well-defined clusters that the K-Means and Euclidean distance vector quantisation strategies tend to follow. No explicit differentiation of the routing process can be observed from the visualisations. The difference in performance may lie in the optimisation process. K-Means does not backpropagate information to the router; weighting backpropagates from the output loss, and vector quantisation backpropagates from a secondary loss.

%% file: experiments/aggregation_schemes.tex
\subsection{Variations of the Aggregation Scheme}
\label{sec:aggregation_variations}

We perform additional experiments on ACRE and RAVEN datasets using the aggregation schemes introduced in Appendix \ref{sec:supp_aggregation}: in the logit and probability spaces. Tables \ref{tab:acre_routing_alternatives} and \ref{tab:raven_routing_alternatives} show the results.

\begin{table*}[ht]
    \centering
    \caption{Accuracy on the ACRE i.i.d and o.o.d test sets.  }
    \begin{tabular}{lcccccc}
        \hline
         & \multicolumn{2}{c}{ACRE} & \multicolumn{2}{c}{-o.o.d-Comp} & \multicolumn{2}{c}{-o.o.d-Sys} \\
        \cline{2-7}
         & Text & Symb & Text & Symb & Text & Symb \\
        \hline
        ICLM* (ours) & 0.653 & \textbf{0.950} & 0.663 & \textbf{0.931} & 0.634 & 0.901 \\
        ICLM-Logits* (ours) & 0.842 & \textbf{0.950} & \textbf{0.881} & 0.901 & \textbf{0.901} & 0.921 \\
        ICLM-Probas* (ours) & \textbf{0.921} & \textbf{0.950} & 0.832 & 0.921 & 0.891 & \textbf{0.931} \\
        \hline
    \end{tabular}
    \label{tab:acre_aggregation_variations}
\end{table*}

\begin{table*}[!ht]
    \centering
    \caption{Accuracy on the RAVEN i.i.d and o.o.d test sets. }
    \begin{tabular}{lcccccc}
        \hline
         & \multicolumn{2}{c}{RAVEN} & \multicolumn{2}{c}{-o.o.d-Four} & \multicolumn{2}{c}{-o.o.d-In-Center} \\
        \cline{2-7}
         & Text & Symb & Text & Symb & Text & Symb \\
        \hline
        ICLM* (ours) & \textbf{1.000} & 0.980 & \textbf{0.703} & 0.703 & 0.515 & 0.228 \\
        ICLM-Logits* (ours) & \textbf{1.000} & \textbf{0.990} & 0.644 & \textbf{0.713} & \textbf{0.703} & 0.268 \\
        ICLM-Probas* (ours) & 0.802 & 0.931 & 0.614 & 0.624 & 0.495 & \textbf{0.297} \\
        \hline
    \end{tabular}
    \label{tab:raven_aggregation_variations}
\end{table*}

As per the routing strategies, the alternative aggregation schemes achieve similar and sometimes superior performance than the base ICLM model. No scheme is systematically better than the others. These results show that less expressive aggregation methods, i.e. weighted sums, can perform similarly to trained dense layers on abstract and causal reasoning tasks.

%% file: experiments/supp_corr.tex
\subsection{Module Correlation Across Hidden States}
\label{sec:supp_corr}

\begin{figure}[t]
    \centering
    \begin{subfigure}{0.19\linewidth}
        \includegraphics[width=\linewidth]{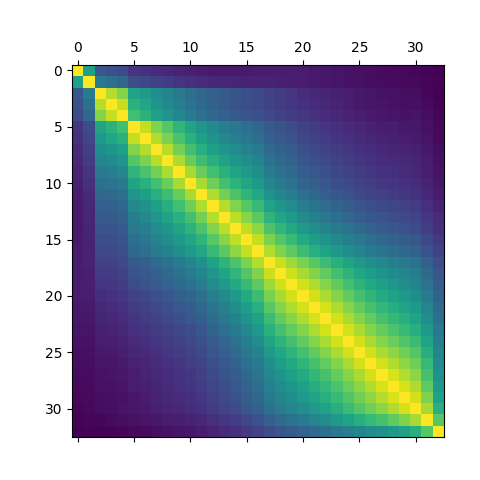}
        \caption{Router-Router. }
        \label{fig:router-router-corr-acre}
    \end{subfigure}
    \hfill
    \begin{subfigure}{0.19\linewidth}
        \includegraphics[width=\linewidth]{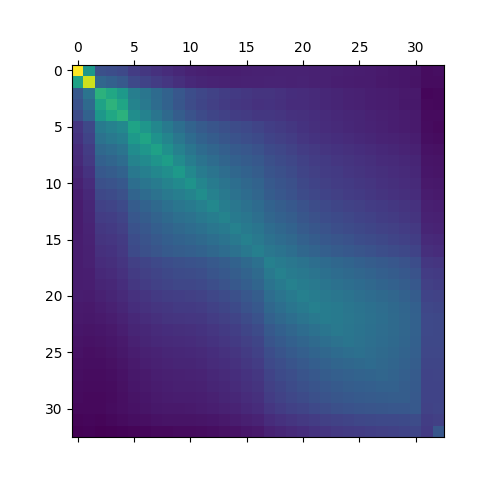}
        \caption{Router-Inv. }
        \label{fig:router-inv-corr-acre}
    \end{subfigure}
    \hfill
    \begin{subfigure}{0.19\linewidth}
        \includegraphics[width=\linewidth]{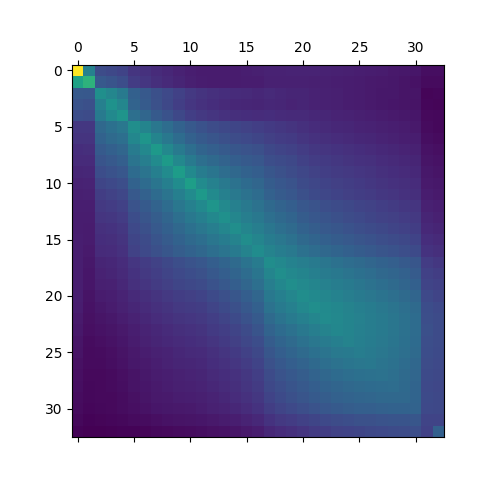}
        \caption{Router-Dom 0. }
        \label{fig:router-0-corr-acre}
    \end{subfigure}
    \hfill
    \begin{subfigure}{0.19\linewidth}
        \includegraphics[width=\linewidth]{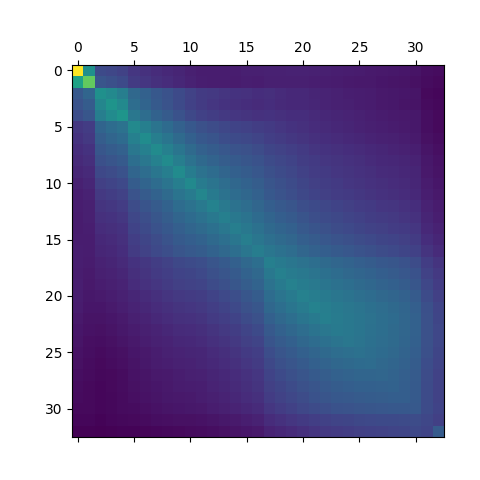}
        \caption{Router-Dom 1. }
        \label{fig:router-1-corr-acre}
    \end{subfigure}
    \hfill
    \begin{subfigure}{0.19\linewidth}
        \includegraphics[width=\linewidth]{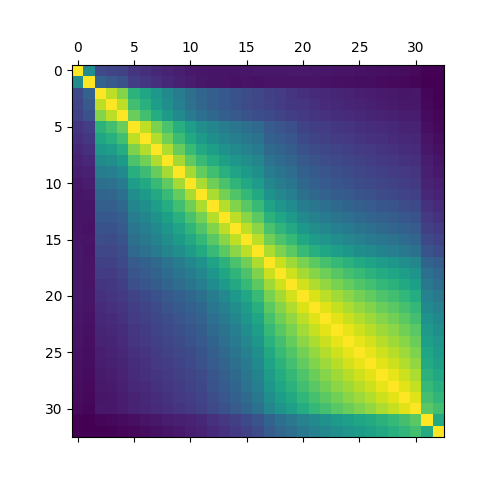}
        \caption{Inv-Inv. }
        \label{fig:inv-inv-corr-acre}
    \end{subfigure}
    \hfill
    \begin{subfigure}{0.19\linewidth}
        \includegraphics[width=\linewidth]{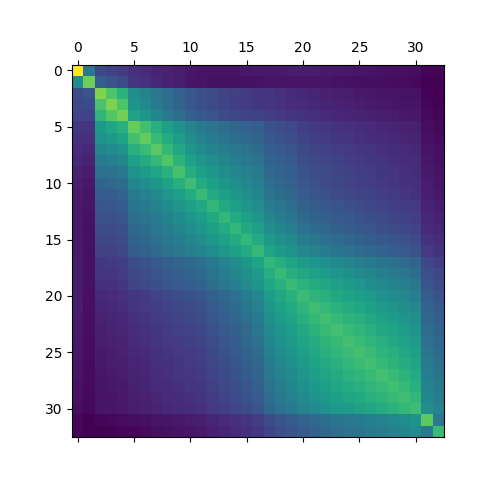}
        \caption{Inv-Dom 0. }
        \label{fig:inv-0-corr-acre}
    \end{subfigure}
    \hfill
    \begin{subfigure}{0.19\linewidth}
        \includegraphics[width=\linewidth]{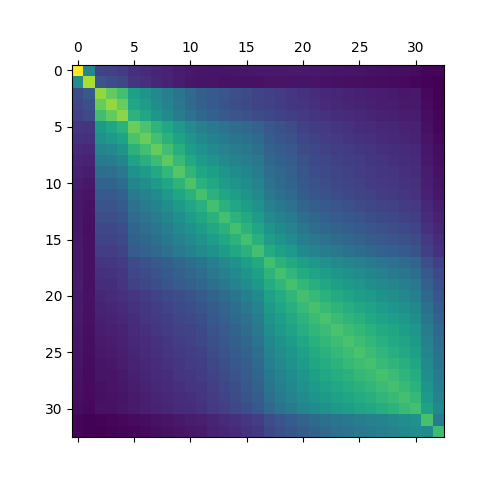}
        \caption{Inv-Dom 1. }
        \label{fig:inv-1-corr-acre}
    \end{subfigure}
    \hfill
    \begin{subfigure}{0.19\linewidth}
        \includegraphics[width=\linewidth]{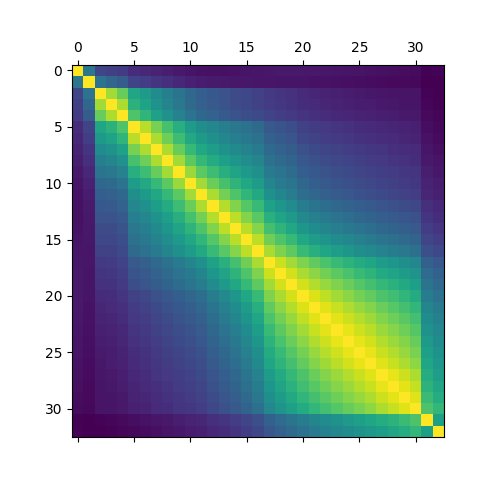}
        \caption{Dom 0-Dom 0. }
        \label{fig:0-0-corr-acre}
    \end{subfigure}
    \hfill
    \begin{subfigure}{0.19\linewidth}
        \includegraphics[width=\linewidth]{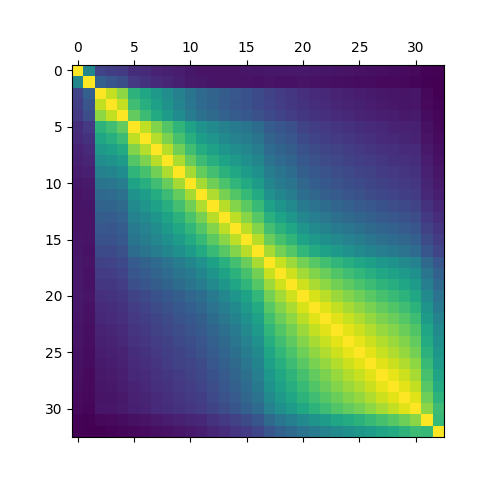}
        \caption{Dom 1-Dom 1. }
        \label{fig:1-1-corr-acre}
    \end{subfigure}
    \hfill
    \begin{subfigure}{0.19\linewidth}
        \includegraphics[width=\linewidth]{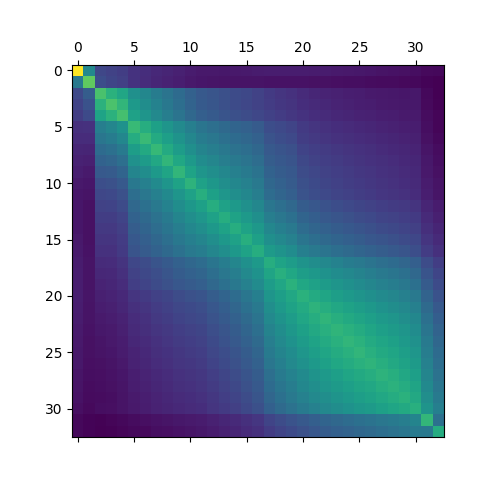}
        \caption{Dom 0-Dom 1. }
        \label{fig:0-1-corr-acre}
    \end{subfigure}
    \caption{Measures of the Pearson Correlation Coefficient between module hidden states during inference on ACRE dataset. Rows and columns represent layers 0 to 33 of a LLaMA2 module. Router refers to the routing module, Inv refers to the domain-invariant module, and Dom \textit{i} refers to the \textit{i} domain-specific module. The caption indicates the modules used for each row-column pair. }
    \label{fig:all-layer-correlations-acre}
\end{figure}

\begin{figure}[t]
    \centering
    \begin{subfigure}{0.19\linewidth}
        \includegraphics[width=\linewidth]{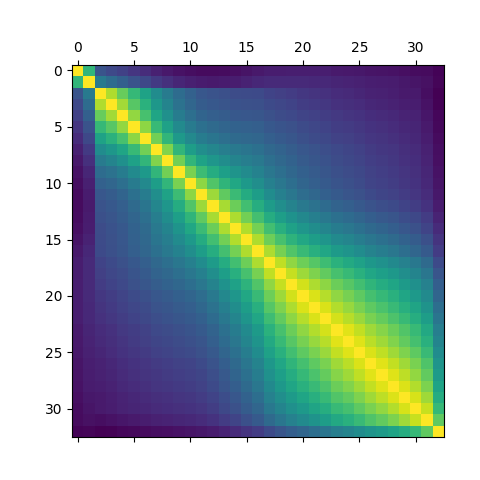}
        \caption{Router-Router. }
        \label{fig:router-router-corr-raven}
    \end{subfigure}
    \hfill
    \begin{subfigure}{0.19\linewidth}
        \includegraphics[width=\linewidth]{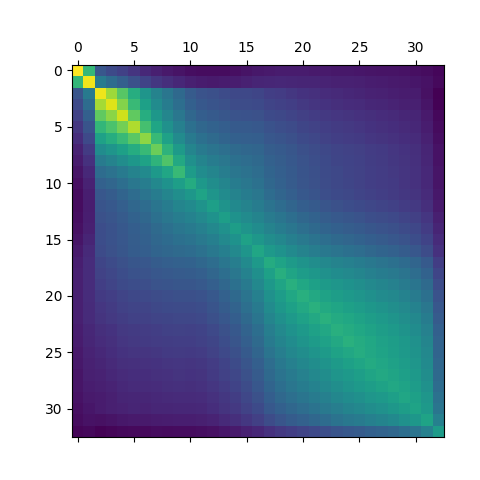}
        \caption{Router-Inv. }
        \label{fig:router-inv-corr-raven}
    \end{subfigure}
    \hfill
    \begin{subfigure}{0.19\linewidth}
        \includegraphics[width=\linewidth]{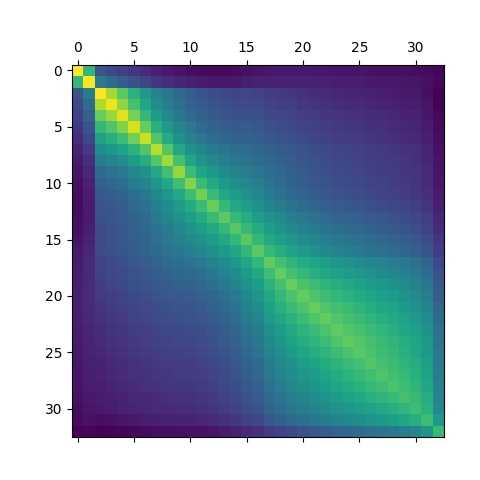}
        \caption{Router-Dom 0. }
        \label{fig:router-0-corr-raven}
    \end{subfigure}
    \hfill
    \begin{subfigure}{0.19\linewidth}
        \includegraphics[width=\linewidth]{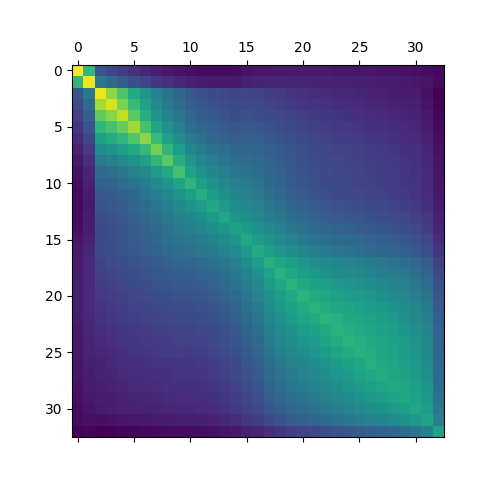}
        \caption{Router-Dom 1. }
        \label{fig:router-1-corr-raven}
    \end{subfigure}
    \hfill
    \begin{subfigure}{0.19\linewidth}
        \includegraphics[width=\linewidth]{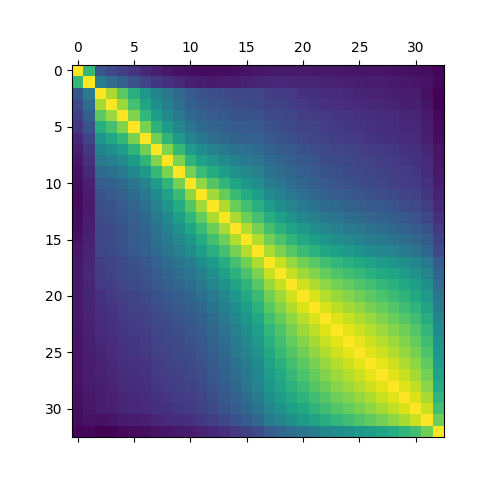}
        \caption{Inv-Inv. }
        \label{fig:inv-inv-corr-raven}
    \end{subfigure}
    \hfill
    \begin{subfigure}{0.19\linewidth}
        \includegraphics[width=\linewidth]{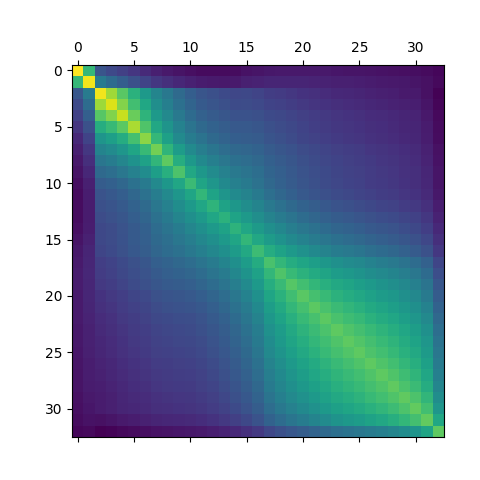}
        \caption{Inv-Dom 0. }
        \label{fig:inv-0-corr-raven}
    \end{subfigure}
    \hfill
    \begin{subfigure}{0.19\linewidth}
        \includegraphics[width=\linewidth]{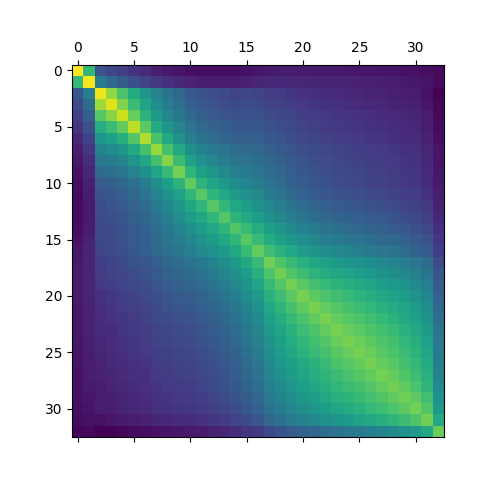}
        \caption{Inv-Dom 1. }
        \label{fig:inv-1-corr-raven}
    \end{subfigure}
    \hfill
    \begin{subfigure}{0.19\linewidth}
        \includegraphics[width=\linewidth]{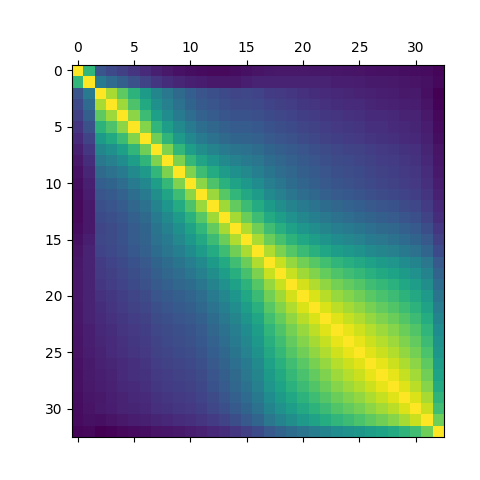}
        \caption{Dom 0-Dom 0. }
        \label{fig:0-0-corr-raven}
    \end{subfigure}
    \hfill
    \begin{subfigure}{0.19\linewidth}
        \includegraphics[width=\linewidth]{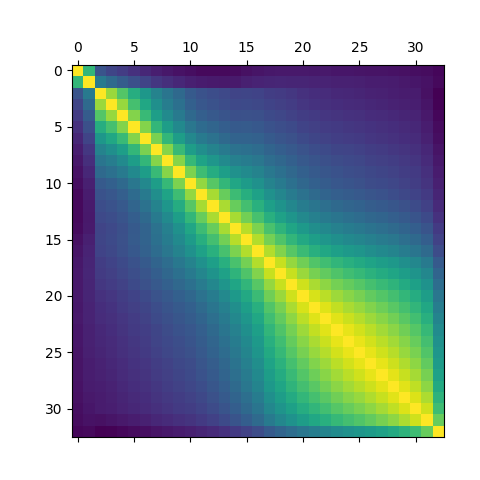}
        \caption{Dom 1-Dom 1. }
        \label{fig:1-1-corr-raven}
    \end{subfigure}
    \hfill
    \begin{subfigure}{0.19\linewidth}
        \includegraphics[width=\linewidth]{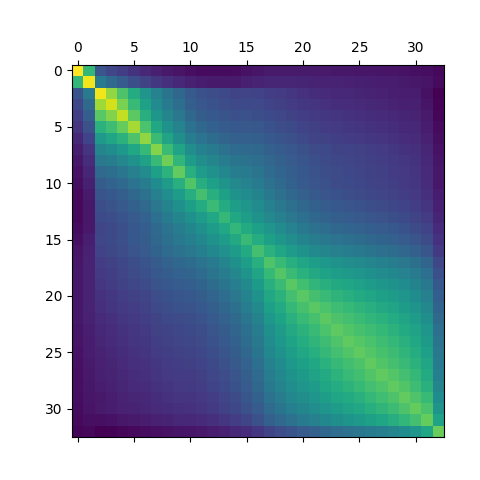}
        \caption{Dom 0-Dom 1. }
        \label{fig:0-1-corr-raven}
    \end{subfigure}
    \caption{Measures of the Pearson Correlation Coefficient between module hidden states during inference on RAVEN dataset. Rows and columns represent layers 0 to 33 of a LLaMA2 module. Router refers to the routing module, Inv refers to the domain-invariant module, and Dom \textit{i} refers to the domain-specific module \textit{i}. The caption indicates the modules used for each row-column pair. }
    \label{fig:all-layer-correlations-raven}
\end{figure}

To further investigate the level of Independence of the LLM modules, we measure the Pearson Correlation Coefficient between all hidden states of the domain-invariant and domain-specific modules during inference on ACRE and RAVEN. Figures \ref{fig:all-layer-correlations-acre} and \ref{fig:all-layer-correlations-raven} show the results. The intra-module correlations (Figures \ref{fig:router-router-corr-acre}, \ref{fig:inv-inv-corr-acre}, \ref{fig:0-0-corr-acre}, \ref{fig:1-1-corr-acre}, \ref{fig:router-router-corr-raven}, \ref{fig:inv-inv-corr-raven}, \ref{fig:0-0-corr-raven} and \ref{fig:1-1-corr-raven}) show that hidden states from close layers in the model are highly correlated. Furthermore, correlation \textit{blocks} are visible, i.e. sequences of layers that demonstrate a high level of correlation between them and a low level of correlation with the layers not in the sequence. We observe five correlation blocks well-defined on ACRE and with fuzzy-edges on RAVEN. 

These correlation blocks hold in the inter-module correlation matrices, indicating that similar mechanisms are shared across modules. However, the fine-tuning and regularisation procedures reduce the influence of the shared mechanisms and enhance specialisation.

\paragraph{Correlation Measures in Transfer Learning Settings}

\begin{figure}[t]
    \centering
    \begin{subfigure}{0.16\linewidth}
        \includegraphics[width=\linewidth]{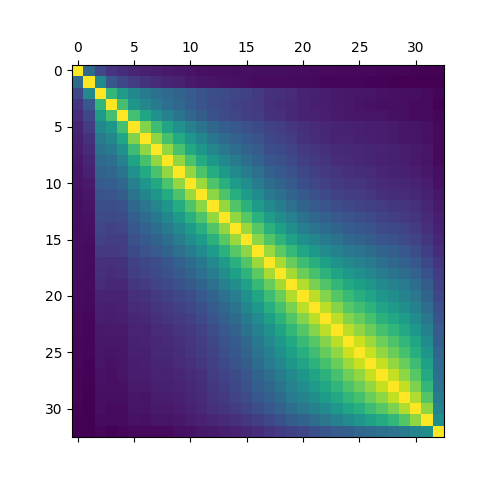}
        \caption{Router-Router. }
        \label{fig:router-router-corr-acre-transfer}
    \end{subfigure}
    \hfill
    \begin{subfigure}{0.16\linewidth}
        \includegraphics[width=\linewidth]{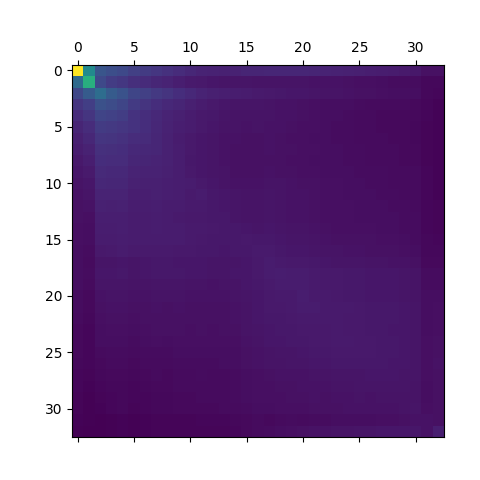}
        \caption{Router-Inv. }
        \label{fig:router-inv-corr-acre-transfer}
    \end{subfigure}
    \hfill
    \begin{subfigure}{0.16\linewidth}
        \includegraphics[width=\linewidth]{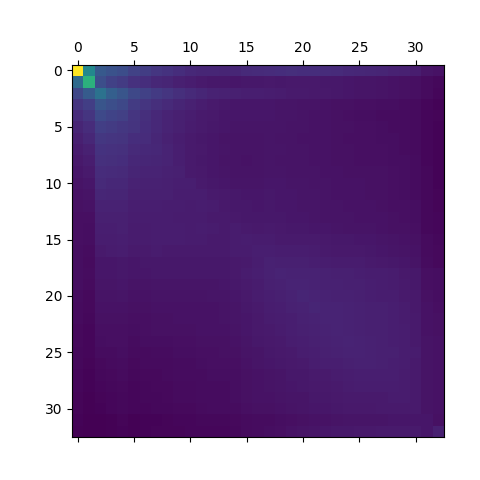}
        \caption{Router-Dom 0. }
        \label{fig:router-0-corr-acre-transfer}
    \end{subfigure}
    \hfill
    \begin{subfigure}{0.16\linewidth}
        \includegraphics[width=\linewidth]{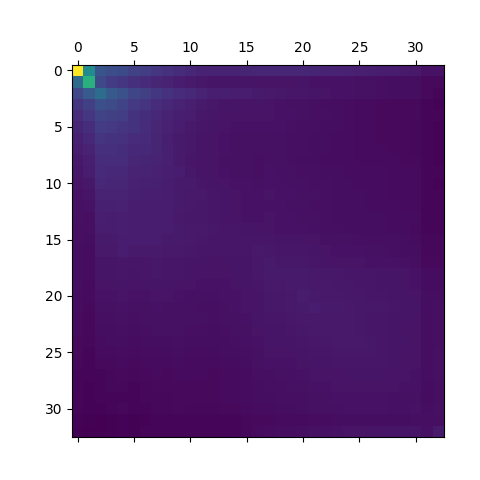}
        \caption{Router-Dom 1. }
        \label{fig:router-1-corr-acre-transfer}
    \end{subfigure}
    \hfill
    \begin{subfigure}{0.16\linewidth}
        \includegraphics[width=\linewidth]{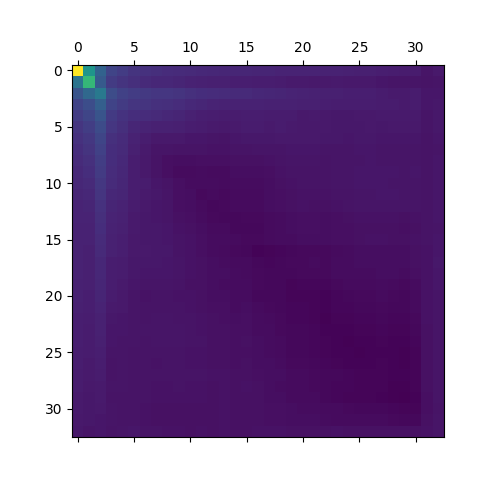}
        \caption{Router-Dom 2. }
        \label{fig:router-2-corr-acre-transfer}
    \end{subfigure}
    \hfill
    \begin{subfigure}{0.16\linewidth}
        \includegraphics[width=\linewidth]{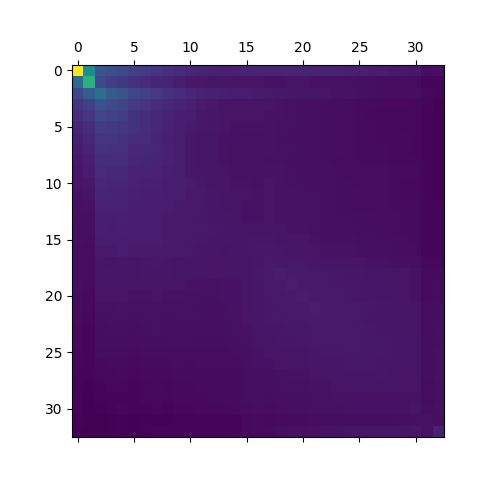}
        \caption{Router-Dom 3. }
        \label{fig:router-3-corr-acre-transfer}
    \end{subfigure}
    \hfill
    \begin{subfigure}{0.16\linewidth}
        \includegraphics[width=\linewidth]{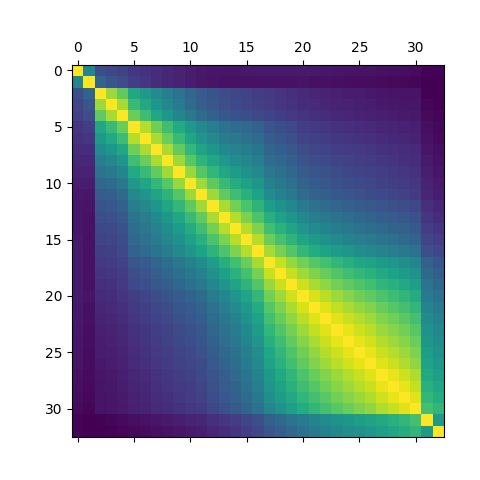}
        \caption{Inv-Inv. }
        \label{fig:inv-inv-corr-acre-transfer}
    \end{subfigure}
    \hfill
    \begin{subfigure}{0.16\linewidth}
        \includegraphics[width=\linewidth]{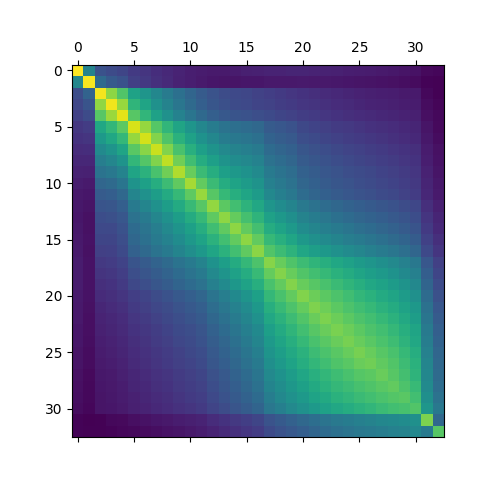}
        \caption{Inv-Dom 0. }
        \label{fig:inv-0-corr-acre-transfer}
    \end{subfigure}
    \hfill
    \begin{subfigure}{0.16\linewidth}
        \includegraphics[width=\linewidth]{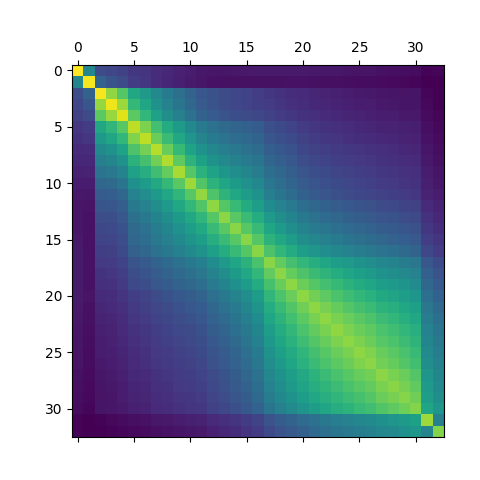}
        \caption{Inv-Dom 1. }
        \label{fig:inv-1-corr-acre-transfer}
    \end{subfigure}
    \hfill
    \begin{subfigure}{0.16\linewidth}
        \includegraphics[width=\linewidth]{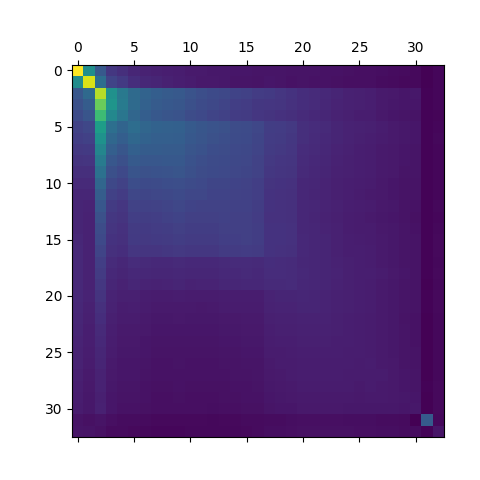}
        \caption{Inv-Dom 2. }
        \label{fig:inv-2-corr-acre-transfer}
    \end{subfigure}
    \hfill
    \begin{subfigure}{0.16\linewidth}
        \includegraphics[width=\linewidth]{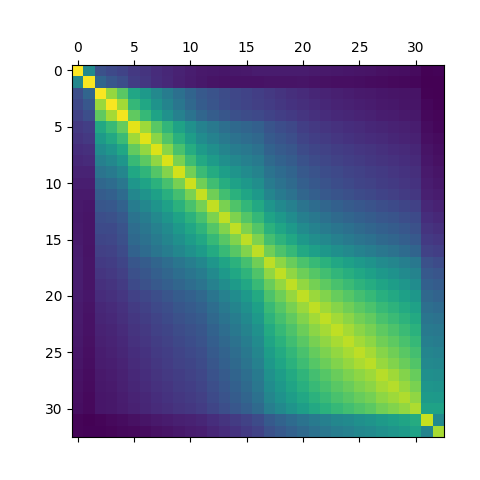}
        \caption{Inv-Dom 3. }
        \label{fig:inv-3-corr-acre-transfer}
    \end{subfigure}
    \hfill
    \begin{subfigure}{0.16\linewidth}
        \includegraphics[width=\linewidth]{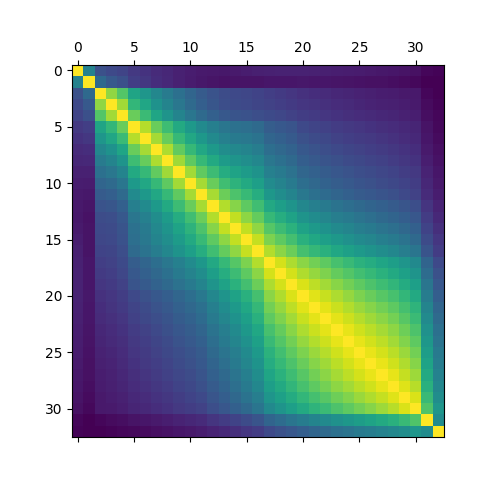}
        \caption{Dom 0-Dom 0. }
        \label{fig:0-0-corr-acre-transfer}
    \end{subfigure}
    \hfill
    \begin{subfigure}{0.16\linewidth}
        \includegraphics[width=\linewidth]{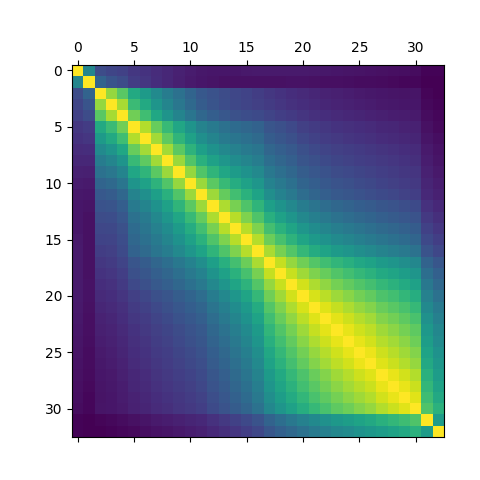}
        \caption{Dom 1-Dom 1. }
        \label{fig:1-1-corr-acre-transfer}
    \end{subfigure}
    \hfill
    \begin{subfigure}{0.16\linewidth}
        \includegraphics[width=\linewidth]{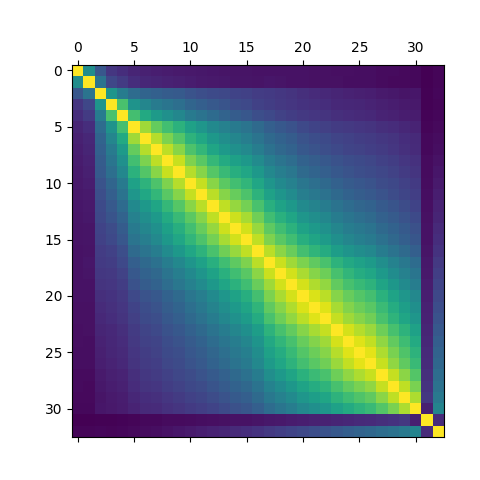}
        \caption{Dom 2-Dom 2. }
        \label{fig:2-2-corr-acre-transfer}
    \end{subfigure}
    \hfill
    \begin{subfigure}{0.16\linewidth}
        \includegraphics[width=\linewidth]{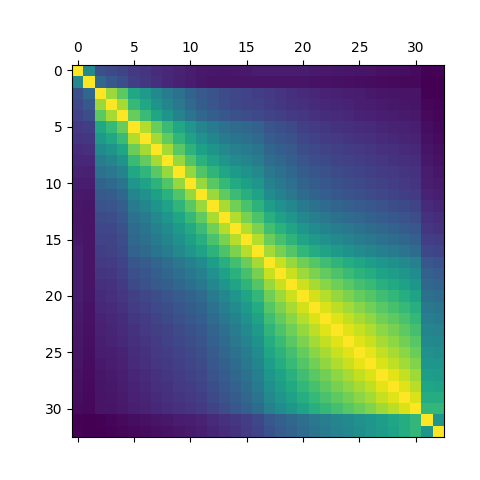}
        \caption{Dom 3-Dom 3. }
        \label{fig:3-3-corr-acre-transfer}
    \end{subfigure}
    \hfill
    \begin{subfigure}{0.16\linewidth}
        \includegraphics[width=\linewidth]{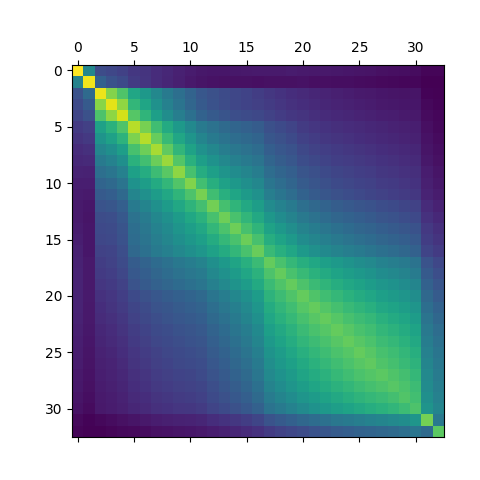}
        \caption{Dom 0-Dom 1. }
        \label{fig:0-1-corr-acre-transfer}
    \end{subfigure}
    \hfill
    \begin{subfigure}{0.16\linewidth}
        \includegraphics[width=\linewidth]{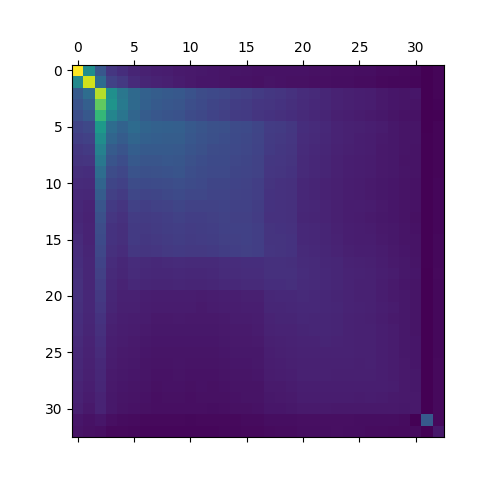}
        \caption{Dom 0-Dom 2. }
        \label{fig:0-2-corr-acre-transfer}
    \end{subfigure}
    \hfill
    \begin{subfigure}{0.16\linewidth}
        \includegraphics[width=\linewidth]{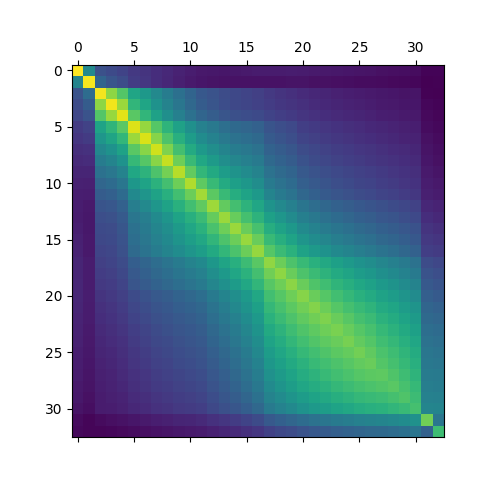}
        \caption{Dom 0-Dom 3. }
        \label{fig:0-3-corr-acre-transfer}
    \end{subfigure}
    \begin{subfigure}{0.16\linewidth}
        \includegraphics[width=\linewidth]{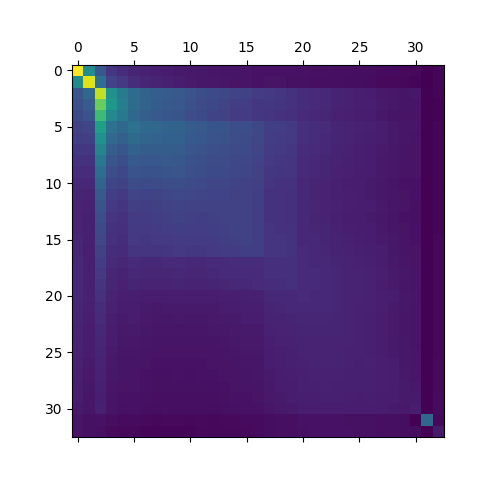}
        \caption{Dom 1-Dom 2. }
        \label{fig:1-2-corr-acre-transfer}
    \end{subfigure}
    \begin{subfigure}{0.16\linewidth}
        \includegraphics[width=\linewidth]{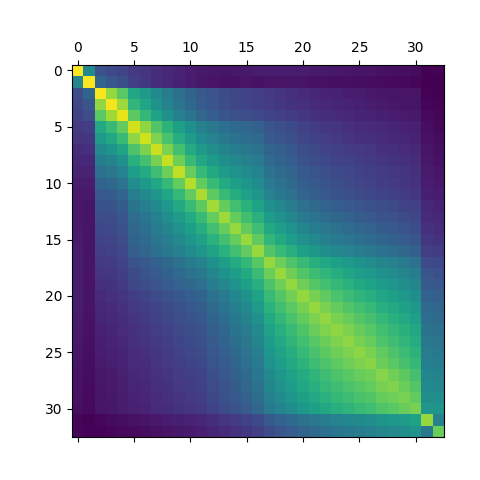}
        \caption{Dom 1-Dom 3. }
        \label{fig:1-3-corr-acre-transfer}
    \end{subfigure}
    \begin{subfigure}{0.16\linewidth}
        \includegraphics[width=\linewidth]{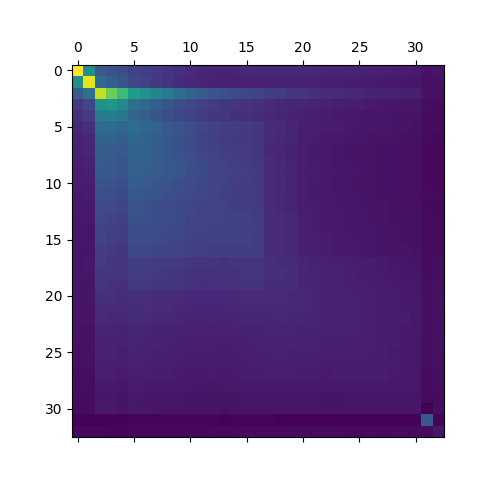}
        \caption{Dom 2-Dom 3. }
        \label{fig:2-3-corr-acre-transfer}
    \end{subfigure}
    \caption{Measures of the Pearson Correlation Coefficient between module hidden states of the transfer learning model during inference on ACRE dataset. Rows and columns represent layers 0 to 33 of a LLaMA2 module. Router refers to the routing module, Inv refers to the domain-invariant module, and Dom \textit{i} refers to the \textit{i} domain-specific module. The caption indicates the modules used for each row-column pair. }
    \label{fig:all-layer-correlations-acre-transfer}
\end{figure}

\begin{figure}[t]
    \centering
    \begin{subfigure}{0.16\linewidth}
        \includegraphics[width=\linewidth]{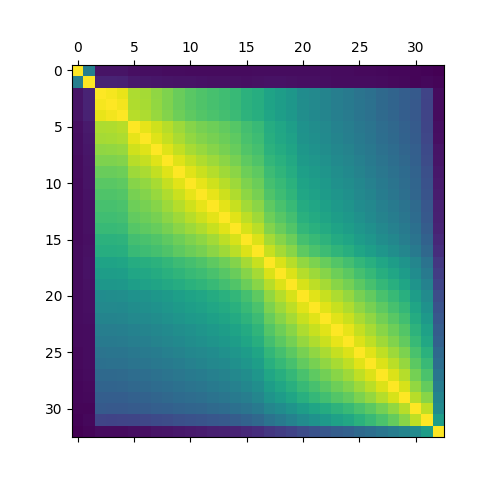}
        \caption{Router-Router. }
        \label{fig:router-router-corr-raven-transfer}
    \end{subfigure}
    \hfill
    \begin{subfigure}{0.16\linewidth}
        \includegraphics[width=\linewidth]{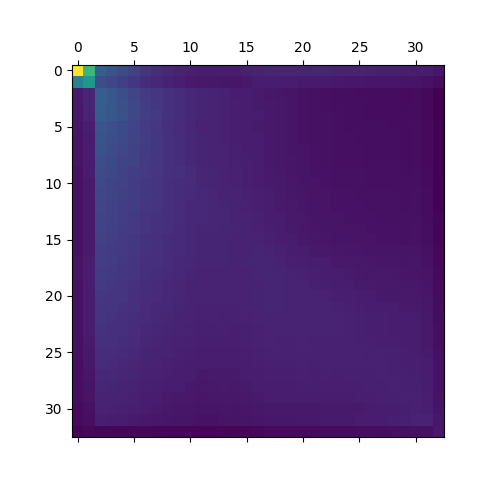}
        \caption{Router-Inv. }
        \label{fig:router-inv-corr-raven-transfer}
    \end{subfigure}
    \hfill
    \begin{subfigure}{0.16\linewidth}
        \includegraphics[width=\linewidth]{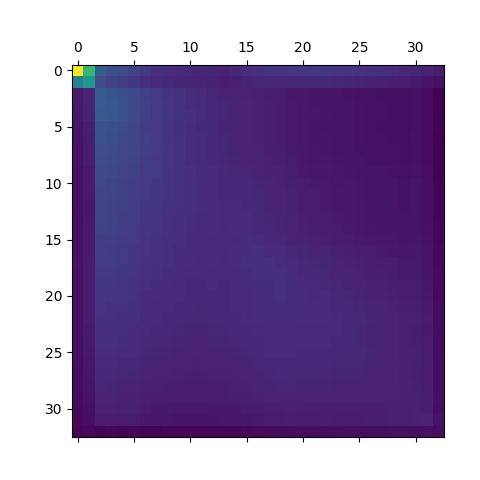}
        \caption{Router-Dom 0. }
        \label{fig:router-0-corr-raven-transfer}
    \end{subfigure}
    \hfill
    \begin{subfigure}{0.16\linewidth}
        \includegraphics[width=\linewidth]{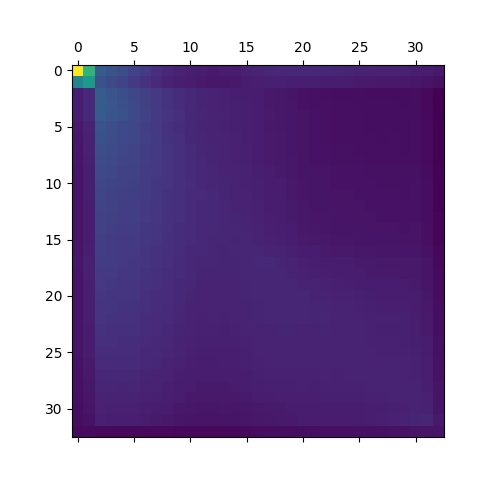}
        \caption{Router-Dom 1. }
        \label{fig:router-1-corr-raven-transfer}
    \end{subfigure}
    \hfill
    \begin{subfigure}{0.16\linewidth}
        \includegraphics[width=\linewidth]{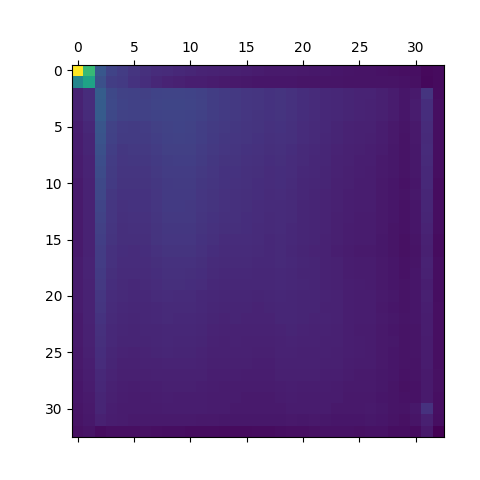}
        \caption{Router-Dom 2. }
        \label{fig:router-2-corr-raven-transfer}
    \end{subfigure}
    \hfill
    \begin{subfigure}{0.16\linewidth}
        \includegraphics[width=\linewidth]{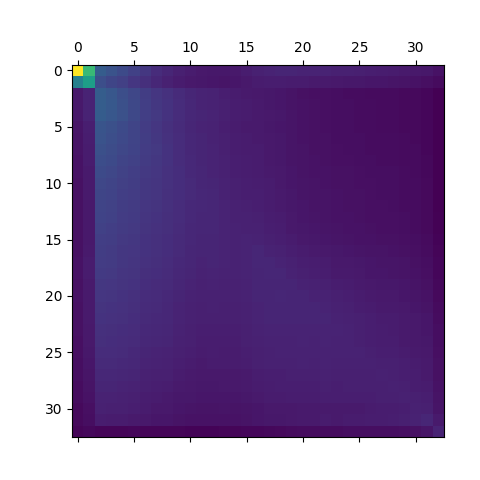}
        \caption{Router-Dom 3. }
        \label{fig:router-3-corr-raven-transfer}
    \end{subfigure}
    \hfill
    \begin{subfigure}{0.16\linewidth}
        \includegraphics[width=\linewidth]{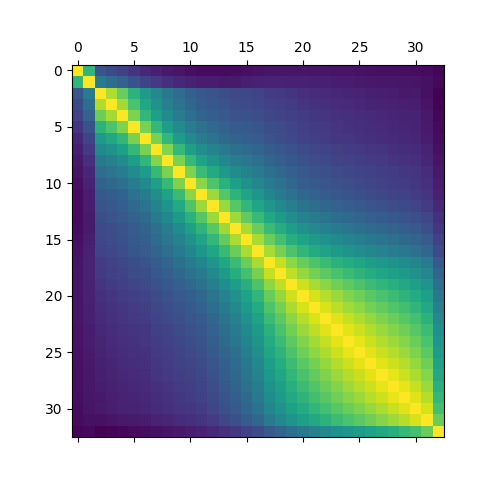}
        \caption{Inv-Inv. }
        \label{fig:inv-inv-corr-raven-transfer}
    \end{subfigure}
    \hfill
    \begin{subfigure}{0.16\linewidth}
        \includegraphics[width=\linewidth]{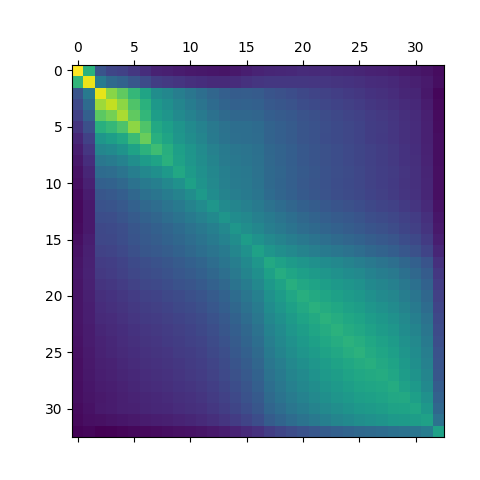}
        \caption{Inv-Dom 0. }
        \label{fig:inv-0-corr-raven-transfer}
    \end{subfigure}
    \hfill
    \begin{subfigure}{0.16\linewidth}
        \includegraphics[width=\linewidth]{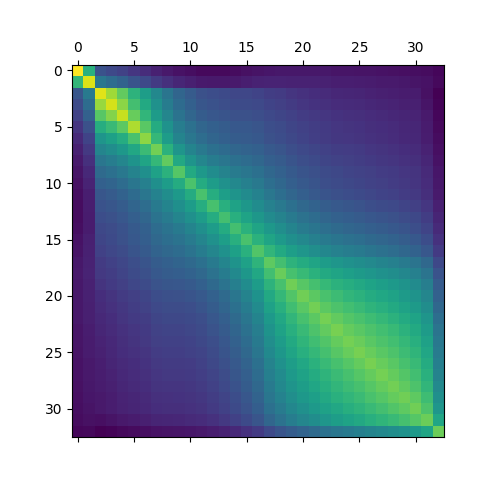}
        \caption{Inv-Dom 1. }
        \label{fig:inv-1-corr-raven-transfer}
    \end{subfigure}
    \hfill
    \begin{subfigure}{0.16\linewidth}
        \includegraphics[width=\linewidth]{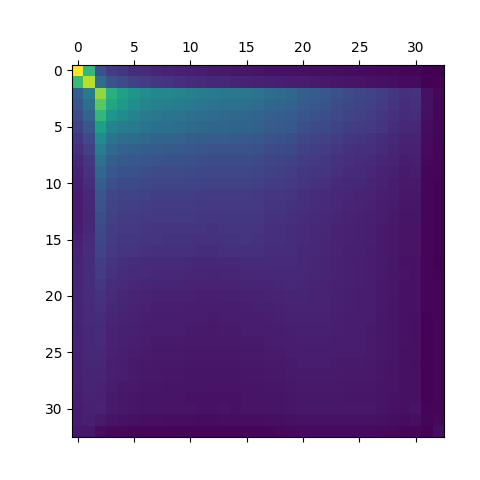}
        \caption{Inv-Dom 2. }
        \label{fig:inv-2-corr-raven-transfer}
    \end{subfigure}
    \hfill
    \begin{subfigure}{0.16\linewidth}
        \includegraphics[width=\linewidth]{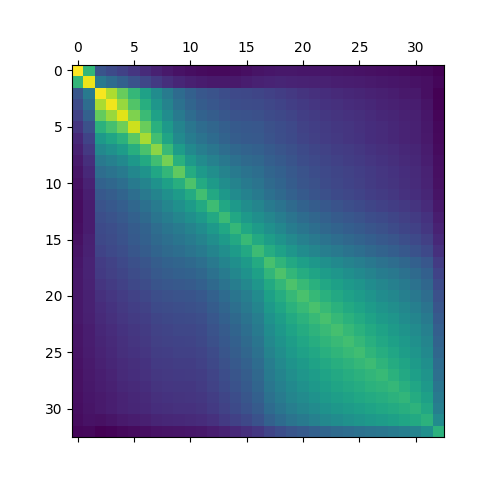}
        \caption{Inv-Dom 3. }
        \label{fig:inv-3-corr-raven-transfer}
    \end{subfigure}
    \hfill
    \begin{subfigure}{0.16\linewidth}
        \includegraphics[width=\linewidth]{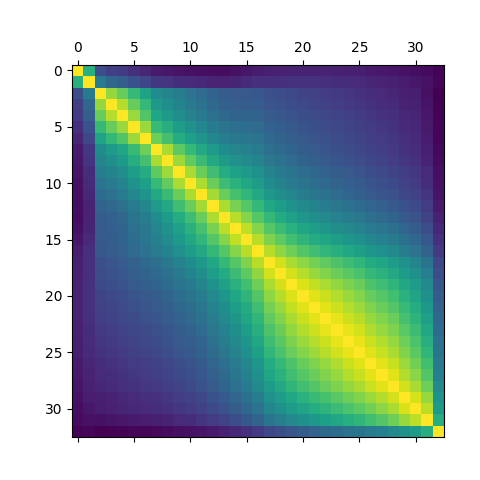}
        \caption{Dom 0-Dom 0. }
        \label{fig:0-0-corr-raven-transfer}
    \end{subfigure}
    \hfill
    \begin{subfigure}{0.16\linewidth}
        \includegraphics[width=\linewidth]{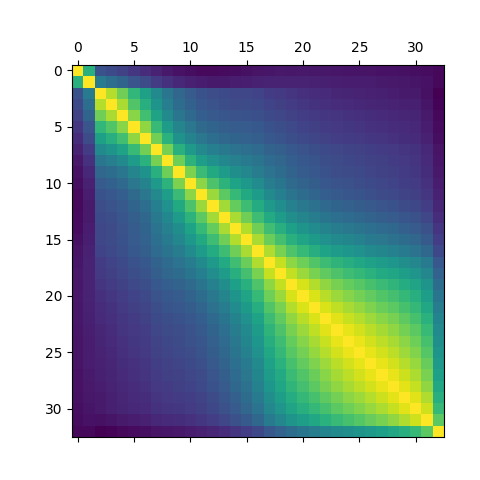}
        \caption{Dom 1-Dom 1. }
        \label{fig:1-1-corr-raven-transfer}
    \end{subfigure}
    \hfill
    \begin{subfigure}{0.16\linewidth}
        \includegraphics[width=\linewidth]{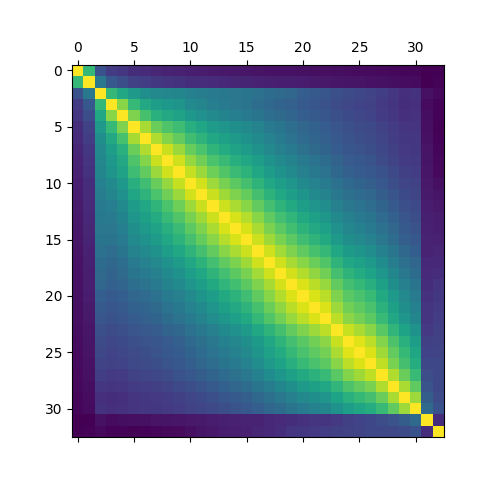}
        \caption{Dom 2-Dom 2. }
        \label{fig:2-2-corr-raven-transfer}
    \end{subfigure}
    \hfill
    \begin{subfigure}{0.16\linewidth}
        \includegraphics[width=\linewidth]{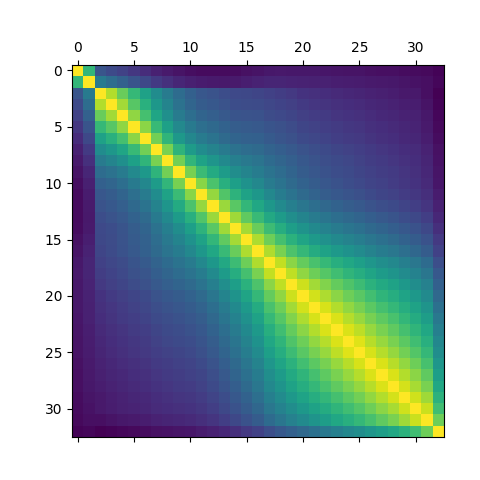}
        \caption{Dom 3-Dom 3. }
        \label{fig:3-3-corr-raven-transfer}
    \end{subfigure}
    \hfill
    \begin{subfigure}{0.16\linewidth}
        \includegraphics[width=\linewidth]{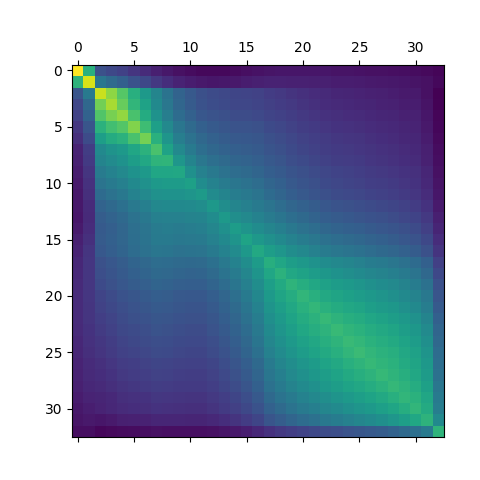}
        \caption{Dom 0-Dom 1. }
        \label{fig:0-1-corr-raven-transfer}
    \end{subfigure}
    \hfill
    \begin{subfigure}{0.16\linewidth}
        \includegraphics[width=\linewidth]{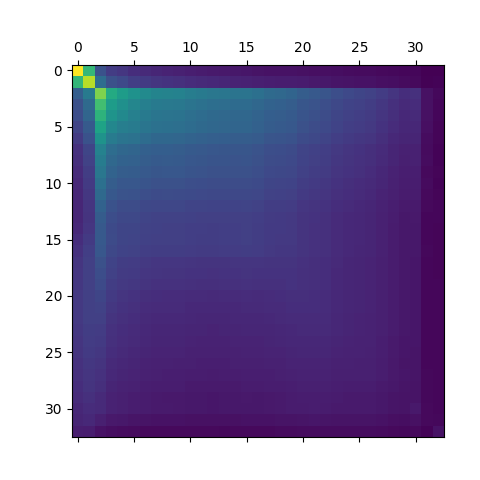}
        \caption{Dom 0-Dom 2. }
        \label{fig:0-2-corr-raven-transfer}
    \end{subfigure}
    \hfill
    \begin{subfigure}{0.16\linewidth}
        \includegraphics[width=\linewidth]{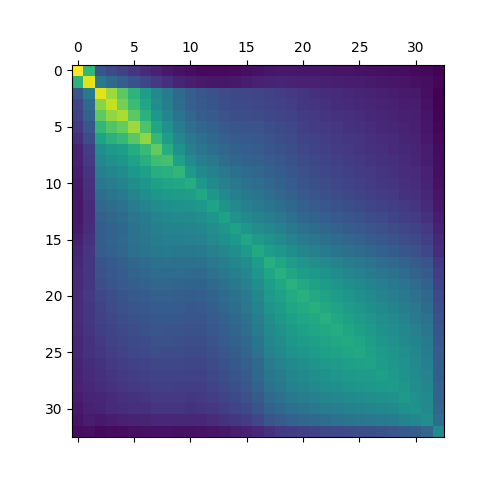}
        \caption{Dom 0-Dom 3. }
        \label{fig:0-3-corr-raven-transfer}
    \end{subfigure}
    \begin{subfigure}{0.16\linewidth}
        \includegraphics[width=\linewidth]{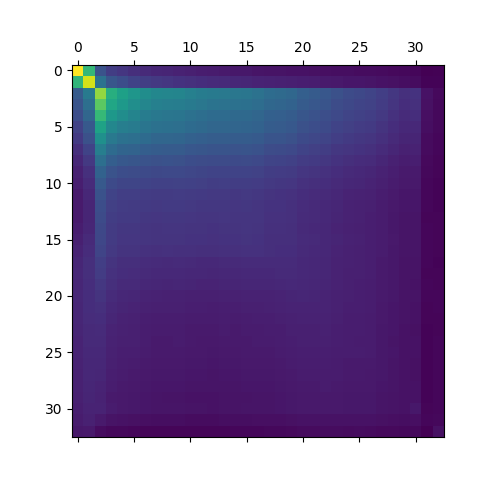}
        \caption{Dom 1-Dom 2. }
        \label{fig:1-2-corr-raven-transfer}
    \end{subfigure}
    \begin{subfigure}{0.16\linewidth}
        \includegraphics[width=\linewidth]{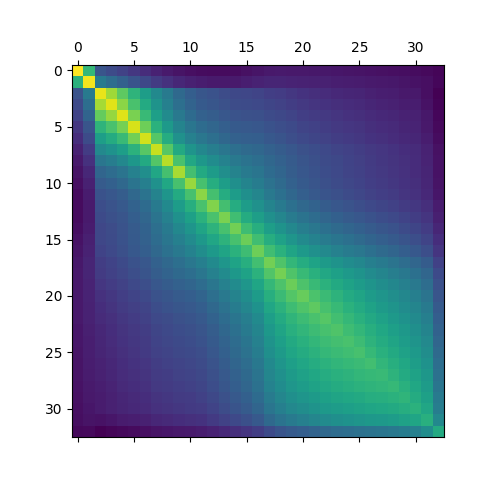}
        \caption{Dom 1-Dom 3. }
        \label{fig:1-3-corr-raven-transfer}
    \end{subfigure}
    \begin{subfigure}{0.16\linewidth}
        \includegraphics[width=\linewidth]{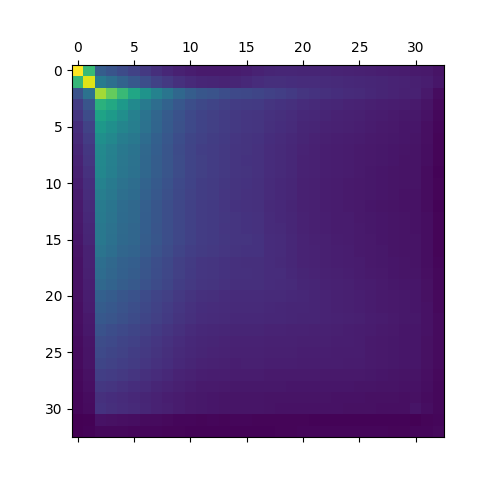}
        \caption{Dom 2-Dom 3. }
        \label{fig:2-3-corr-raven-transfer}
    \end{subfigure}
    \caption{Measures of the Pearson Correlation Coefficient between module hidden states of the transfer learning model during inference on RAVEN dataset. Rows and columns represent layers 0 to 33 of a LLaMA2 module. Router refers to the routing module, Inv refers to the domain-invariant module, and Dom \textit{i} refers to the \textit{i} domain-specific module. The caption indicates the modules used for each row-column pair. }
    \label{fig:all-layer-correlations-raven-transfer}
\end{figure}

We perform the same measures in the transfer learning settings. Figures \ref{fig:all-layer-correlations-acre-transfer} and \ref{fig:all-layer-correlations-raven-transfer} show the results. The studied ICLM model has four modules: while domain-$2$ and dmain-$3$ are well aligned with the respective text and symbolic  RAVEN splits, the two remaining modules do not fully align with the two ACRE splits. Particularly, the domain-$2$ shows a very low level of correlation with the other modules and even with its own layers. This visualisation of the layers confirms the results observed in Table \ref{tab:continual_learning_results} and demonstrates that the poor performance of this module is due to a collapse of the module during the transfer learning: the module inputs are no longer correlated with the module outputs.

